\documentclass[11pt]{article}
\usepackage{amsmath,amsfonts}
\usepackage{subcaption}
\usepackage{multirow}
\usepackage{caption}
\usepackage{float}
\usepackage{natbib}
\usepackage{bbm}
\usepackage{amssymb}
\usepackage{natbib}
\usepackage{url}
\usepackage{mathtools}
\usepackage{makecell}
\usepackage{tablefootnote}
\usepackage[colorlinks,linkcolor=red,citecolor=blue,urlcolor=magenta]{hyperref}

\usepackage{graphicx,xypic,xcolor} %graph

\usepackage{algorithm}
\usepackage{algorithmic}
\usepackage{colortbl}

\usepackage[mathscr]{eucal}
\usepackage{mathrsfs}
\usepackage{caption}
\newtheorem{definition}{Definition}
\newtheorem{proposition}{Proposition}

\newtheorem{remark}{Remark}

\usepackage[singlelinecheck=false]{caption}
\usepackage{enumerate}
\usepackage{amsmath}
\newcommand{\bm}[1]{\boldsymbol{#1}}

\usepackage{natbib}
\newcommand{\bpi}{\boldsymbol{\pi}}
\newcommand{\bPi}{\boldsymbol{\Pi}}

\usepackage{comment}

\allowdisplaybreaks[4]

\graphicspath{{../}}

\title{Reward-Directed Score-Based Diffusion Models via q-Learning}

\author{
	Xuefeng Gao\thanks{Department of Systems Engineering and Engineering Management, The Chinese University of Hong Kong, Hong Kong, China. E-mail: xfgao@se.cuhk.edu.hk}
	\and 
	Jiale Zha\thanks{Department of Systems Engineering and Engineering Management, The Chinese University of Hong Kong, Hong Kong, China. E-mail: jialezha@link.cuhk.edu.hk }
	\and 
	Xun Yu Zhou\thanks{Department of Industrial Engineering and Operations Research and The Data Science Institute, Columbia University, New York, NY 10027, USA. Email: xz2574@columbia.edu}
}

\begin{document}

\maketitle
%\blindmathpaper

\begin{abstract}
We propose a new reinforcement learning (RL) formulation for training continuous-time score-based diffusion models for generative AI to generate samples that maximize reward functions while keeping the generated distributions close to the unknown target data distributions. Different from most existing studies, ours does not involve any pretrained model for the unknown score functions of the noise-perturbed data distributions, nor does it attempt to learn the score functions.  Instead, we formulate the problem as entropy-regularized continuous-time RL and show that the optimal stochastic policy  has a Gaussian distribution with a known covariance matrix.
Based on this result, we parameterize the mean of Gaussian policies and develop an actor--critic type (little) q-learning algorithm to solve the RL problem. A key ingredient in our algorithm design is to obtain noisy observations from the unknown score function  via a ratio estimator. Our formulation can also be adapted to solve pure score-matching and fine-tuning pretrained models.
Numerically, we show the effectiveness of our approach by comparing its performance with two state-of-the-art RL methods that fine-tune pretrained models on several generative tasks including high-dimensional image generations. Finally, we discuss extensions of our RL formulation to probability flow ODE implementation of diffusion models and to conditional diffusion models.

%\noindent{\textbf{Keywords:}}  Generative AI, score-based diffusion models, reward function, continuous-time reinforcement learning, q-learning, stochastic differential equations
\end{abstract}

\section{Introduction}

Diffusion models form a powerful family of probabilistic generative AI models that can capture complex high-dimensional data distributions \citep{Sohl2015, SongErmon2019, Ho2020, SongICLR2021}.
The basic idea is to use a forward process to gradually turn the (unknown) target data distribution to a simple noise distribution, and then reverse this process to generate new samples. A key technical barrier is that the time-reversed backward process involves
a so-called score function that depends on the unknown data distribution; thus learning the score functions (called ``score matching") becomes the main objective of these models.
Diffusion models have achieved state-of-the-art performances in various applications such as image and audio generations \citep{rombach2022high, ramesh2022hierarchical} and molecule generation \citep{hoogeboom2022equivariant, wu2022diffusion}. See, e.g., \cite{yang2022diffusion} for a survey on diffusion models.

In standard diffusion models, the goal is typically to generate new samples whose distribution closely resembles the target data distribution (e.g. ``generate more cat pictures", or ``write a Shakespearean play"). Standard score-based diffusion models are trained (i.e. estimating score functions using neural nets) by minimizing weighted combination of score matching losses \citep{ hyvarinen2005estimation, vincent2011connection, song2020sliced}.
However,
in many applications, we often have preferences about the generated samples from diffusion models (e.g. ``generate prettier cat pictures", or ``write a Shakespearean thriller that happened in New York"). A common way to capture this is to use a reward function, either handcrafted (e.g. a utility function) or learned (through human feedbacks), to evaluate the quality of generated samples related to the preferences. This has led to an interesting  recent line of research on how to adapt standard diffusion models to optimizing reward functions. Several approaches have been proposed, including discrete-time reinforcement learning (RL) \citep{black2023training, fan2024reinforcement}, continuous-time stochastic control/RL \citep{uehara2024fine,tang2024score},
backpropagation of reward function gradient through sampling \citep{clark2023directly}, supervised learning \citep{lee2023aligning}, and guidance \citep{dhariwal2021diffusion}. These studies, however, focus on fine-tuning certain {\it pretrained} diffusion models, whose score functions have already be learned, to maximize the values of additional reward functions.

In this paper, we put forward a significantly different  approach where we directly train a diffusion model from scratch for reward optimization using RL, {\it without involving any pretrained model}. There are two motivations behind this approach. 
One is practical: in many applications, especially for new or special-purpose tasks,  a pretrained model may not even exist  or existing pretrained models may not be easily adapted and fine-tuned. The other is (more importantly) conceptual: fine-tuning a pretrained model is essentially a model-driven approach {\it for dealing with the unknown score function} (i.e. first estimate the score function and then optimize), which is prone to model misspecification and misleading risks.\footnote{We stress the word ``model" here specifically refers to the score function, and ``model misspefication" refers to poor estimations of the score. Existing works on reward maximization and fine-tuning pretrained models may still use a data-driven RL approach to deal with an unknown reward function.}
By contrast, our approach is model-free and data-driven, not relying on a good pretrained model. More on this later.

We consider \textit{continuous-time} score-based diffusion models in \cite{SongICLR2021}, which utilize stochastic differential equations (SDEs) for noise blurring and sample generations. The denoising/reverse process also follows an SDE whose drift includes the unknown score function (the gradient of the log probability density of the noise-perturbed data distribution). We take the continuous-time framework because a) the SDE-based formulation is general and unifies several celebrated diffusion models, including the score matching with Langevin dynamics (SMLD) \citep{SongErmon2019} and the denoising diffusion probabilistic modeling (DDPM) \citep{Ho2020}; b) it leads to not only the SDE-based implementation of score diffusions, but also the probability flow ordinary difference equation (ODE) implementation \citep{SongICLR2021}; and c) there are more analytical tools available for the continuous setting that enable a rigorous and thorough analysis leading to interpretable (instead of black-box) and general
(instead of ad hoc) algorithms.

Due to the need of optimizing rewards of the generated samples,
 we propose a continuous-time RL formulation for adapting diffusion models to reward functions. In this formulation, because the score function in the drift of the denoising process is unknown, we take it as the control (action) variable. This idea of regarding scores as actions, first put forth in \cite{uehara2024fine, tang2024score},
 is quite natural because the problem now has two somewhat competing criteria: score matching (i.e. the generated samples should be close to the true distribution) and terminal reward (i.e. the samples should align with the preferences); so an adjustable  control can be used to achieve the best trade-off between the two.
Specifically, we introduce an objective function that is a weighted combination of two parts: a running reward that penalizes (regularizes) the Kullback--Leibler (KL) divergence of the denoising process from the true score function, and
a terminal reward of the generated samples based on the preference.
This naturally leads to an RL problem with continuous state and action spaces in which the system dynamics are known but the running reward function is unknown (because it involves the true yet unknown score function) and the terminal reward is possibly unknown.
We can therefore adapt and apply the theory and algorithms developed recently for general continuous-time RL with controlled diffusions \citep{Wang20, JZ21, JZ22, jia2023q}.

However, there is a critical issue one needs to address in this formulation. While the general theory in \citep{Wang20, JZ21, JZ22, jia2023q} allows unknown reward functions, it requires access to a (if noisy) observation (a ``reinforcement signal") of the reward every time a state is visited.
It is natural to assume  that we have such signals from the terminal reward (e.g. human rating of aesthetic quality of a generated image). However, obtaining signals from the running reward in our RL problem is subtle, because the KL divergence term in the objective involves the unknown score function. To overcome this difficulty, we express the true score function as the ratio of two expectations with respect to the data distribution. Because we have access to i.i.d. samples from the unknown data distribution, we derive a simple ratio estimator as a noisy observation from the true score value, allowing us to obtain a reinforcement signal from the running reward whenever an action is applied. This procedure is computationally very cheap -- indeed, we will show that a mere {\it single}  sample for computing the ratio estimator already achieves satisfactory performance in an image generation task. % as we do not need to generate new samples from any distributions.

To solve the resulting RL problem for continuous-time diffusion models, we adapt the approach in \cite{jia2023q} to develop theory and algorithms for our RL problem. Specifically, we take
the entropy-regularized, exploratory framework of \cite{Wang20} and optimize over stochastic policies. We show that the optimal stochastic policy for our problem is Gaussian with a known covariance matrix and an unknown mean function.
This key theoretical result suggests that we need to consider Gaussian policies only and parameterize their mean functions when designing RL algorithms.
This insight inspires us to develop an actor--critic type algorithm based on the (little) q-learning theory established in \cite{jia2023q} to solve our RL problem.  We also present a convergence analysis of our algorithm. Furthermore, we implement the q-learning algorithm and evaluate its performances through a series of experiments on three diverse datasets: synthetic data from a one-dimensional Gaussian mixture distribution, a two-dimensional Swiss rolls dataset \citep{sohl2015deep, lai2023fp}, and the CIFAR-10 image dataset \citep{krizhevsky2009learning}. The results demonstrate that our algorithm achieves strong performance in adapting diffusion models to optimize
various reward functions.

Thanks to the continuous-time setting, we can extend our RL formulation to ODE-based models in a straightforward manner. The probability flow ODE implementation of diffusion models is another mainstream approach for sample generation, in addition to the SDE-based one; see, e.g., \cite{song2020denoising}, \cite{SongICLR2021, Karras2022, lu2022dpm}. Compared with SDE-based samplers,  ODE-based deterministic samplers often converge to the data distribution much faster with fewer sampling steps, at the cost of slightly inferior sample quality.
We adapt our continuous-time RL formulation to ODE-based models, where the system dynamics are now described by controlled ODEs. We show that the optimal stochastic policy is still Gaussian, and the algorithm designed for the SDE-based formulation still applies after one replaces the SDE-based sampler with ODE counterparts. We implement the resulting algorithm for two ODE-based samplers: ODE-Euler, which is based on Euler discretization of the probability flow ODE, and DDIM of \cite{song2020denoising}. We find that the algorithm performs well and indeed accelerates the training process compared with the SDE-based formulation. On the other hand, while we mainly focus on unconditional diffusion models, our RL formulation can be readily extended to conditional diffusion models which are used for conditional data generation, such as in text-to-image models.

We now explain the key differences between our work and several closely related ones. \cite{black2023training} propose to fine-tune \textit{discrete-time} diffusion models using RL and directly optimize the reward function (\textit{without} KL regularization).
The denoising process is formulated as a multi-step Markov decision process and the action at each step corresponds to the next denoising state. They then present a policy gradient algorithm, referred to as DDPO, to solve their RL problem.
\cite{fan2024reinforcement} have a similar discrete-time RL formulation, but  add the KL divergence between the fine-tuned model and the pretrained model to the objective to prevent overfitting the reward. They also use a policy gradient method called  DPOK to solve the problem.
Our paper differs from these two in a few important aspects.
First, we consider a continuous-time RL formulation for \textit{continuous-time} diffusion models, where the action is a substitute for the unknown score controlling the drift of the reverse-time SDE. Therefore, our framework and the resulting theory and algorithm are fundamentally different from theirs. Second, compared with \cite{black2023training, fan2024reinforcement} that focus on the DDPM based denoising process with stochastic transitions, our approach not only applies to SDE-based implementation of diffusion models, but also can be readily extended to probability flow ODE implementation. Last but most importantly, in dealing with the unknown score function, the pretrained approach is essentially model-driven, whose performance crucially depends on the availability of a {\it good} pretrained model, whereas our approach is driven by data --  noisy signals from the score. To wit, we penalize the deviation from the (unknown) true score model in our RL objective, rather than from a (known) pretrained model as in \cite{fan2024reinforcement}. As a consequence, our formulation encourages the generated samples to stay close to the true data distribution while maximizing the reward, whereas theirs is to incentivize the samples to stay close to the pretrained data distribution, which is not necessarily always of high quality.

To compare our approach with the fine-tuning approach in \cite{black2023training} and \cite{fan2024reinforcement}, we conduct experiments on the 2-dimensional Swiss Roll data. Although DDPO and DPOK were originally developed for conditional generations, they can be easily adapted to unconditional data generations.
We find that DDPO of \cite{black2023training} suffers from the issue of reward over-optimization, where the generated distribution diverges too far from the original data distribution.
On the other hand, DPOK of \cite{fan2024reinforcement} has a similar performance as our q-learning algorithm, provided that the pretrained model is of good quality. However, if the pretrained model is not good enough in the sense that the pretrained distribution is not close to the data distribution, our q-learning algorithm outperforms DPOK significantly. Notably, this observation holds true for the CIFAR-10 image dataset as well, where our q-learning approach exhibits superior performance compared to DPOK.

We also mention two contemporary  studies \citep{uehara2024fine, tang2024score} that consider fine-tuning  pretrained \textit{continuous-time} diffusion models using entropy regularized control/RL.
The key difference in the problem formulations is again that they penalize the deviation from a \textit{known} pretrained score/diffusion model, while we penalize the deviation from the unknown true score model.
In addition, our objective involves an unknown running reward, making ours a genuine  RL problem instead of a stochastic control one as in \cite{uehara2024fine}. Therefore, our theory and algorithms are conceptually different from these two papers. On the other hand, recent works on GFlowNets (Generative Flow Networks) employ a discrete-time RL formulation to  sample from distributions with given unnormalized probability mass/density functions; see e.g. \cite{bengio2023gflownet, lahlou2023theory, zhang2022unifying}. Moreover, a concurrent work \cite{zhang2024improving} studies fine-tuning discrete-time diffusion models using GFlowNets, with the goal of generating high-reward images with relatively high probability.
Our study is entirely different from this line of work too in that we focus on the generative diffusion setting combined with optimizing reward functions via continuous-time RL.

Finally, it is important to note that our approach and results can be easily modified to cover pretrained and fine-tuning
formulation as a special case. Indeed, if a score has been trained and given, then we can generate the reinforcement signals by simply computing the trained score function values instead of using the ratio estimator, and the rest is entirely the same. 

The remainder  of the paper is organized as follows. In Section~\ref{sec:prelim}, we briefly review continuous-time score-based diffusion models. In Section~\ref{sec:prob-sde}, we discuss the continuous-time RL formulation for reward maximizations in diffusion models. Section~\ref{sec:theory} presents our main theoretical result, based on which we design an actor--critic type q-learning algorithm in Section~\ref{sec:algorithm}. In Section~\ref{sec:experiment}, we present the results of the numerical experiments for two toy examples, whereas in Section~\ref{sec:image}, we test our algorithm on a reward-directed image generation task.
Section~\ref{sec:ext} highlights extensions to ODE models and conditional diffusion models. Finally, Section~\ref{sec:conclusion} concludes.

%%%%%%%%%%%%%%%%%%%%%%
\section{\mbox{Quick Review on Continuous-Time Score-Based Diffusion Models}}\label{sec:prelim}
For reader's convenience we briefly recall the continuous-time score-based diffusion models with SDEs \citep{SongICLR2021}. Denote by $p_{0} \in \mathcal{P} (\mathbb{R}^d) $ the unknown continuous data distribution, where $\mathcal{P}(\mathbb{R}^{d})$ is the space of all probability measures on $\mathbb{R}^{d}$. Given i.i.d samples from $p_{0}$, standard diffusion models aim to generate new samples whose distribution closely resembles the data distribution. In this classical setting, one does not consider the preference/reward of the generated samples.

\begin{itemize}

\item\textbf{Forward process and reverse process.}

Fix $T>0.$ We consider a $d-$dimensional forward process $(\mathbf{x}_{t})_{t \in [0, T]}$, which is a Ornstein–-Uhlenbeck (OU) process satisfying the following SDE
\begin{equation}\label{OU:SDE}
d\mathbf{x}_{t}=-f(t)\mathbf{x}_{t}dt+g(t)d\mathbf{B}_{t},
\end{equation}
where $\mathbf{x}_{0}$ is a random variable following the (unknown target) distribution $p_{0}$, $(\mathbf{B}_{t})$ is a standard $d$-dimensional Brownian motion which is independent of $\mathbf{x}_{0}$, and both $f(t)\geq0$ and $g(t)\geq0$ are (prescribed and known) scalar-valued continuous functions of time $t$.  The solution to \eqref{OU:SDE} is
\begin{equation}\label{SDE:solution}
\mathbf{x}_{t}=e^{-\int_{0}^{t}f(s)ds}\mathbf{x}_{0}+\int_{0}^{t}e^{-\int_{s}^{t}f(v)dv}g(s)d\mathbf{B}_{s}, \quad t \in[0, T].
\end{equation}
Note that the  forward model \eqref{OU:SDE} is fairly general and it includes variance exploding SDEs, variance preserving SDEs, and sub variance preserving SDEs that are commonly used in the literature; see \cite{SongICLR2021} for details.

Denote by $p_t(\cdot)$ the probability density function of $\mathbf{x}_{t}$, and $p_{t|0}(\cdot | \mathbf{x}_{0})$ the density function of $\mathbf{x}_{t}$ given $\mathbf{x}_{0}$ for $t \in[0, T].$
By \eqref{SDE:solution}, $p_{t|0}(\cdot | \mathbf{x}_{0})$ is Gaussian which has an analytical form. %We only use this fact in \eqref{eq:scoreMC} in our analysis.

Now consider the reverse (in time) process $(\tilde{\mathbf{x}}_{t})_{t\in[0,T]}$, where
\begin{align}\label{eq:rev}
\tilde{\mathbf{x}}_{t} := \mathbf{x}_{T-t}, \quad  t\in[0,T].
\end{align}
Under mild assumptions, the reverse process still satisfies an SDE \citep{Anderson1982, haussmann1986time, cattiaux2023time}:  %For technical reasons, we assume the score function $\nabla\log p_{t}(\mathbf{x})$ has polynomial growth in $\mathbf{x}$ throughout the paper.
\begin{equation}\label{eq:yt}
d\tilde{\mathbf{x}}_{t}=[f(T-t)\tilde{\mathbf{x}}_{t}+(g(T-t))^{2}\nabla\log p_{T-t}(\tilde{\mathbf{x}}_{t})]dt
+g(T-t)dW_t,  \quad  t\in[0,T], %\quad \tilde{\mathbf{x}}_{0}\sim p_{T},
\end{equation}
where $(W_t)$ is a standard Brownian motion in $\mathbb{R}^d$, and the term $\nabla_{\mathbf{x}}\log p_{t}(\cdot)$ in \eqref{eq:yt} is called the {\it score function}. By \eqref{eq:rev}, the reverse process starts from a random location $\tilde{\mathbf{x}}_{0}\sim p_{T}$, where $p_T$ is the probability density/distribution of $\mathbf{x}_T$ (here and henceforth, probability distribution and probability density function are used interchangeably). Thus, at time $T$, we have
$\tilde{\mathbf{x}}_{T}\sim p_{0}$, where $p_0$ is the target distribution we want to generate samples from.
However, the distribution $p_{T}$ is unknown because it depends on the unknown target distribution $p_{0}$. Because
the aim of generative diffusion models is to convert random noises into random samples from $p_0$, by (2) we can choose a random (Gaussian) noise $\nu$, with
\begin{equation} \label{eq:hatp}
\nu:=\mathcal{N}\left(\mathbf{0},\int_{0}^{T}e^{-2\int_{t}^{T}f(s)ds}(g(t))^{2}dt\cdot I_{d}\right),
\end{equation}
as a proxy to $p_T$, where $I_d$ is the $d-$dimensional identity matrix; see \cite{SongICLR2021, leeconvergence}.
Note that $\nu$ is the distribution of the random variable $\int_{0}^{t}e^{-\int_{s}^{t}f(v)dv}g(s)d\mathbf{B}_{s}$ in
\eqref{SDE:solution}, which is easy to sample because it is Gaussian, and will be henceforth referred to as the {\it prior distribution}.\footnote{Although one could take an empirical estimate of $\mathbb{E}[\mathbf{x}_T] = e^{-\int_{0}^{T}f(s)ds} \mathbb{E}[\mathbf{x}_{0}]$ as the mean of the prior distribution $\nu$, in practice $\nu$ is often chosen with mean zero because of two reasons: a) one can rescale the data to make them to have zero mean; see e.g. \cite{Ho2020}; and b) the mean is close to zero when $T$ is sufficiently large due to the exponential discounting (for variance preserving SDEs).}
%With a large amount of noise added to the data in the forward process, $p_T$ is
%practically indistinguishable from the pure Gaussian noise $\nu$.

\begin{remark}
When considering variance preserving SDEs where $f(t) = \frac{1}{2}\alpha(t) $ and $g(t) = \sqrt{ \alpha(t) }$ for some nondecreasing positive function $\alpha$, it is also common to use the stationary distribution of the forward SDE, which is $\mathcal{N}(0, I_{d})$, as the prior distribution. Taking $\alpha(t) \equiv 2$ as an example, we deduce from \eqref{eq:hatp} that $\nu=\mathcal{N}\left(0, \left(1 - e^{-2T}\right)\cdot I_d\right)$, which converges to $\mathcal{N}(0, I_{d})$ as $T$ becomes large.
\end{remark}

In view of \eqref{eq:yt}, we now consider the SDE:
\begin{equation}\label{eq:zt}
d\mathbf{z}_{t}=\left[f(T-t)\mathbf{z}_{t}+(g(T-t))^{2}\nabla\log p_{T-t}(\mathbf{z}_{t})\right]dt
+g(T-t)dW_t, \quad \mathbf{z}_{0}\sim\nu,
\end{equation}
where $\nu$ is independent of the Brownian motion $W$.
Because $\nu \approx p_T$, one expects that the distribution
of $\mathbf{z}_{T}$ will be close to that of $\tilde{\mathbf{x}}_{T}$ which is $p_{0}$.

%%%%%%%%%%%%%%%%%%%%%%%%%%%%%%%

\item\textbf{Training diffusion models via score matching.}

The score function $\nabla_{\mathbf{x}}\log p_{t}(\cdot)$ in \eqref{eq:zt} is unknown because the data distribution $p_0$ is unknown. One can estimate it with a time--state score model $s_{\theta}(\cdot,\cdot)$, which is often a deep neural network parameterized by $\theta$, by minimizing the score matching loss:
\begin{align}\label{eq:score-matching}
& \min_{\theta} \mathbb{E}_{t \sim U[0, T]}  \left[ \lambda(t) \mathbb{E}_{\mathbf{x}_{t}}  \left\Vert s_{\theta}(t,\mathbf{x}_t)-\nabla_{\mathbf{x}_{t}}\log p_{t}(\mathbf{x}_{t})\right\Vert^2   \right] .
\end{align}
Here, $\lambda(\cdot): [0, T] \rightarrow \mathbb{R}_{>0}$ is some positive weighting function (e.g. $\lambda(t)= g(t)^2$), and $U[0, T]$ is the uniform distribution on $[0, T].$  This objective is intractable because $\nabla_{\mathbf{x}}\log p_{t}(\cdot)$ is unknown. Several approaches have been developed in literature to tackle this issue, including denoising score matching \citep{vincent2011connection, SongICLR2021}, sliced score matching \citep{song2020sliced} and implicit score matching \citep{hyvarinen2005estimation}. Here, we take denoising score matching for illustration. One can show that \eqref{eq:score-matching} is equivalent to the following objective
\begin{align}\label{eq:score-matching2}
\min_{\theta} \mathbb{E}_{t \sim U[0, T]}  \left[ \lambda(t) \mathbb{E}_{\mathbf{x}_{0}}  \mathbb{E}_{\mathbf{x}_{t} | \mathbf{x}_{0}} \left\Vert s_{\theta}(t,\mathbf{x}_t)-\nabla_{\mathbf{x}_{t}}\log p_{t|0}(\mathbf{x}_{t} | \mathbf{x}_{0})\right\Vert^2   \right],
\end{align}
where $\mathbf{x}_{0} \sim p_0$ is the data distribution, and $p_{t|0}(\cdot | \mathbf{x}_{0})$ is the density of $\mathbf{x}_{t}$ given $\mathbf{x}_{0}$, which is Gaussian by \eqref{SDE:solution}. Because we have access to i.i.d. samples from $p_0$ (i.e. training data),  the objective in \eqref{eq:score-matching2} can be approximated by Monte Carlo, and the resulting loss function can be then optimized using e.g. stochastic gradient descent.

\item\textbf{Algorithms.}

After the score function is estimated,  the true score function in \eqref{eq:zt} is replaced by the estimated score $s_{\theta}$, and  the reverse SDE \eqref{eq:zt} is discretized to obtain an implementable algorithm. Alternatively, one can use the probability flow ODE based implementation; see Section 4 of \cite{SongICLR2021} for details. The generated samples at time $T$ are expected to follow approximately the data distribution, {\it provided} that the score is estimated accurately.

\end{itemize}
%%%%%%%%%%%%%%%%%%%%%%%%%%%%%%%%%

\section{Problem Formulation}\label{sec:prob-sde}
Standard training of diffusion models via score matching \eqref{eq:score-matching2} -- called {\it pretrained models} -- does not consider preferences about the generated samples.
 A standard way to capture preferences is to add a reward function that evaluates the quality of generated samples related to the specified  preferences. This is essentially a model-based approach -- first learn a pre-trained model by estimating the score function and then optimize. This paper takes a conceptually different approach, one that is in the core spirit of RL, namely, to learn optimal policies {\it directly} without attempting to first learn a model. This leads naturally to a continuous-time RL formulation.

\subsection{Reward-directed diffusion models}

The  key idea is that because the score term $\nabla\log p_{T-t}(\mathbf{z}_{t})$ in
\eqref{eq:zt} is unknown, we regard it as a control. With a reward function, this leads to
the following stochastic control problem:
\begin{align}\label{eq:CT-RL}
\max_{{\mathbf{a}} = (a_t: 0 \le t \le T)}  \left\{ \beta \cdot \mathbb{E}[h( \mathbf{y}^{\mathbf{a}}_{T})] -  \mathbb{E} \left[ \int_0^T (g(T-t))^{2} \cdot {| \nabla\log p_{T-t}({\mathbf{y}}^{\mathbf{a}}_{t}) - a_t|^2} dt \right]  \right\}
\end{align}
subject to
\begin{equation}\label{eq:u-sys}
d\mathbf{y}^{\mathbf{a}}_{t}=\left[f(T-t)\mathbf{y}^{\mathbf{a}}_{t}+(g(T-t))^{2} a_t \right]dt
+g(T-t)d W_{t}, \quad \mathbf{y}_{0}\sim\nu.
\end{equation}
Here, $a_t \in \mathbb{R}^d$ denotes the control action at time $t$, $(\mathbf{y}^{\mathbf{a}}_{t})$ is the controlled state process, $h$ is the terminal reward function for the generated sample $ \mathbf{y}^{\mathbf{a}}_{T},$ and $\beta \ge 0$ is a weighting coefficient.

The first term in the objective function \eqref{eq:CT-RL} captures the preference on
the generated samples. In this study, we do not require the reward/utility function $h$ to be differentiable as in \cite{clark2023directly}, nor do we necessarily assume that its functional form is known. What we do assume is that given a generated sample $\mathbf{y}^{\mathbf{a}}_{T}$ we can obtain a noisy observation (``a reward signal") of the reward $h$ evaluated at that sample (e.g. a human evaluation of the aesthetic quality of an image).

The second term in the objective \eqref{eq:CT-RL} has the following interpretation. Consider a deterministic feedback policy so that $a_t = w(t, \mathbf{y}_t)$ for some deterministic function $w$.
Let $\mathbb{P}^{\mathbf{z}}$ and $\mathbb{P}^{\mathbf{y}^{\mathbf{a}}}$ be the induced distribution (i.e., path measures over $C([0, T], \mathbb{R}^d)$) by the SDEs
\eqref{eq:zt} and \eqref{eq:u-sys}, respectively. Then Girsanov's theorem gives that under some regularity conditions (see, e.g.  \citealp[Appendix C]{uehara2024fine} or \citealp[Proposition 3.3]{tang2024fine})
\begin{align}\label{eq:KL}
 \text{KL} (\mathbb{P}^{\mathbf{y}^{\mathbf{a}}} || \mathbb{P}^{\mathbf{z}}) & = \frac{1}{2} \mathbb{E} \left[ \int_0^T (g(T-t))^{2} \cdot{| \nabla\log p_{T-t}({\mathbf{y}}^{\mathbf{a}}_{t}) - w(t, \mathbf{y}_t^{\mathbf{a}}) |^2} dt \right] \nonumber \\
& = \frac{1}{2} \mathbb{E} \left[ \int_0^T (g(T-t))^{2} \cdot{| \nabla\log p_{T-t}({\mathbf{y}}^{\mathbf{a}}_{t}) - a_t|^2} dt \right],
\end{align}
where the expectation is taken with respect to $(\mathbf{y}^{\mathbf{a}}_{t})$ where $\mathbf{y}_{0}\sim\nu.$ This justifies the second term of \eqref{eq:CT-RL} in terms of KL-divergence.

\begin{remark}[Connections between \eqref{eq:KL} and score matching]
	It is worth noting that if we let $\mathbf{y}_{0}\sim{p}_{T}$ and $\mathbf{z}_{0}\sim{p}_{T}$, then we can compute that
	\begin{align}\label{eq:KL2}
		\text{KL} (\mathbb{P}^{\mathbf{z}}|| \mathbb{P}^{\mathbf{y}^{\mathbf{a}}} ) &= \frac{1}{2} \mathbb{E} \left[ \int_0^T (g(T-t))^{2} \cdot{| \nabla\log p_{T-t}({\mathbf{z}}_{t}) - w(t, \mathbf{z}_t) |^2} dt \right] \nonumber \\
		&= \frac{1}{2} \mathbb{E} \left[ \int_0^T (g(t))^{2} \cdot{| \nabla\log p_{t}({\mathbf{z}}_{T-t}) - w(T-t, \mathbf{z}_{T-t}) |^2} dt \right] \nonumber \\
		&= \frac{1}{2} \mathbb{E} \left[ \int_0^T (g(t))^{2} \cdot{| \nabla\log p_{t}({\mathbf{x}}_{t}) - w(T-t, \mathbf{x}_{t}) |^2} dt \right],
	\end{align}
	where the last equality follows from the time reversal of diffusion processes (see \eqref{eq:yt}), and the expectation there is taken with respect to the randomness in the forward process $({\mathbf{x}}_{t})$ in
	\eqref{OU:SDE}. This is exactly the score matching loss in \eqref{eq:score-matching} with the likelihood weighting $\lambda(t) = \frac{1}{2}g(t)^2$;  see \cite{song2021maximum} for details. In particular, it was shown in \cite{song2021maximum} that $\text{KL} (\mathbb{P}^{\mathbf{z}}|| \mathbb{P}^{\mathbf{y}^{\mathbf{a}}} ) \ge \text{KL} (p_0 || \mathbf{y}^{\mathbf{a}}_{T})$ when $\mathbf{y}_0 \sim p_T$ and
	$\mathbf{z}_{0}\sim{p}_{T}$, and \eqref{eq:KL2}  serves as an efficient proxy for maximum likelihood training of score-based generative models.
	Finally, the hyperparameter $\beta$ in \eqref{eq:CT-RL} represents the weights placed on score matching and human preferences. When $\beta=0$, the problem \eqref{eq:CT-RL}
	reduces to a score matching problem. Thus, our formulation can be used also for pure score matching.
\end{remark}

Denote
\begin{align}\label{eq:rt}
 r(t, \mathbf{y}, a) := -(g(T-t))^{2} \cdot {| \nabla\log p_{T-t}({\mathbf{y}}) - a|^2},
\end{align}
 the  running reward function (or instantaneous reward) at time $t$ in the objective \eqref{eq:CT-RL}. Because the true score function $\nabla\log p_{T-t}(\cdot)$ is unknown, this running/instantaneous reward function is also unknown. Moreover, recall that the true terminal reward function $h$ is also generally unknown.
Hence, we need an RL formulation of the problem \eqref{eq:CT-RL}--\eqref{eq:u-sys}, where exploration is necessary due to these unknown rewards. This is discussed in the next subsection.

%%%%%%%%%%%%%%%%%%%%%%%%%

\subsection{Stochastic policies and exploratory formulation}\label{sec:explora}

We now adapt the exploratory formulation for continuous-time RL in \cite{Wang20,jia2023q} to our problem setting. % and develop q-learning algorithms to solve our RL problem.
A distinctive feature of our problem is that  the RL agent knows the environment (the functions $f, g$ are known in the SDE model \eqref{eq:u-sys}), but she does not know the instantaneous reward function $r$  and the terminal reward function $h$. So she still needs to do ``trial and error" -- to try a strategically designed sequence of actions, observe the corresponding state process and a stream of running rewards and terminal reward samples/signals, and continuously update and improve her action plans based on these observations. For this purpose, the agent employs stochastic policies in our RL setting.

Mathematically, let $\bpi: (t,\mathbf{y}) \in [0,T] \times \mathbb{R}^d \rightarrow  \bpi (\cdot| t, \mathbf{y}) \in \mathcal{P} (\mathbb{R}^d)$ be a given stochastic feedback policy, where $\mathcal{P} (\mathbb{R}^d)$ is a collection of probability density functions defined on $\mathbb{R}^d$.
For such a stochastic policy $\bpi,$  \cite{Wang20,jia2023q} propose  an ``exploratory" formulation with the following ``averaged" dynamic
        \begin{align}\label{eq: exp_state_SDE}
        d \tilde {\mathbf{y}}^{\bpi}_s= \left[f(T- s) \tilde{\mathbf{y}}^{\bpi}_{s} +g^2(T-s) {\int_{\mathbb{R}^d} a {\bpi}(a|s, \tilde {\mathbf{y}}^{\bpi}_s ) da} \right]  ds + g(T-s)  dW_s  , \quad \tilde{\mathbf{y}}_{0}^{\bpi} \sim\nu.
    \end{align}
In contrast to \eqref{eq:u-sys}, here the action randomization has been averaged out in the drift term of \eqref{eq: exp_state_SDE}.
The associated entropy-regularized control objective is given by:
\begin{align} \label{eq: value_function}
        &J(t, y, \bpi) \nonumber\\
        = &\mathbb{E}_{t,y}\bigg[  \int_t^T \int_{\mathbb{R}^d} \big( {r(s, \tilde{\mathbf{y}}_s^{\bpi}, a)}\ - { \theta \log \bpi(a|s, \tilde{\mathbf{y}}_{s}^{\bpi})\big) \bpi(a|s, \tilde{\mathbf{y}}_{s}^{\bpi}) da} ds + {\beta h(\tilde{\mathbf{y}}_T^{\bpi} )}\bigg],
    \end{align}
    where the reward has also been averaged over action randomization, and
    $\theta>0$ is the temperature parameter that controls the level of exploration.\footnote{The entropy regularization is a commonly used technique to improve exploration in RL, originally introduced for MDPs; see e.g. \cite{haarnoja2018soft}.}
    The function $J(\cdot, \cdot, \bpi)$ in \eqref{eq: value_function} is called the (exploratory) value function of the stochastic policy $\bpi$. The goal of RL is to solve the following optimization problem: 
\begin{align*} %\label{eq:Jstar}
	\max_{\bpi \in \bPi} \int J(0, y, \bpi) d \nu(y),
\end{align*}
where $\boldsymbol{\Pi}$ is the set of admissible stochastic policies to be specified  in Definition~\ref{def:policy} below. Define
\begin{align}\label{eq:Jstar2}
J^*(t, y) =	\max_{\bpi \in \bPi} J(t, y, \bpi).
\end{align}
%One can expect that
As the temperature parameter $\theta \rightarrow 0$,
the function $J^*$ in \eqref{eq:Jstar2} converges to the optimal value function of the original control problem (without entropy regularization) in \eqref{eq:CT-RL} under some technical conditions \citep{tang2022exploratory}. Hence, solving the exploratory RL problem \eqref{eq:Jstar2} when $\theta$ is small provides an approximate solution to \eqref{eq:CT-RL}. %In the next section we provide our theoretical results.

\begin{definition}\label{def:policy}
	A policy $\bpi = \bpi(\cdot| \cdot, \cdot)$ is called admissible, if
	\begin{enumerate}[(i)]
		\item $\bpi(\cdot| t,y) \in \mathcal{P} (\mathbb{R}^d)$,  $\text{supp}\;\bpi(\cdot| t,y) = \mathbb{R}^d$ for every $(t,y) \in [0, T] \times \mathbb{R}^d$ and $\boldsymbol{\pi} (a| t,y): (t,y,a) \in [0, T] \times   \mathbb{R}^d \times \mathbb{R}^d \rightarrow \mathbb{R}$ is measurable;
		
		\item $\int_{\mathbb{R}^d}\left|\bpi(a|t,y)-\bpi(a|t',y')\right|da\to 0$ as $(t',y')\to (t,y)$. Moreover, there is a constant $C>0$ independent of $(t,a)$ such that
		\begin{equation*}
			\int_{\mathbb{R}^d}\left|\bpi(a|t,y)-\bpi(a|t,y')\right|da\leq C |y-y'|,\ \forall y,y'\in\mathbb{R}^d;
		\end{equation*}
	
		\item $\forall (t,y)$, $\int_{\mathbb{R}^d} |\log \boldsymbol{\pi}(a|t,y)| \boldsymbol{\pi}(a |t,y) da \le C (1+ |y|^p)$ for some $p\geq 1$ and $C$ is a positive constant; for any $k\geq 1$, $\int_{\mathbb{R}^d} |a|^k\bpi(a|t,y) da \le C_k (1+ |y|^{k'})$  for some $k'\geq 1$ and $C_k$ is a positive constant that can depend on $k$.

	\end{enumerate}
\end{definition}

%%%%%%%%%%%%%%%%%%%%%%%%%%%%%%%%%%%%%%%%%%%%%%

\section{Theory}\label{sec:theory}
In this section, we present our main theoretical results, which provide the optimal stochastic policy to the RL problem in \eqref{eq:Jstar2} with
the system dynamic \eqref{eq:u-sys} and the running reward \eqref{eq:rt}.  Introduce the (generalized) Hamiltonian $H: [0, T] \times \mathbb{R}^d \times \mathbb{R}^d \times \mathbb{R}^d  \times \mathbb{R}^{d \times d} \rightarrow \mathbb{R}$ associated with the original problem  \eqref{eq:CT-RL}--\eqref{eq:u-sys} (see e.g. \citealp{yong2012stochastic}):
\begin{align}\label{eq:Hal}
&H(t, y, a, p,q) \nonumber \\
&=   -(g(T-t))^{2}  | \nabla\log p_{T-t}(y) - a|^2  +  [f(T-t) y +(g(T-t))^{2} a ] \circ p + \frac{1}{2}(g(T-t))^{2} \circ  q.
\end{align}
Equation (13) of \cite{jia2023q} yields that the optimal stochastic policy $\bpi^* (\cdot| t, y)$ for the problem \eqref{eq:Jstar2} is given by
\begin{align*} %\label{eq:opt-pol}
	\boldsymbol{\pi}^*(a| t, y) \propto \exp \left(\frac{1}{\theta} H(t,y, a, J_y^*(t, y), J_{yy}^*(t, y))  \right),
\end{align*}
where $J_y^*$ and $J_{yy}^*$ are the first and second order partial derivatives of $J^*$ with respect to $y$, respectively.
However, $H$ in our case, \eqref{eq:Hal}, is quadratic in $a$, leading to the following result.

%Because the system dynamics \eqref{eq:u-sys} is linear in action $a$, and the running reward \eqref{eq:rt} is quadratic in $a$, it follows that the Hamiltonian $H$ \eqref{eq:Hal} is quadratic in $a$. Hence, we immediately obtain the following result from Equation (13) of\cite{jia2023q}.
\begin{proposition}\label{prop:optimal-policy}
The optimal stochastic policy $\bpi^* (\cdot| t, y)$ is a Gaussian distribution in $\mathbb{R}^d$:
\begin{align}\label{eq:pi-star}
\bpi^{*}(\cdot|t,y)   \sim  \mathcal{N}\left(\mu^{*}(t,y), \frac{\theta}{2 g^2(T-t)}  \cdot I_{d} \right) , %\left(\cdot \mid  \right)
\end{align}
where
\begin{align}\label{eq:mustar}
\mu^{*}(t,y) =  \nabla\log p_{T-t}(y)  + \frac{1}{2} \cdot J_y^*(t, y).
\end{align}
\end{proposition}
This result provides several interesting insights. First, the mean $\mu^{*}(t,y)$ of the optimal stochastic policy consists of two parts: the score $\nabla\log p_{T-t}(y)$ that we try to match in \eqref{eq:CT-RL}, and an additional term $\frac{1}{2} \cdot J_y^*(t, y)$ that arises due to the consideration of maximizing the terminal reward $h$ of the generated samples. The optimal value function $J^*(t, y) $ in \eqref{eq:Jstar2} can be shown to satisfy the following so-called exploratory HJB equation, a nonlinear PDE (see Equation (14) of \citealp{jia2023q}):
\begin{align}\label{eq:expHJB}
&\frac{\partial J^*}{ \partial t} (t, y ) + \theta \log \left[   \int_{\mathbb{R}^d} \exp \left(\frac{1}{\theta} H(t, y, a, J^*_y,  J^*_{yy})  \right)  da \right]  = 0, \\
& J^*(T, y) =\beta \cdot h(y).
\end{align}
Note that we do not attempt to solve this PDE  because the Hamiltonian $H$ involves the unknown true score function (we never solve any HJB equations in the realm of RL).
However, \eqref{eq:expHJB} illustrates the impacts of the terminal reward function $h$ as well as the temperature parameter $\theta$ and the score $\nabla\log p_{T-t}(y)$ on the mean of the optimal Gaussian policy $\eqref{eq:mustar}$.
%In particular, one has $J_u^*(t, u) = \beta \cdot \nabla h(u)$ typically only when $t=T$, if $h$ is differentiable.
 We also observe from \eqref{eq:pi-star} that the optimal Gaussian policy has a covariance matrix $\frac{\theta}{2 g^2(T-t)}  \cdot I_{d}$, whose magnitude is, naturally,  proportional to the temperature parameter $\theta>0$.
 Meanwhile, the exploration level is inversely proportional to $g^2(T-t).$ This is intuitive because $g$ represents the strength of the noise we add to blur the original samples. The higher this noise the less {\it additional} noise we need for exploration. %The Technically, this is because the weight $g^2(T-t)$ appears as the quadratic coefficient in the Hamiltonian \eqref{eq:Hal} (as a function of the action), which in turn originates from the KL-divergence interpretation \eqref{eq:KL} of the second term in the objective \eqref{eq:CT-RL}. If we
%penalize the deviations of actions from the true sore function by an unit weight instead of $g^2$ in \eqref{eq:CT-RL}, then the exploration noise will not depend on $g$ anymore, even though the dynamics \eqref{eq:u-sys} depends on $g$.

Inspired by Proposition~\ref{prop:optimal-policy}, in our RL algorithm to be presented in Section~\ref{sec:algorithm}, we only need to consider Gaussian policies in the following form:
\begin{align}\label{eq:pi-psi}
\bpi^{\psi}(\cdot|t,y)   \sim  \mathcal{N}\left(\mu^{\psi}(t,y), \frac{\theta}{2 g^2(T-t)}  \cdot I_{d} \right) \quad \text{for all $(t, y)$,}%\left(\cdot \mid  \right)
\end{align}
where the mean $\mu^{\psi}(t,y)$ is parameterised by some vector $\psi$.

%%%%%%%%%%%%%%%%%%%%%%%%%%%%%%%%%%%%%%%%%%%

\section{Reward Signals}

%%%%%%%%%%%%%%%%%%%%%%%%%%%%%%%%%%%%%%%%%%%%%%%%%%%%%%%%%%%%%%%%
Before presenting our algorithm, we introduce a new component essential to its design.
In general RL problems, we either have access to the reward function or have its (noisy) observations for  learning. A unique feature of the current problem is that the running/instantaneous reward function \eqref{eq:rt} involves the unknown true score $\nabla\log p_{T-t}(\cdot)$. To obtain signals from this unknown running reward (or score), we express a score function value as the ratio of two expectations with respect to the data distribution.
%However, one issue with this RL problem is that we do not have samples from the running reward $r_t$ at any given time $t$, precisely because we do not know the true score $\nabla\log p_{T-t}(\cdot)$.
%To overcome this difficulty, we express the true score function as the ratio of two expectations.

Specifically, note that
\begin{equation*}
p_{t}(\mathbf{x})=\int_{\mathbb{R}^{d}}p_{t|0} (\mathbf{x}|\mathbf{x}_{0})p_{0}(\mathbf{x}_{0})d\mathbf{x}_{0} = \mathbb{E}_{\mathbf{x}_{0} \sim p_0 } [ p_{t|0}(\mathbf{x} | \mathbf{x}_{0})],
\end{equation*}
where we recall that $p_{t|0}(\cdot|\mathbf{x}_{0})$ is the conditional density of  $\mathbf{x}_{t}$ given $\mathbf{x}_{0}$. By the forward process \eqref{SDE:solution}, we know that $p_{t|0}(\cdot|\mathbf{x}_{0})$ is Gaussian with
\begin{equation}\label{eq: conditional_density}
p_{t|0}(\mathbf{x}|\mathbf{x}_{0})
=\frac{1}{\left(2\pi\int_{0}^{t}e^{-2\int_{s}^{t}f(v)dv}(g(s))^{2}ds\right)^{d/2}}\exp\left(-\frac{\Vert\mathbf{x}-e^{-\int_{0}^{t}f(s)ds}\mathbf{x}_{0}\Vert^{2}}{2\int_{0}^{t}e^{-2\int_{s}^{t}f(v)dv}(g(s))^{2}ds}\right).
\end{equation}
It follows that
\begin{align}\label{eq:scoreMC}
\nabla_{\mathbf{x}}\log p_{t}(\mathbf{x}) & = \frac{ \nabla_{\mathbf{x}} p_{t}(\mathbf{x})}{p_{t}(\mathbf{x}) } = \frac{\mathbb{E}_{\mathbf{x}_{0} \sim p_0 } [  \nabla_{\mathbf{x}}  p_{t|0}(\mathbf{x} | \mathbf{x}_{0})]}{ \mathbb{E}_{\mathbf{x}_{0} \sim p_0 } [ p_{t|0}(\mathbf{x} | \mathbf{x}_{0})] }  \nonumber \\
& = \frac{\mathbb{E}_{\mathbf{x}_{0} \sim p_0 } \left[  p_{t|0}(\mathbf{x} | \mathbf{x}_{0})  \cdot  \frac{-( \mathbf{x}-e^{-\int_{0}^{t}f(s)ds}\mathbf{x}_{0}  ) }{ \int_{0}^{t}e^{-2\int_{s}^{t}f(v)dv}(g(s))^{2}ds} \right] }{ \mathbb{E}_{\mathbf{x}_{0} \sim p_0 } [ p_{t|0}(\mathbf{x} | \mathbf{x}_{0})] } \nonumber \\
& = \frac{1} { \int_{0}^{t}e^{-2\int_{s}^{t}f(v)dv}(g(s))^{2}ds} \cdot \left( -  \mathbf{x} +  \frac{\mathbb{E}_{\mathbf{x}_{0} \sim p_0 } \left[  p_{t|0}(\mathbf{x} | \mathbf{x}_{0})  \cdot  {\mathbf{x}_{0}  } \right] }{ \mathbb{E}_{\mathbf{x}_{0} \sim p_0 } [ p_{t|0}(\mathbf{x} | \mathbf{x}_{0})] } \cdot e^{-\int_{0}^{t}f(s)ds}  \right).
\end{align}
Therefore, given $m$ i.i.d. samples $( \mathbf{x}^{i}_{0} )$ from the data distribution $p_0,$
a simple ratio estimator for the true score $\nabla_{\mathbf{x}}\log p_{t}(\mathbf{x}) $ at given $(t,\mathbf{x})$ is
\begin{align}\label{eq:approxi-score}
\widehat{ \nabla_{\mathbf{x}}\log p_{t}(\mathbf{x}) }   = \frac{1} { \int_{0}^{t}e^{-2\int_{s}^{t}f(v)dv}(g(s))^{2}ds} \cdot \left( -  \mathbf{x} +  \frac{ \sum_{i=1}^m  \left[  p_{t|0}(\mathbf{x} | \mathbf{x}^{i}_{0})  \cdot  {\mathbf{x}^{i}_{0}  } \right] }{ \max\{ \sum_{i=1}^m  [ p_{t|0}(\mathbf{x} | \mathbf{x}^i_{0})], m \epsilon \} } \cdot e^{-\int_{0}^{t}f(s)ds}  \right),
\end{align}
where $\epsilon>0$ is some prespecified small constant.
Note that we use $\max\{ \sum_{i=1}^m  [ p_{t|0}(\mathbf{x} | \mathbf{x}^i_{0})], m \epsilon \}$ instead of $\sum_{i=1}^m  [ p_{t|0}(\mathbf{x} | \mathbf{x}^i_{0})]$ in \eqref{eq:approxi-score} to avoid division by extremely small numbers for numerical stability. In view of the running reward function $r$ in \eqref{eq:rt}, if we define
\begin{align}\label{eq:hat-r}
\hat r(t, \mathbf{y}, a) = -g^2(T-t) \cdot {|  \widehat{ \nabla\log p_{T- t}(\mathbf{y}) }  - a|^2},
\end{align}
then a noisy observation of the running reward at time $t$ is
\begin{align}\label{eq:sample-r}
\hat r_t =  \hat r(t, \mathbf{y}_t, a_t).
\end{align}

For fixed $(t,\mathbf{x})$, the simple ratio estimator in \eqref{eq:approxi-score} provides a nonparametric approach to estimate the true score $\nabla_{\mathbf{x}}\log p_{t}(\mathbf{x})$. It is efficient to compute the estimator, because $p_{t|0}(\mathbf{x}|\mathbf{x}_{0})$ has an explicit form. In addition, while this ratio estimator is generally biased, it is asymptotically exact as the number of data samples $m \rightarrow \infty$ by the strong law of large numbers. On the other hand,  the true score function is a function of {\it all} $(t,\mathbf{x})$. Thus, using this method to accurately estimate the score function for every $(t,\mathbf{x})$ is simply impossible. %can be computationally expensive, and this is not what we pursue.
In our study, the estimator  \eqref{eq:approxi-score} allows us to compute the reward signal \eqref{eq:sample-r} as we go, namely we compute it only when a state--action pair $(\mathbf{y}_t, a_t) $ is visited at time $t <T$. On the other hand, it is even unnecessary to use up all the data samples from $p_0$ each time in computing \eqref{eq:approxi-score}, though in theory a larger sample size will generally lead to a lower variance of the running reward signal \eqref{eq:sample-r}. In the training process, suppose that we have $M$ training samples in total, it is more practical to use $m < M$ samples to estimate the score function value and compute the reward signal, so that the computational cost can be significantly  reduced. Indeed, later in an image generation task we will show that even $m=1$ suffices. Details will be discussed in the subsequent experiments.

\begin{remark}
Given the estimator  \eqref{eq:approxi-score} of the score function value $\nabla_{\mathbf{x}}\log p_{t}(\mathbf{x})$, it seems tempting, at least for the pure score-matching problem (without preference), to consider directly running the reverse process \eqref{eq:zt} with the ratio estimator to generate new samples, i.e. to produce samples $\mathbf{y}_T$ via simulating
\begin{equation}\label{eq:sim-ratio}
    d\mathbf{y}_{t}=\left[f(T-t)\mathbf{y}_{t}+(g(T-t))^{2}\nabla\widehat{\log p_{T-t}}(\mathbf{y}_{t})\right]dt
+g(T-t)dW_t, \quad \mathbf{y}_{0}\sim\nu.
\end{equation}
However, this approach essentially replicates those training samples rather than generating {\em new} samples from the target distribution. Indeed, assuming we always use all the $M$ training samples to estimate the score function values $\widehat{ \nabla_{\mathbf{x}}\log p_{t}(\mathbf{x}) }$ in \eqref{eq:approxi-score}, according to the derivation in \eqref{eq:scoreMC} we have
\begin{align*}
    \widehat{ \nabla_{\mathbf{x}}\log p_{t}(\mathbf{x}) }
    &= \frac{1} { \int_{0}^{t}e^{-2\int_{s}^{t}f(v)dv}(g(s))^{2}ds} \cdot \left( -  \mathbf{x} +  \frac{\mathbb{E}_{\mathbf{x}_{0} \sim \hat p_0 } \left[  p_{t|0}(\mathbf{x} | \mathbf{x}_{0})  \cdot  {\mathbf{x}_{0}  } \right] }{ \mathbb{E}_{\mathbf{x}_{0} \sim \hat p_0 } [ p_{t|0}(\mathbf{x} | \mathbf{x}_{0})] } \cdot e^{-\int_{0}^{t}f(s)ds}  \right) \\
    &= \frac{\mathbb{E}_{\mathbf{x}_{0} \sim \hat p_0 } [  \nabla_{\mathbf{x}}  p_{t|0}(\mathbf{x} | \mathbf{x}_{0})]}{ \mathbb{E}_{\mathbf{x}_{0} \sim \hat p_0 } [ p_{t|0}(\mathbf{x} | \mathbf{x}_{0})] }= \nabla_{\mathbf{x}}\log \hat p_{t}(\mathbf{x})
\end{align*}
where $\hat p_0$ is the empirical distribution of the samples $\mathbf{x}_0^1, \cdots \mathbf{x}_0^M$ and $\hat p_t$ is the time marginal distribution of the forward process \eqref{OU:SDE} but starting from $\mathbf{x}_0\sim \hat p_0$. Then running the reverse process \eqref{eq:sim-ratio} essentially generates samples from $\hat p_0$, i.e. replicating samples in the training dataset. It is therefore not a truly generative model for it can not generate new and diverse samples that are different from the training data. In our continuous-time RL approach for training reward-directed diffusion models (which cover the pure score-matching as a special case), we use the ratio estimator only for generating running reward signals, and employ  a neural network to parameterize the policy which is related to the true score; see \eqref{eq:pi-psi} and \eqref{eq:mustar}.
\end{remark}

%%%%%%%%%%%%%%%%%%%%%%%%%%%%%%%%%%%%%%%%%%%%%
\section{q-Learning Algorithm and Convergence}\label{sec:algorithm}
In this section, we present an algorithm to solve the continuous-time RL problem \eqref{eq:Jstar2} and discuss its convergence.

\subsection{q-Learning Algorithm}
 Our algorithm is based on the general q-learning algorithms developed recently in \cite{jia2023q}, which are of the actor--critic type. For the reader's convenience, we first introduce some definitions and theoretical results in \cite{jia2023q} that are important for developing our algorithm.

Following \cite{jia2023q}, we define the so-called optimal $q-$function by
\begin{align*} %\label{eq:opt-q}
	q^*(t, y, a) & =
	\frac{\partial J^*(t, y)  }{\partial t } +   H(t, y, a, J^*_y(t, y),  J^*_{yy}(t, y), J^*(t, y)), \quad (t,y,a) \in [0, T] \times \mathbb{R}^d \times \mathbb{R}^d,
\end{align*}
where $J^*$ is the optimal value function given in \eqref{eq:Jstar2}, and the Hamiltonian function $H$ is defined in \eqref{eq:Hal}. One can readily infer from \eqref{eq:expHJB} that (see Proposition 8 in \citealp{jia2023q})
\begin{align*}
 \int_{\mathbb{R}^d } {\exp \left( \frac{1}{\theta}  q^*(t,y,a) \right) da} = 1, \quad \text{for all $(t,y) \in [0, T] \times \mathbb{R}^d $},
\end{align*}
and  the  optimal stochastic policy $\bpi^* (\cdot| t, y)$ in Proposition \ref{prop:optimal-policy} is simply
\begin{align*}
\bpi^* (a| t, y) = \exp \left( \frac{1}{\theta}  q^*(t, y,a) \right).
\end{align*}

Note that the exploratory (or ``averaged") dynamic in \eqref{eq: exp_state_SDE} is introduced mainly for the purpose of theoretical analysis. Its solutions are not observable (i.e. not data), and hence can not be used in the actual implementation of RL algorithms. We next introduce the discretely sampled state processes, which can be observed based on the agent interactions with the environment.
      Given an admissible feedback policy $\bpi$ and  $(t,y) \in [0,T) \times \mathbb{R}^d,$ consider a time grid $\mathbb{S}= \{t = t_0 < t_1 < \ldots < t_K = T\}$ of $[t, T]$. In algorithmic implementations, we sample actions from $\bpi$ only at the grid points in $\mathbb{S}$.
The corresponding state process, which is referred to as the \emph{discretely sampled state process}, satisfies the following SDE: for all $i = 0, . . . , K - 1$ and all $s \in [t_{i}, t_{i+1})$,
\begin{align} \label{eq:samplestate}
d \mathbf{y}^{\bpi, \mathbb{S}}_s
= [f(T-s) \mathbf{y}^{\bpi, \mathbb{S}}_s + (g(T-s))^{2} {\bf a}^{\bpi}_{t_i} ] ds + g(T-s)  dW_s,
\end{align}
where ${\bf a}^{\bpi}_{t_i}= \phi (t_i, \mathbf{y}^{\bpi, \mathbb{S}}_{t_i}, U_{i+1})$. Here, the function $\phi: (t, y, u) \in [0,T] \times \mathbb{R}^d \times [0,1]^{d} \rightarrow  \phi(t,y,u ) \in \mathbb{R}^d$ is given such that $\phi(t,y, U) \sim \bpi(\cdot|t,y)$ where $U$ is a uniform random vector on $[0,1]^{d}$. The random vectors $(U_i)$ are i.i.d $d$-dimensional uniform random vectors, and they are independent of the Brownian motion $W$. We also define the action process $a_{s}^{\bpi, \mathbb{S}} =  {\bf a}^{\bpi}_{t_i}$ for $s \in [t_{i}, t_{i+1})$. Hence, the action is a constant over each subinterval $[t_{i}, t_{i+1})$. Finally, we denote by $\mathcal{F}^{\mathbb{S}}$ the (right-continuous) filtration generated by the Brownian motion $W$ and the random vectors $(U_i)$.

We next state the martingale condition that characterizes the optimal value function $J^*$ and the optimal q-function; see Theorem 9 in \citep{jia2025erratum}.  This result is essential because it provides the theoretical foundation for designing our q-learning algorithm.

\begin{proposition}\label{thm:opt-mart}
Let a function $\hat J^* \in C^{1,2}([0,T)\times\mathbb{R}_{+})\cap C([0,T]\times\mathbb{R}_{+})$ and a continuous function $\hat q^*: [0, T] \times \mathbb{R}^d \times \mathbb{R}^d \rightarrow \mathbb{R}$ be given satisfying
\begin{align} \label{eq:boundary}
	\hat J^*(T, y) = h (y), \quad \int_{\mathbb{R}^d } {\exp \left( \frac{1}{\theta} \hat q^*(t,y,a) \right) da} = 1, \quad \text{for all $(t,y) \in [0, T] \times \mathbb{R}^d $}.
\end{align}
Assume that $\hat J^*$ and $\hat J^*_y$ both have polynomial growth.
Then
\begin{enumerate}
	\item [(i)] If $\hat J^*$ and $\hat q^*$ are respectively the optimal value function and the optimal $q$-function,  then for any $\boldsymbol{\pi} \in \boldsymbol{\Pi}$ , for all
	$(t,y) \in [0, T] \times \mathbb{R}^d $ and any grid $\mathbb{S}$ of $[t,T]$, the following process
	\begin{align} \label{eq:mart-opt}
		 \hat J^*(s, \mathbf{y}_s^{ \boldsymbol{\pi}, \mathbb{S}}) +    \int_t^s  [ r(\tau, \mathbf{y}_{\tau}^{\boldsymbol{\pi}, \mathbb{S}}, a_{\tau}^{\boldsymbol{\pi}, \mathbb{S}}) - \hat q^* (\tau, \mathbf{y}_{\tau}^{\boldsymbol{\pi} , \mathbb{S}}, a_{\tau}^{\boldsymbol{\pi}, \mathbb{S}}) ] d\tau
	\end{align}
	is an $\mathcal{F}^{\mathbb{S}}$-martingale, where $\mathbf{y}^{\boldsymbol{\pi}, \mathbb{S}} = \{ \mathbf{y}^{\boldsymbol{\pi}, \mathbb{S}}_s: t \le s \le T\}$ satisfies \eqref{eq:samplestate}  with $\mathbf{y}^{\boldsymbol{\pi}, \mathbb{S}}_t =y$.

	\item [(ii)] If there exists one $\boldsymbol{\pi} \in \boldsymbol{\Pi}$ such that for all $(t,y)$ and any grid $\mathbb{S}$ of $[t,T]$, the process \eqref{eq:mart-opt} is an $\mathcal{F}^{\mathbb{S}}$-martingale where $\mathbf{y}^{\boldsymbol{\pi}, \mathbb{S}}_t=y$, then $\hat J^*$ and $\hat q^*$ are respectively the optimal value function and the optimal $q$-function.

\end{enumerate}

\end{proposition}

We are now ready to develop and state the algorithm to solve our continuous-time RL problem \eqref{eq:Jstar2}.
Consider the parameterized Gaussian policies $\bpi^{\psi}(\cdot|t,y)$ in \eqref{eq:pi-psi}.
%\begin{align*}
%\bpi^{\psi}(\cdot|t,y)   \sim  \mathcal{N}(\mu^{\psi}(t,y), \frac{\theta}{2 g^2(T-t)}  \cdot I_{d} ) \quad \text{for all $(t, y)$.}%\left(\cdot \mid  \right)
%\end{align*}
It is useful to note that
\begin{align} \label{eq:pi-psi_q}
\bpi^{\psi} (a| t, y) = \exp \left( \frac{1}{\theta} q^{\psi}(t, y, a) \right),
\end{align}
where
 \begin{align}\label{eq:q-psi}
q^{\psi}(t, y, a) = - g^2(T-t) \cdot |a - \mu^{\psi}(t,y)|^2 - \frac{\theta d}{2} \log \left(\frac{ \pi \theta }{g^2(T-t)} \right).
%T [\Sigma^{\psi}(t,u)]^{-1} [a - \mu^{\psi}(t,u)] + \frac{\theta}{2} \log \det   [\Sigma^{\psi}(t,u)]^{-1} - \frac{m \theta}{2} \log 2 \pi.
\end{align}
This function $q^{\psi}$ is a resulting parameterization  of the optimal $q$-function $q^*$  which satisfies the equation $\int_{\mathbb{R}^d} \exp \left( \frac{1}{\theta} q^{\psi}(t, y, a) \right) da =1$.
We also choose the value function approximator $J^{\Theta}$ parametrized by $\Theta$ for the optimal value function $J^*$.  We aim to learn the optimal value function and q-function simultaneously by updating $(\Theta,\psi)$,  based on the martingale characterization in Proposition~\ref{thm:opt-mart}. 
In particular, we follow \cite{jia2023q} and use the so-called martingale orthogonality conditions. These conditions state that a diffusion process $M$ is a (square-integrable) martingale with respect to a filtration $\mathcal{F}$ if and only if $\mathbb{E} \int_0^T H_t dM_t =0$ for any process $H$ (called a test function) on $[0,T]$ that is progressively measurable (with respect to $\mathcal{F}$) with $\mathbb{E}[\int_0^T |H_t|^2 d \langle M \rangle_t] < \infty,$ where $\langle M \rangle_t$ denotes the quadratic variation of the martingale $M$. See \cite[Section 4.2]{JZ21}.

Because the optimal value function $J^*$ and q-function $q^*$ are both infinite-dimensional, theoretically one should vary all possible test functions to obtain infinitely many equations from the martingale orthogonality conditions in order to determine $(J^*, q^*)$. This is naturally infeasible for computer implementation. Numerically we need only to determine the parameters $(\Theta, \psi) \in \mathbb{R}^k$, for some $k$, of  the function approximators $(J^{\Theta}, q^{\psi})$. This calls for $k$ test functions. We choose the following test functions $\xi_t$ and $\zeta_t$ for our problem:
\begin{align}\label{eq: q_test_func}
\xi_t = \frac{\partial J^{\Theta}}{\partial \Theta}(t, \mathbf{y}_t^{\bm\pi^{\psi}}),  \quad \zeta_t = \frac{\partial q^{\psi}}{\partial \psi}(t, \mathbf{y}_t^{\bm\pi^{\psi}},a_t^{\bm\pi^{\psi}}) = \frac{\partial }{\partial \psi} \log \pi^{\psi}(t, \mathbf{y}_t^{\bm\pi^{\psi}},a_t^{\bm\pi^{\psi}}),
\end{align}
where we use the fact that $q^{\psi}(t, y, a) = \theta \log \bpi^{\psi} (a| t, y)$.
The above choice of $\xi_t$ is motivated by the TD(0) method  in discrete-time RL (e.g. \citealp{bhandari2021finite}; see \citealp[Section 4.2]{JZ21} for a detailed discussion), whereas that of $\zeta_t$ is inspired by the policy gradient method \citep{schulman2015high, JZ22}.

In our implementation, we execute stochastic policies on a uniform time grid with step size $\Delta t$; that is, the grid $\mathbb{S}= \{0= t_0 < t_1 < \ldots < t_K = T\}$, where $t_{i+1} - t_i = \Delta t$ for all $i$. For notational simplicity, in \eqref{eq: q_test_func} and the discussions below  we denote the discretely sampled state process $\mathbf{y}_t^{\bm\pi^{\psi}, \mathbb{S}}$ by $\mathbf{y}_t^{\bm\pi^{\psi}}$. Likewise, the (piecewise constant) action process is denoted by $a_t^{\bm\pi^{\psi}}$.

The martingale characterization in Proposition~\ref{thm:opt-mart} together with the martingale orthogonality conditions lead to the following system of equations in $(\Theta, \psi)$:
\begin{align}
&\mathbb{E} \left[  \int_0^T  \xi_t     \left( dJ^{\Theta} (t, \mathbf{y}_t^{\bm\pi^{\psi}}) +  r(t, \mathbf{y}_t^{\bpi^{\psi}}, a_t^{\bpi^{\psi}}) dt - q^{\psi}(t, \mathbf{y}_{t}^{\bpi^{\psi}}, a_t^{\bpi^{\psi}}) dt \right)     \right]=0,  \label{eq:SA1}\\
& \mathbb{E} \left[  \int_0^T  \zeta_t     \left( dJ^{\Theta} (t, \mathbf{y}_t^{\bm\pi^{\psi}}) +  r(t, \mathbf{y}_t^{\bpi^{\psi}}, a_t^{\bpi^{\psi}}) dt - q^{\psi}(t, \mathbf{y}_{t}^{\bpi^{\psi}}, a_t^{\bpi^{\psi}}) dt \right)     \right] =0. \label{eq:SA2}
\end{align}
%Note that if the parameters $(\Theta, \psi) \in \mathbb{R}^k$ for some $k$, then in principle we need at least $k$ equations in order to fully determine $(\Theta, \psi)$.
Equations~\eqref{eq:SA1}-\eqref{eq:SA2} have $k$ equations used to determine $(\Theta, \psi) \in \mathbb{R}^k$. %, assuming the existence of solutions to these equations.
We solve these equations using stochastic approximation (SA) and replace the true (unknown) running reward function $r$ by its noisy estimate $\hat r$ in \eqref{eq:hat-r} to update $(\Theta, \psi)$:
\begin{align}
& \Theta \leftarrow \Theta + \alpha_{\Theta}    \int_0^T  \xi_t     \left( dJ^{\Theta} (t, \mathbf{y}_t^{\bm\pi^{\psi}}) +  \hat r(t, \mathbf{y}_t^{\bpi^{\psi}}, a_t^{\bpi^{\psi}}) dt - q^{\psi}(t, \mathbf{y}_{t}^{\bpi^{\psi}}, a_t^{\bpi^{\psi}}) dt \right)     ,  \label{eq:SA-update1}\\
& \psi \leftarrow \psi + \alpha_{\psi}  \int_0^T  \zeta_t     \left( dJ^{\Theta} (t, \mathbf{y}_t^{\bm\pi^{\psi}}) +  \hat r(t, \mathbf{y}_t^{\bpi^{\psi}}, a_t^{\bpi^{\psi}}) dt - q^{\psi}(t, \mathbf{y}_{t}^{\bpi^{\psi}}, a_t^{\bpi^{\psi}}) dt \right)    . \label{eq:SA-update2}
\end{align}
%Given our choice of the test functions in \eqref{eq: q_test_func}, the update rule of $\Theta$ resembles the TD(0) method with function approximation in discrete-time RL; see e.g. \cite{bhandari2021finite}), and the update rule of $\psi$ is similar to the policy gradient method; see e.g. \citep{schulman2015high, JZ22}.
The above updating rule leads to an actor--critic type algorithm based on the temporal difference $dJ^{\Theta} (t, \mathbf{y}_t^{\bm\pi^{\psi}})$.  Algorithm~\ref{algo:offline episodic} gives the pseudocode. We also remark that it is possible to consider other test functions in \eqref{eq:SA1}--\eqref{eq:SA2}, which will lead to different algorithms; see \cite{JZ21,jia2023q} for further discussions.

\begin{algorithm}[hbtp]
\caption{q-Learning Algorithm (SDE-based unconditional generation)}
\textbf{Inputs}: $M$ samples from data distribution, number of samples $m$ that used to estimate the score function, horizon $T$, time step $\Delta t$, number of episodes $N$, number of mesh grids $K = T/\Delta t$, initial learning rates $\alpha_{\Theta},\alpha_{\psi}$ and a learning rate schedule function $l(\cdot)$ (a function of the number of episodes), functional forms of parameterized value function $J^{\Theta}(\cdot,\cdot)$ and  $\mu^{\psi}(\cdot,\cdot)$, temperature parameter $\theta$, functions $f, g$ in \eqref{OU:SDE}, and $\epsilon$ in \eqref{eq:approxi-score}.

\textbf{Required program}: environment simulator $(y', \hat r) = \textit{Environment}_{\Delta t}(t,y,a, m)$ that takes current time--state pair $(t,y)$ and action $a$ as inputs and generates state $y'$ (by a numerical solver of SDE \eqref{eq:samplestate}) at time $t+\Delta t$ and sample instantaneous reward $\hat r$ (by randomly selecting $m$ samples from the dataset and implementing \eqref{eq:sample-r}) at time $t$ as outputs. Policy $\bpi^{\psi}(\cdot|t, y)$ in \eqref{eq:pi-psi}, and q-function $q^{\psi}(t, y, a)$ in \eqref{eq:q-psi}.

\textbf{Learning procedure}:
\begin{algorithmic}
\STATE Initialize $\Theta,\psi$.
\FOR{episode $j=1$ \TO $N$} \STATE{Initialize $k = 0$. Sample initial state $y_0 \sim\nu$ and store $y_{t_k} \leftarrow  y_0$.

\WHILE{$k < K$} \STATE{
Generate action $a_{t_k}\sim \bm{\pi}^{\psi}(\cdot|t_k,y_{t_k})$.

Apply $a_{t_k}$ to environment simulator $(y, \hat r) = Environment_{\Delta t}(t_k, y_{t_k}, a_{t_k}, m)$, and observe new state $y$ and reward $\hat r$ as outputs. Store $y_{t_{k+1}} \leftarrow y$ and $r_{t_k} \leftarrow \hat r$.

Update $k \leftarrow k + 1$.
}
\ENDWHILE

For every $i = 0,1,\cdots,K-1$, compute and store test functions
\begin{align*}
\xi_{t_i}= \frac{\partial J^{\Theta}}{\partial \Theta}(t_{i},y_{t_{i}}),  \quad \zeta_{t_i} = \frac{\partial q^{\psi}}{\partial \psi}(t_{i}, y_{t_{i}},a_{t_i}).
\end{align*}	

Compute  % (note when $i=K-1$, we use $J^{\Theta}(t_{K}, u_{t_{K}}) =\beta h(u_{t_{K}}) = \beta h(u_T)$ instead of the value computed from NN)
\[ \Delta \Theta = \sum_{i=0}^{K-1} \xi_{t_i} \big[ J^{\Theta}(t_{i+1}, y_{t_{i+1}}) - J^{\Theta}(t_{i},y_{t_{i}}) + r_{t_i}\Delta t -q^{\psi}(t_{i}, y_{t_{i}},a_{t_i})\Delta t  \big], \]
\[
\Delta \psi =   \sum_{i=0}^{K-1}\zeta_{t_i}\big[ J^{\Theta}(t_{i+1},y_{t_{i+1}}) - J^{\Theta}(t_{i}, y_{t_{i}}) + r_{t_i}\Delta t - q^{\psi}(t_{i},y_{t_{i}},a_{t_i})\Delta t \big] .
\]

Update $\Theta$ and $\psi$ by
\begin{align}
 &\Theta \leftarrow \Theta + l(j)\alpha_{\Theta} \Delta \Theta , \label{Theta-update}\\
& \psi \leftarrow \psi + l(j)\alpha_{\psi} \Delta \psi .  \label{Delta-update}
\end{align}

}
\ENDFOR
\end{algorithmic}
\label{algo:offline episodic}
\end{algorithm}

\begin{remark}\label{adam}

In Algorithm~\ref{algo:offline episodic}, the update rule for $(\Theta, \psi)$ in \eqref{Theta-update}-\eqref{Delta-update} is essentially  the stochastic gradient descent (SGD) type method. To see this, define
	\begin{align*}
		 G(t_i; \Theta, \psi) := \hat J(t_{i+1},y_{t_{i+1}}) - J^{\Theta}(t_{i}, y_{t_{i}}) + r_{t_i}\Delta t - q^{\psi}(t_{i},y_{t_{i}},a_{t_i})\Delta t,
	\end{align*}
	where $\hat J(t_{i+1},y_{t_{i+1}}) := J^\Theta(t_{i+1},y_{t_{i+1}})$,  $i=0, ..., K - 1$, which however are treated  as constants free of parameter $\Theta$. Then
	\begin{align*}
		\Delta \Theta = - \frac{1}{2}\sum_{i=0}^{K - 1} \frac{\partial}{\partial\Theta}\left[ G(t_i; \Theta, \psi)\right]^2, \quad \Delta \psi = - \frac{1}{2\Delta t}\sum_{i=0}^{K - 1} \frac{\partial}{\partial\psi}\left[ G(t_i; \Theta, \psi)\right]^2.
	\end{align*}
	Therefore, the update of $(\Theta, \psi)$ in \eqref{Theta-update}-\eqref{Delta-update} can be considered as a gradient descent method for minimizing the following mean-square loss function:
	\begin{align}\label{eq:loss}
		L(\Theta, \psi) := \sum_{i=0}^{K - 1} \left[ G(t_i; \Theta, \psi)\right]^2.
	\end{align}
As a result, instead of the SA in \eqref{Theta-update}-\eqref{Delta-update}, we can also optimize the loss function \eqref{eq:loss} by applying other optimization methods such as the Adam optimizer \citep{kingma2014adam} to update these parameters. In our experiments, we use Adam because it converges much faster than SA/SGD. In addition, to further accelerate the training process, the above loss function could be estimated more accurately through the empirical mean of a batch of losses under the same $(\Theta, \psi)$. Specifically, we collect a batch of $B$ trajectories simultaneously, calculate the above loss for each of the $B$ sample paths, denoted by $L^{(b)}(\Theta, \psi), b=1,..., B$, and then use the following batch loss for optimization:
\begin{equation*}
    L_{\text{batch}}(\Theta, \psi):= \frac{1}{B}\sum_{b=1}^BL^{(b)}(\Theta, \psi).
\end{equation*}
\end{remark}

\begin{remark}
In the classical setting of training diffusion models using score matching, the training data (i.e. samples from $p_0$) are used in Monte Carlo approximation of the expectation in the (denoising) score matching objective \eqref{eq:score-matching2}.
% \begin{align}
% \min_{\theta} \mathbb{E}_{t \sim U[0, T]}  \left[ \lambda(t) \mathbb{E}_{\mathbf{x}_{0}}  \mathbb{E}_{\mathbf{x}_{t} | \mathbf{x}_{0}} \left\Vert s_{\theta}(\mathbf{x}_t,t)-\nabla_{\mathbf{x}_{t}}\log p_{t|0}(\mathbf{x}_{t} | \mathbf{x}_{0})\right\Vert^2   \right].
% \end{align}
In our RL formulation, however, the training data are used for approximating the score {\it values} via the ratio estimators in \eqref{eq:approxi-score} in order to obtain noisy samples of the running reward. This again highlights one of the key differences between the two approaches.

\end{remark}

\subsection{Convergence}\label{sec:convergence}
In this section, we discuss the convergence  of the q-Learning algorithm. We focus on the analysis of the SA scheme in \eqref{eq:SA-update1}-\eqref{eq:SA-update2} for updating $(\Theta, \psi)$. In particular,
we do not consider errors of
time discretization of the integrals in actual implementations of the $q-$learning algorithm,  Algorithm~\ref{algo:offline episodic}, because including
these errors would only complicate the analysis and cloud the key insights from $q-$learning. In implementation, such errors can be controlled by appropriately choosing the time steps; see a recent paper \cite{jia2025accuracy} for a detailed study.

To analyze the scheme \eqref{eq:SA-update1}-\eqref{eq:SA-update2} for updating $(\Theta, \psi)$, we rewrite it as follows: for each episode $n,$
\begin{align}\label{eq: qL_SA}
    \left(\Theta_{n + 1}, \psi_{n + 1}\right) = \left(\Theta_{n}, \psi_{n}\right) + \left(\alpha_{\Theta}(n + 1), \alpha_{\psi}(n + 1)\right)\cdot\mathcal{H}\bigg((\Theta_{n}, \psi_{n}), \underbrace{(\mathbf{y}_t^{\bpi^{\psi_n}}, a_t^{\bpi^{\psi_n}})_{t \in [0, T]}}_{\Xi_{n + 1}}, \mathcal{Z}\bigg).
\end{align}
Here, we have %$n$ denotes the episode/iteration number, and
\begin{align} \label{eq:H-SA}
    \mathcal{H}\big((\Theta, \psi), \Xi, \mathcal{Z} \big) := \left(\int_{0}^T\xi_t d \mathcal{G}(t; \Theta, \psi), \int_{0}^T\zeta_t d \mathcal{G}(t; \Theta, \psi)\right),
\end{align}
with $\xi_t$ and $\zeta_t$ being the test functions in \eqref{eq: q_test_func}, $\Xi =(\mathbf{y}_t^{\bpi^{\psi}}, a_t^{\bpi^{\psi}})_{t\in[0, T]}$,
\begin{align*}
    d \mathcal{G}(t; \Theta, \psi):=  dJ^{\Theta} (t, \mathbf{y}_t^{\bm\pi^{\psi}}) +  \hat r(t, \mathbf{y}_t^{\bpi^{\psi}}, a_t^{\bpi^{\psi}}) dt - q^{\psi}(t, \mathbf{y}_{t}^{\bpi^{\psi}}, a_t^{\bpi^{\psi}}) dt,
\end{align*}
and $\mathcal{Z} = (\mathcal{Z}_t)_{t \in \mathbb{S}}$ where $\mathcal{Z}_t$ is a mini batch of $m$ data points independently sampled from the whole dataset (of size $M$)  at each grid point $t$ in order to estimate the score function value there, {and $\mathcal{Z}$ is independent of $\Xi_{n + 1}$ for all $n$.}
Note that we allow time-varying learning rates  $\left(\alpha_{\Theta}(n + 1), \alpha_{\psi}(n + 1)\right)$, which helps in the convergence analysis of SA schemes \citep{kushner2003stochastic}.
The update scheme \eqref{eq: qL_SA} is an instance of the Robbins--Monro algorithm (see e.g. p.343 of \citealp{benveniste2012adaptive}). This is because in each episode $n$, the initial state $\mathbf{y}_0$ is sampled independently from a Gaussian distribution $\nu$ in \eqref{eq:hatp}, which implies that
the conditional distribution of $(\Xi_{n+1}, \mathcal{Z})$, knowing the past, %({\color{red} i.e. $\sigma$-field generated by the variables $\Xi_n, \Xi_{n - 1}, \cdots \psi_n, \psi_{n - 1}\cdots$}),
depends only on $\psi_n$.
Define
\begin{align*}
    e(\Theta, \psi):= \mathbb{E} \big[\mathcal{H}((\Theta, \psi), \Xi,  \mathcal{Z}) \big| \Theta, \psi \big],
\end{align*}
where the expectation is taken with respect to the distribution of $\mathcal{Z} = (\mathcal{Z}_t)_{t \in \mathbb{S}}$ and the path measure induced by $\Xi=(\mathbf{y}_t^{\bpi^{\psi}}, a_t^{\bpi^{\psi}})_{t\in[0, T]}$ when $\Theta$ and $\psi$ are given.

We can now state a convergence result for \eqref{eq: qL_SA}, which follows from Chapter 5.3 in Part II of \cite{benveniste2012adaptive}.

% \begin{assumption}\label{assumption: convergence_cond}
% ~\\
% \begin{enumerate}

%     \item There exists $C > 0$ such that $\mathbb{E} \big[|\mathcal{H}((\Theta_n, \psi_n, ), U_{n + 1})|^2\big| \Theta_n, \psi_n\big] \le C(1 + |\Theta_n^2| + |\psi_n|^2)$.

%     \item  There exists $(\Theta^*, \psi^*)$ such that $\sup_{\epsilon \le |(\Theta - \Theta^*, \psi - \psi^*)| \le 1/\epsilon }(\Theta - \Theta^*, \psi - \psi^*)^T\cdot e (\Theta, \psi) <0$ for all $\epsilon>0$.

%     \item The stepsizes satisfy $\sum_{n} \alpha(n) = \infty$ and  $\sum_{n} \alpha(n)^2 < \infty$.
% \end{enumerate}
% \end{assumption}

% The following results are from the Theorem 1 of the Section 5.1 in \cite{benveniste2012adaptive}
\begin{proposition}\label{prop:SA-convergence}
   Assume the following conditions hold:
   \begin{enumerate}

    \item There exist a twice continuously differentiable Lyapunov function $V$ with a uniformly bounded second derivative and  $(\Theta^*, \psi^*)$ satisfying
    \begin{align*}
    \sup_{\epsilon \le |(\Theta - \Theta^*, \psi - \psi^*)| \le 1/\epsilon }\nabla V(\Theta, \psi) \cdot e (\Theta, \psi) <0, \quad \text{for all $\epsilon>0$. }
    \end{align*}
    Moreover, $\mathbb{E} \big[|\mathcal{H}((\Theta, \psi, ), \Xi, \mathcal{Z})|^2\big| \Theta, \psi\big] \le C(1 + V(\Theta, \psi))$ for some $C>0$.

    % \item  There exists $(\Theta^*, \psi^*)$ such that $\sup_{\epsilon \le |(\Theta - \Theta^*, \psi - \psi^*)| \le 1/\epsilon }(\nabla V(\Theta, \psi))^T\cdot e (\Theta, \psi) <0$ for all $\epsilon>0$.

    \item The learning rates satisfy $\sum_{n} \alpha_{\gamma}(n) = \infty$ and  $\sum_{n} \alpha_{\gamma}(n)^2 < \infty$ for $\gamma= \Theta, \psi$.
\end{enumerate}
Then the sequence $(\Theta_n, \psi_n)$ in \eqref{eq: qL_SA} converges to $(\Theta^*, \psi^*)$ almost surely.
\end{proposition}

The second assumption in Proposition~\ref{prop:SA-convergence} is standard that guides the choices of learning rates.
Verifying the first assumption, however, is nontrivial. The difficulty arises mainly from the fact that our setting involves continuous time and continuous state/action spaces, and the Lyapunov function depends on the specific choice of the function approximations (neural networks) of  the value function
and the $q$-function (i.e. $J^{\Theta}$ and $q^{\psi}$).
%we need function approximations for both the value function and the $q$-function (which are parameterized using neural networks in our experiments). Furthermore, the function $\mathcal{H}$ in \eqref{eq:H-SA} includes time integrals and stochastic integrals, adding additional complexity to the analysis.
As such, a  rigorous verification of this assumption (i.e. finding the desired Lyapunov function) is beyond the scope of this work and is left for future investigation.

\section{Experiments for Two Toy Examples}\label{sec:experiment}
In this section, we implement Algorithm \ref{algo:offline episodic} and conduct ``proof-of-concept" experiments for two toy examples with synthetic training data, one involving a one-dimensional (1D) Gaussian mixture distribution and the other a two-dimensional (2D) Swiss rolls dataset. We further show the effectiveness of our RL approach by comparing its performance against that of two state-of-the-art RL fine-tuning methods.

\subsection{General setup}\label{sec: 1d_sde_setup}
We first provide some details on the implementation, including the environment (i.e. SDE model \eqref{eq:samplestate}) and its simulator, as well as the structure of the neural networks used for the parameterized actor $\pi^\psi$ and critic $J^\Theta$.
\begin{itemize}
	\item \textbf{SDE/Environment simulator.}

	The implementation of Algorithm \ref{algo:offline episodic} requires an environment simulator describing the (controlled) denoising process. This corresponds to a numerical approximation of the controlled SDE \eqref{eq:samplestate}.

	For  \eqref{eq:samplestate}, we take
	\begin{align}\label{eq:SDE_fg}
		f(t)\equiv 1,\quad g(t)\equiv \sqrt{2},\quad 0\le t\le T,
	\end{align}
	for both the 1D and 2D examples, leading to the prior distribution $\nu$ in \eqref{eq:hatp} as
	\begin{align*}
		\nu:=\mathcal{N}\left(\mathbf{0}, \left(1 - e^{-2T}\right)\cdot I_d\right),\quad d=1, 2.
	\end{align*}
 Other forms of $f$ and $g$ can also be considered, but we choose \eqref{eq:SDE_fg} because it is simple and such a choice already yields satisfactory numerical results as we will see below.
	We numerically solve the controlled SDE \eqref{eq:samplestate} via the standard Euler--Maruyama discretization, i.e., for some small $\Delta t > 0$ and $t < T$, $\mathbf{y}^{\bpi}_{t + \Delta t} - \mathbf{y}^{\bpi}_{t} \approx \left[f(T- t) \mathbf{y}^{\bpi}_{t} +(g(T-t))^{2} a^{\bpi}_t \right] \cdot\Delta t + g(T-t)\sqrt{\Delta t} \cdot \xi$, where $\xi\sim \mathcal{N}(\mathbf{0}, I_d)$. The SDE (or environment) simulator under a policy $\bpi$ is given by (with a slight abuse of notation for $\mathbf{y}^{\bpi}_{t}$)
	\begin{align} \label{eq:simulator}
		\mathbf{y}^{\bpi}_{t + \Delta t} - \mathbf{y}^{\bpi}_{t}= \left[ \mathbf{y}^{\bpi}_{t} +2a^{\bpi}_t \right] \Delta t+ \sqrt{2\Delta t} \cdot \xi,  \quad \mathbf{y}^{\bpi}_{0} \sim \nu.
	\end{align}

	\item  \textbf{Neural network approximators.}

	We use two neural networks (NNs) for the policy $\mu^\psi$ (i.e., the mean of the Gaussian policy in \eqref{eq:pi-psi}) and value function $J^\Theta$, respectively. Because we consider low-dimensional examples, the structure of those two NNs are similar, except that the output dimensions are different; see Table \ref{table:NN_setting} below for details. As noted in Remark~\ref{adam}, we choose the Adam optimizer  to optimize these two NNs, and we will specify the hyperparameters values for the training processes later.

	\begin{table}[H]
		\centering
		\begin{tabular}{c | c c c}
			\hline Layer ID & Input Dimension & Output Dimension & Activation Function\\
			\hline 1 (Input) & $d + 1$ &  64 & Tanh\\
			2 (Hidden) & 64 & 64 & ReLU\\				
			3 (Output) & 64 & $\mu^\psi: d,\quad J^\Theta: 1$\\ \hline
		\end{tabular}
		\captionsetup{justification=centering}
		\caption{Neural networks setting}
		\label{table:NN_setting}
	\end{table}

\end{itemize}

\subsection{1D example - mixed Gaussian}
We first consider an example where the RL agent is provided with $m=300$ samples drawn from a mixture of two 1D Gaussians: % specifically $d=1$ and
\begin{align*}
p_0 = \frac{1}{2} \mathcal{N}(-3, 1) + \frac{1}{2} \mathcal{N}(3, 1).
\end{align*}
We consider a reward function
\begin{align}\label{eq:h-1}
	h(y)= - (y - 6)^2,
\end{align}
 indicating the agent's preference for the generated samples to be close to $6$ (which is located at the right tail of $p_0$).

\begin{figure}[h]
    \centering
    %\captionsetup{justification=raggedright, singlelinecheck=false}
    \includegraphics[scale=0.45]{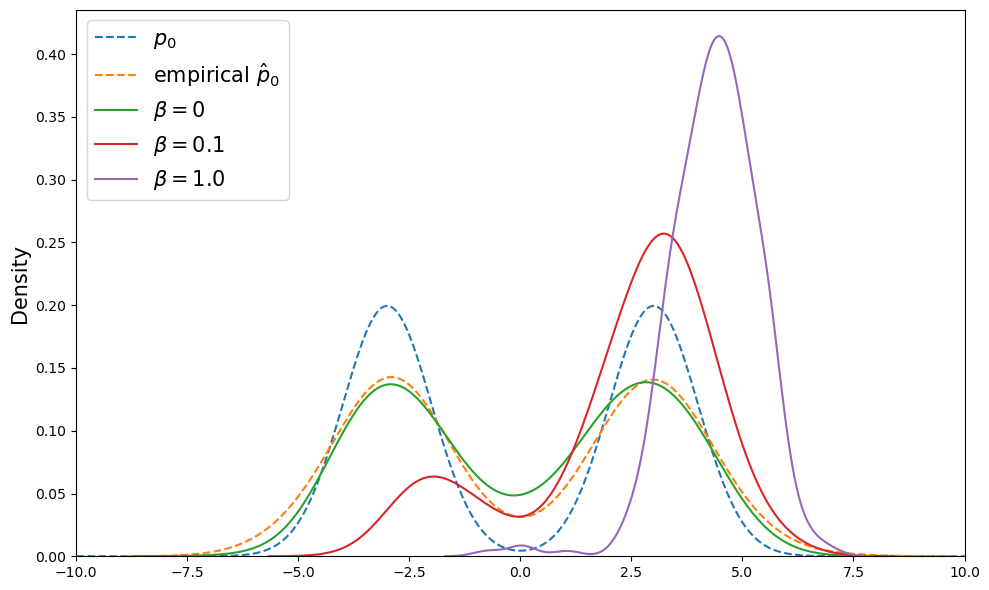}
    \caption{Learned probability densities of generated samples under different $\beta$'s in 1D example}
    \label{fig:example_1D_300}
\end{figure}

In our experiment, we consider different $\beta$'s and implement Algorithm \ref{algo:offline episodic} under hyperparameters shown in Table \ref{table:param_toy}. For each trained model with a particular value of $\beta$, we generate 300 samples (i.e. 300 terminal states $y_T$ from the SDE simulator \eqref{eq:simulator}). The resulting probability density functions (PDFs), computed via kernel density estimation using the Python seaborn library, are plotted in Figure \ref{fig:example_1D_300}.

We have the following observations from Figure \ref{fig:example_1D_300}.
First, when $\beta=0$, our continuous-time RL formulation \eqref{eq:CT-RL} essentially reduces to score matching, and our RL algorithm indeed
 generates samples with a PDF that matches closely with the empirical PDF $\hat p_0$ that are computed using the 300 samples from the Gaussian mixture distribution. It is worthwhile to note that the empirical PDF still deviates away from the true density $p_0$ due to the small number of samples ($m=300$) taken, and the two densities will become closer with a larger sample size.
 Second, when $\beta$ is 0.1, our RL algorithm generates more samples that are closer to the right mode of the Gaussian mixture distribution, due to the specific reward function \eqref{eq:h-1}. Finally, when $\beta$ is large, say 1, the reward term $\beta \cdot h$ dominates the score matching term in \eqref{eq:CT-RL}. In this case, the  algorithm generates samples that are closer to 6, and the generated distribution is significantly different from the original data distribution.

\begin{table}[h]
    \captionsetup{justification=centering}
    \begin{minipage}{0.5\linewidth}
	\centering
        %\captionsetup{justification=center}
	\begin{tabular}{l | r}
		\hline Inputs/Hyperparameters & Setting\\
		\hline Sample Size $m$ & 300\\
		Terminal Time $T$ &  5\\
		Time Step $\Delta t$ & 0.25\\
            Number of Episodes & 50,000\\
            Batch Size $B$ & 300\\
		Learning Rate $\alpha_\psi $ & 0.001 \\
		Learning Rate $\alpha_\Theta $ & 0.001 \\
		Scheduler $l(episode)$ & 1\\
            Temperature $\theta$ & 5\\
            Lower Bound $\epsilon$ & $10^{-20}$\\ \hline
	\end{tabular}
	\caption{Hyperparameters}
	\label{table:param_toy}
	\end{minipage}%
    \begin{minipage}{0.5\linewidth}
	\centering
	%c\captionsetup{justification=raggedright, singlelinecheck=false}
	\includegraphics[scale=0.48]{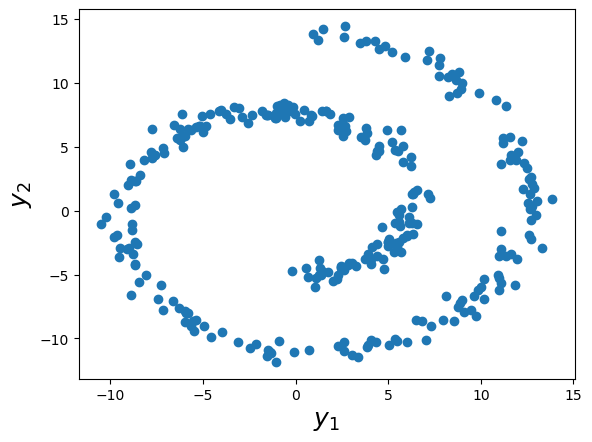}
	\captionof{figure}{300 samples from Swiss roll data}
	\label{fig:example_2D_x0}
    \end{minipage}
\end{table}

Before proceeding, we briefly discuss the choice of hyperparameters presented in Table \ref{table:param_toy}. Key hyperparameters that impact the RL training process include $T, \Delta t, \alpha_\psi, \alpha_\Theta,$ and $\theta$. We employ a grid search to determine the optimal combination of these hyperparameters. The search space included:  $T \in \{0.1, 0.5, 1, 2, 5, 10\}$, $\Delta t\in\{0.01, 0.025, 0.05, 0.1, 0.25, 0.5\}$, $\alpha_\psi = \alpha_\Theta \in \{10^{-1}, 10^{-2}, 10^{-3}, 10^{-4}, 10^{-5}\}$, and $\theta\in\{j\times 10^{k}: \forall j=1,5, k=-3, -2, -1, 0, 1\}$. By tuning these key hyperparameters, we aim to achieve a balance between sample quality and computational efficiency in our training process.

\subsection{2D example - Swiss roll} \label{sec:swissroll}

We next consider 2D Swiss Rolls data, and plot all $M=300$ training samples in the dataset in Figure \ref{fig:example_2D_x0}.\footnote{We use the dataset provided by the sklearn package of Python in \url{https://scikit-learn.org/stable/modules/generated/sklearn.datasets.make_swiss_roll.html}. This dataset is also used in other studies on diffusion models; see e.g. \cite{sohl2015deep} and \cite{lai2023fp}.} Consider a non-differentiable reward function
\begin{align*}
	h(y_1, y_2) = 1_{ y_1\in [-5, 6]}, \quad (y_1, y_2) \in \mathbb{R}^2.
\end{align*}
This reward function encourages generated samples to stay  within the rectangular region $[-5, 6] \times \mathbb{R}$.

\begin{figure}[h]
	\centering
	\includegraphics[scale=0.3]{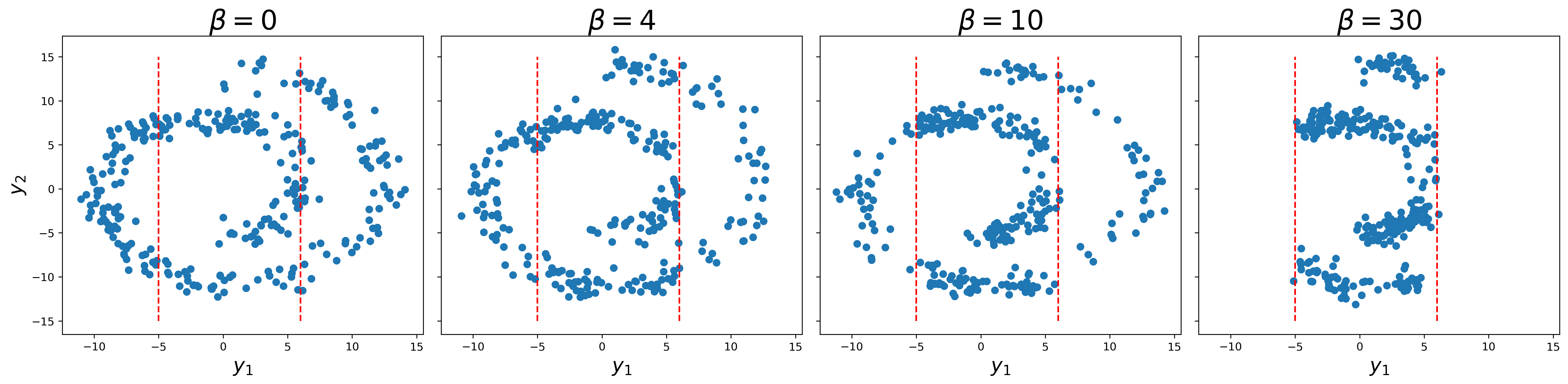}
	\caption{300 samples generated from diffusion models trained under different terminal weights $\beta$ for 2D Swiss rolls.}
	\label{fig:example_2D}
\end{figure}

 We again consider different $\beta$'s and implement Algorithm \ref{algo:offline episodic} with hyperparameters shown in Table \ref{table:param_toy}. The numerical results are visualized in Figure \ref{fig:example_2D}.
Similar to the 1D example, when $\beta = 0$, our RL algorithm produces samples distributed close to the data distribution shown in Figure \ref{fig:example_2D_x0}. As we increase the value of  $\beta$,  the generated samples become more concentrated on the rectangular region $[-5, 6] \times \mathbb{R}$. When $\beta$ is large enough, say 30, all the generated samples tend to stay inside the region where $y_1 \in[-5, 6]$,
visually resembling a French croissant.

\begin{table}[h]
	\centering
	\begin{tabular}{c || c | c c}
        \hline \multirow{2}{*}{Sample Size $m$} & $\beta = 0$ &  \multicolumn{2}{c}{$\beta > 0$} \\\cline{2-4}
        & KL$(p(\mathbf{y}_T) || p_0)$ $\downarrow$ & Reward $\mathbb{E}[h(\mathbf{y}_T)]$ & KL$(p(\mathbf{y}_T) || p_0)$ $\downarrow$\\\hline
		30 &  0.33 $\pm$ 0.017 & 0.80 $\pm$ 0.006 & 0.88 $\pm$ 0.026 \\	

		60 & 0.32 $\pm$ 0.016 & 0.80 $\pm$ 0.004 & 0.51 $\pm$ 0.027 \\

		90 & 0.20 $\pm$ 0.019 & 0.80 $\pm$ 0.004 & 0.42 $\pm$ 0.025\\				
        120 & \textbf{0.17 $\pm$ 0.016} & 0.80 $\pm$ 0.004 & 0.40 $\pm$ 0.029 \\
	
		150 &  0.18 $\pm$ 0.016 & 0.80 $\pm$ 0.004 & \textbf{0.38 $\pm$ 0.025} \\

        300 & \textbf{0.17 $\pm$ 0.019} & 0.80 $\pm$ 0.003 & 0.39 $\pm$ 0.026
		\\ \hline
		
	\end{tabular}
	\captionsetup{justification=centering}
	\caption{Effect of sample size $m$. }
	\label{table: 2D_target_size_cmp}
\end{table}

 We now discuss the effect of the sample size $m$ in the score estimator \eqref{eq:approxi-score} on the performance of our q-Learning algorithm.
 As described in Algorithm \ref{algo:offline episodic}, whenever the algorithm runs the environment simulator and generates the running reward signal $\hat r$, the simulator will first randomly select $m$ samples from the dataset of $M$ training samples, and then calculate $\widehat{ \nabla_{\mathbf{x}}\log p_{t}(\mathbf{x}) } $ and $\hat r$ according to \eqref{eq:approxi-score} and \eqref{eq:sample-r}, respectively. To see the effect of sample size $m$ on the performance of Algorithm \ref{algo:offline episodic}, we conduct two different experiments. In the first one, we run Algorithm \ref{algo:offline episodic} without terminal reward, i.e. $\beta = 0$, using various sample sizes of $m=30,60, 90, 120, 150$ and $300$, and then compute the KL-divergences between the generated data and the Swiss roll training data. In the second experiment, we fix the expected terminal reward at 0.8 and compute the same KL-divergences with different sample sizes. This is achieved by choosing different values of the weight $\beta$ for the reward in our RL formulation.

 The results are summarized in Table~\ref{table: 2D_target_size_cmp}.
 To obtain the performance metrics in the table, we generate 100 batches of $m$ samples (i.e. $\mathbf{y}_T$), compute the KL divergence to the Swiss roll data $p_0$ for each batch via the estimation method introduced in \cite{wang2009divergence}, and compute the 95\% confidence interval based on the 100 batches.
 Table~\ref{table: 2D_target_size_cmp} shows  that with a larger sample size in the ratio estimator, the generated data are generally of a higher quality in the sense that they are more similar to the training data in terms of a smaller KL divergence for this example. On the other hand, a sample size of $m=120$ or $m=150$ is already enough to produce satisfactory performance of the q-Learning algorithm. In particular, it is not necessary to always use all the available samples to estimate the score function values in our algorithm. This can help us reduce the computational time and speed up the training process. While this point is less prominent  for this two-dimensional example because training is fast anyway, it will become significant in high-dimensional applications as  we will see in image generation tasks later.

To further investigate how sample size $m$ affects the quality of score estimation, we calculate the bias, variance, and mean-squared error (MSE) of the ratio estimators \eqref{eq:approxi-score} for various $m$ and time points $t$. To compute these metrics for each $t$, we randomly generate 10,000 states $\mathbf{x}\in\mathbb{R}^2$, and for each $\mathbf{x}$, we first use 500,000 samples from $p_0$ to calculate the score estimator in \eqref{eq:approxi-score} and regard it as the true score function value $\nabla\log p_{t}(\mathbf{x})$. Then, we calculate the estimator $\widehat{ \nabla\log p_{t}(\mathbf{x}) }$  by randomly drawing $m$ samples and repeat this estimation process for 100 times. This allows us to compute (approximately) for each state--time pair $(\mathbf{x}, t)$: (i) the MSE of $\widehat{ \nabla\log p_{t}(\mathbf{x}) }$ in \eqref{eq:approxi-score}, and (ii) the bias and variance for each dimension of $\widehat{ \nabla\log p_{t}(\mathbf{x}) }$.
Since the score function value $\nabla\log p_{t}(\mathbf{x})$ is a two-dimensional vector in this example, the bias of the estimator is a 2D vector and the corresponding covariance is a $2\times2$ matrix. To focus on key summary statistics, we compute the $L_1$ norm of the bias vector, as well as the  trace of the covariance matrix which is simply the sum of variances across the two dimensions of the score estimator (referred to as ``variance" below). For each $t$, we average the bias,  variance, and MSE over 10,000 states $\mathbf{x}$. The resulting bias, variance, and MSE values for the ratio estimator \eqref{eq:approxi-score} are summarized in Table \ref{table: ratio_bias_variance}.

From Table \ref{table: ratio_bias_variance}, we observe that as the sample size $m$ increases, the bias, variance, and MSE of the ratio estimator generally decrease. Moreover, the accuracy of the estimator depends on the time dimension of the true score function $\nabla\log p_{t}(\cdot)$, with smaller MSE values obtained for larger $t$. Intuitively, as $t \rightarrow T=5,$ the distribution $p_T$ becomes approximately Gaussian, making the score easier to learn compared to the scenario where $t \rightarrow 0$, which corresponds to the true unknown data distribution.

 The above analysis suggests that while using a larger sample size in the score estimator \eqref{eq:approxi-score} in general and in theory  increases accuracy, its benefit may not outweigh the resulting training cost as a smaller sample size may suffice to achieve satisfactory performance with a shorter training time.

\begin{table}[h]
	\centering
	\begin{tabular}{c || c c c}
        \hline Sample Size $m$ & $\|$Bias$\|_1$ & Variance & MSE  \\\hline
        \multicolumn{4}{c}{$t=1$} \\\hline
		30 & $6.178\times10^{-2}$ & $5.036\times 10^{-2}$ & $5.348\times 10^{-2}$ \\

		90 & $3.346\times10^{-2}$  & $1.515\times 10^{-2}$& $1.938\times 10^{-2}$\\

		300 & $2.151\times10^{-2}$ & $4.458\times 10^{-3}$ & $8.682\times 10^{-3}$ \\\hline

        \multicolumn{4}{c}{$t=2$} \\\hline
        30 & $2.294\times10^{-2}$ & $7.504\times10^{-3}$ & $8.273\times10^{-3}$  \\
        90 & $1.393\times10^{-2}$  & $2.440\times 10^{-3}$ & $3.213\times 10^{-3}$\\
        300 & $8.723\times10^{-3}$ & $7.302\times10^{-4}$ & $1.501\times10^{-3}$ \\\hline

        \multicolumn{4}{c}{$t=4$} \\\hline
        30 & $3.069\times10^{-3}$ & $1.015\times10^{-3}$& $1.112\times10^{-3}$ \\	
        90 & $2.524\times10^{-3}$  & $3.386\times 10^{-4}$ & $4.504\times 10^{-4}$\\
        300 & $1.457\times10^{-3}$ & $1.024\times10^{-4}$ & $3.209\times10^{-4}$ \\	\hline
		
	\end{tabular}
	\caption{The bias, variance, and mean-squared error of the ratio estimator \eqref{eq:approxi-score} for the 2D Swiss roll example. }
	\label{table: ratio_bias_variance}
\end{table}

\subsection{Comparison with pretrain-then-fine-tune approach}\label{sec:finetue}

In this section, we compare our continuous-time RL approach with those developed for fine-tuning pretrained discrete-time diffusion models.

Specifically, we consider two benchmark fine-tuning methods, DDPO and DPOK, proposed by \cite{black2023training} and \cite{fan2024reinforcement}, respectively. Both DDPO and DPOK are online RL methods for fine-tuning a pretrained diffusion model to generate samples with higher terminal rewards. Note that the key difference between their formulations and ours is that the latter does not involve a pretrained model nor is it designed for fine-tuning.
For reader's convenience, we first briefly review their formulations. % while highlighting the differences compared with our formulation.

\cite{black2023training} and \cite{fan2024reinforcement} formulate the denoising process of a discrete-time diffusion model, called the denoising diffusion probabilistic model (DDPM), as a Markov decision process (MDP). Specifically, denote by $\{\mathbf{y}_{t_k}\}_{k=0}^{K}$ the denoising process (with a slight abuse of notations), where $\mathbf{y}_0$ is sampled from a normal distribution, and let the one-step transition probabilities $\{p_{\phi}(\mathbf{y}_{t_{k}}| \mathbf{y}_{t_{k - 1}})\}_{k=1}^K$ (often modeled as Gaussian) be parameterized by $\phi$. Given a pretrained diffusion model $p_{pre}=p_{\phi_0}$,
DDPO and DPOK aim to fine-tune it to maximize the expected reward of generated samples by updating $\phi$. They regard the transition $p_\phi(\mathbf{y}_{t _{k + 1}} | \mathbf{y}_{t_k})$ as a stochastic policy and apply the policy gradient algorithm. To do so, they formulate a finite-horizon MDP with state $\mathbf{s}$, action $\mathbf{a}$, policy $\bpi^\phi$, deterministic transition dynamics $P$ and reward $R$ as follows:
	\begin{align*}
		&\mathbf{s}_{t_k} := \mathbf{y}_{t_k},\quad \mathbf{a}_{t_k} := \mathbf{y}_{t_{k + 1}}, \quad \bpi^\phi(\mathbf{a}_{t_k} | \mathbf{s}_{t_k}) := p_\phi(\mathbf{y}_{t _{k + 1}} | \mathbf{y}_{t_k}),\quad P(\mathbf{s}_{t_{k + 1}} | \mathbf{s}_{t_k}, \mathbf{a}_{t_k}) := 1_{\mathbf{s}_{t_{k + 1}} = \mathbf{a}_{t_{k}}},\\
		& R(\mathbf{s}_{t_k}, \mathbf{a}_{t_k}) = 0,\ \forall k=0, ..., K - 2,\quad R(\mathbf{s}_{t_{K - 1}}, \mathbf{a}_{t_{K - 1}}) = h(\mathbf{a}_{t_{K - 1}}) = h(\mathbf{y}_{t_K}).
	\end{align*}
Denote by $p_\phi(\mathbf{y}_{0:K}):= p(\mathbf{y}_0)\prod_{k=1}^{K}p_{\phi}(\mathbf{y}_{t_{k}} | \mathbf{y}_{t_{k - 1}})$ the joint distribution of the denoising trajectory $\{\mathbf{y}_{t_k}\}_{k=0}^{K}$.
 The objective of DDPO and DPOK is given below, with $\gamma = 0$ for DDPO (i.e. purely reward-directed) and $\gamma > 0$ for DPOK:
\begin{equation}\label{eq:objective_DPOK}
	\min_{\phi}\mathbb{E}_{p_\phi(\mathbf{y}_{0:K})}\left[-\beta\cdot h(\mathbf{y}_{t_K})+ \gamma\cdot \sum_{k=1}^{K}\text{KL}\left(p_{\phi}(\mathbf{y}_{t_{k}} | \mathbf{y}_{t_{k - 1}}) || p_{\text{pre}}(\mathbf{y}_{t_{k}} | \mathbf{y}_{t_{k - 1}})\right)\right].
\end{equation}
Here, $p_{\text{pre}}(\mathbf{y}_{t_{k}} | \mathbf{y}_{t_{k - 1}})$ is the one-step transition probability distribution under the pretrained diffusion model, KL$(\cdot || \cdot)$ is the KL-divergence between distributions, and $\beta, \gamma$ are weights balancing the trade-off between the terminal reward $h$ and the deviation from the pretrained model.  One can easily see that our RL formulation  is significantly different from the ones in \cite{black2023training} and \cite{fan2024reinforcement} at least in two aspects: First, we penalize the deviation from the (unknown) true score model, rather than a (known) pretrained score or diffusion model.
Second, the setting (i.e. continuous-time) and definitions of actions and policies in our formulation are entirely different from theirs.

We now compare the performance of our algorithm with those of DDPO and DPOK. For illustrations, we focus on the 2D Swiss roll data discussed in Section~\ref{sec:swissroll} along with the reward function $h(y_1, y_2) = 1_{y_1\in[-5, 6]}$.
To adapt DDPO and DPOK to score-based diffusion models,
we consider the discrete-time denoising process $\{\mathbf{y}_{t_k}\}_{k=0}^{K}$ similar to \eqref{eq:simulator}:
\begin{equation*}
    \mathbf{y}_{t_{k + 1}} = \mathbf{y}_{t_k} + \left[\mathbf{y}_{t_k} + 2\mu^\psi(t_k, \mathbf{y}_{t_k})\right]\Delta t + \sqrt{2\Delta t}\cdot \xi,\quad \mathbf{y}_{0}\sim \nu
\end{equation*}
where $t_{k + 1} = t_k + \Delta t$, and $\mu^\psi$ is the policy or score neural network discussed earlier (see \eqref{eq:pi-psi}). Then, the one-step transition $p_{\psi}(\mathbf{y}_{t_{k}}| \mathbf{y}_{t_{k - 1}})$ is Gaussian with mean effectively parameterized by $\mu^\psi$, and one can compute the KL-divergence in \eqref{eq:objective_DPOK} explicitly.
The pretrained score network $\mu^{pre}$, as required by DDPO and DPOK, is obtained by applying our Algorithm \ref{algo:offline episodic} with hyperparameters shown in Table \ref{table:param_toy} and $\beta = 0$, although one can use other standard score matching methods. For our q-learning algorithm, the controlled state process $\mathbf{y}_t^{\mathbf{a}}$ is updated according to the dynamic \eqref{eq:simulator}. To make a fair comparison, the neural network $\mu^\psi$ used in all the three algorithms share the same structure which is the one displayed in Table \ref{table:NN_setting}. Moreover,  the pretrained network $\mu^{pre}$ is used as the initialization of $\mu^\psi$ in our algorithm.
Finally, for all the algorithms, we set $T = 10, \Delta t = 0.25$ and $K = T / \Delta t = 40$.

\begin{table}[h]
	\centering
	\begin{tabular}{c | c c }
		\hline Algorithm & Reward ($\mathbb{E}[h(\mathbf{y}_T)]$) $\uparrow$& KL$(p(\mathbf{y}_T) || p_0)$ $\downarrow$\\\hline
		Pretrained model A &  0.52 $\pm$ 0.006 & 0.17 $\pm$ 0.019  \\
		DDPO  & 1.00 $\pm$ 0.001  & 2.97 $\pm$ 0.024  \\\hline\hline
		%		DDPO  & 0.70 $\pm$ 0.00  & 0.41 $\pm$ 0.03  \\
		DPOK &  0.70 $\pm$ 0.004 &  \textbf{0.21 $\pm$ 0.021} \\				
		q-Learning  & 0.70 $\pm$ 0.005 & 0.29 $\pm$ 0.023 \\ \hline
		%		DDPO  & 0.80 $\pm$ 0.00  & 0.76 $\pm$ 0.03  \\
		DPOK  & 0.80 $\pm$ 0.004 & \textbf{0.38 $\pm$ 0.023}\\				
		q-Learning    & 0.80 $\pm$ 0.003 & 0.39 $\pm$ 0.026  \\\hline
		%		DDPO  & 0.90 $\pm$ 0.00  & 1.23 $\pm$ 0.02  \\
		DPOK &  0.90 $\pm$ 0.003 &  0.69 $\pm$ 0.024 \\				
		q-Learning  & 0.90 $\pm$ 0.003 & \textbf{0.66 $\pm$ 0.024} \\ \hline
		
	\end{tabular}
	%%\captionsetup{justification=raggedright, singlelinecheck=false}
	\caption{Empirical mean rewards and KL-divergences (with 95\% confidence) based on (100 batches of) 300 samples generated by diffusion models fine-tuned/trained with different algorithms, when a good pretrained model is available.}
	\label{table:pretrain_cmp_good}
\end{table}

\begin{figure}[h]
	\centering
	%\captionsetup{justification=raggedright, singlelinecheck=false}
	\includegraphics[scale=0.6]{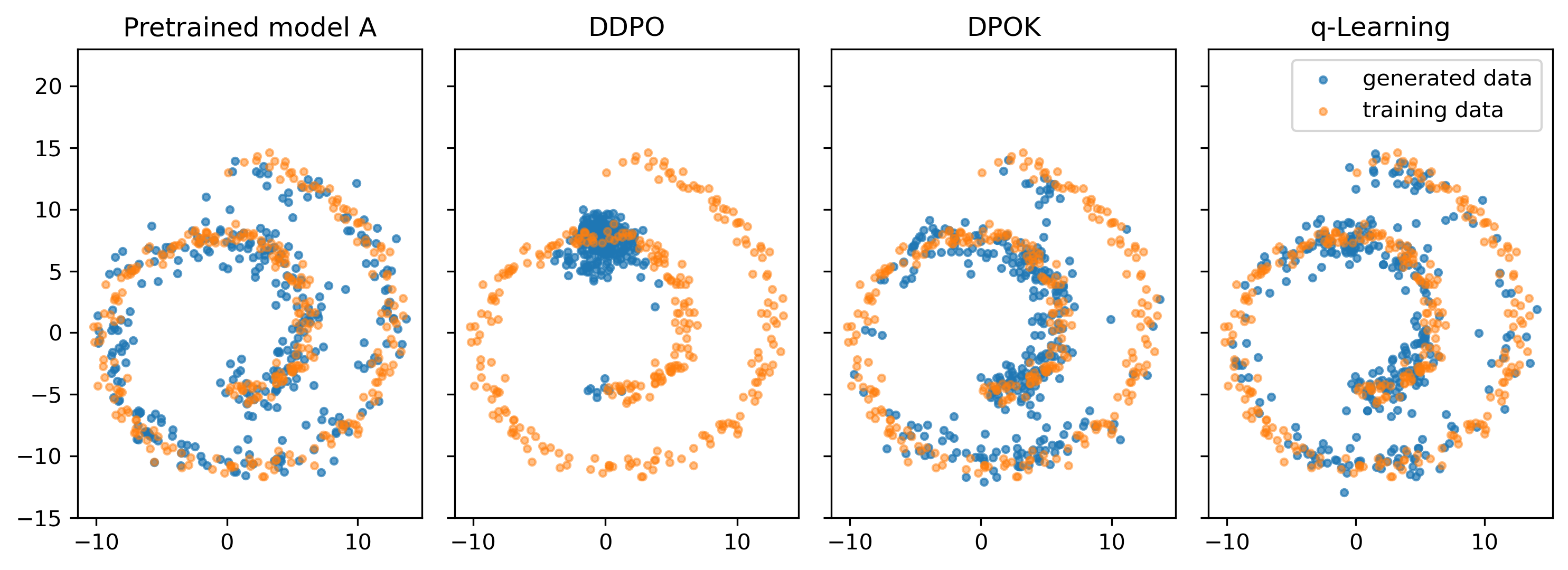}
	\caption{300 samples generated by diffusion models fine-tuned/trained by different algorithms. DPOK and $q-$Learning have a fixed expected reward of 0.80. The samples obtained from (good) Pretrained Model A are also plotted for reference.}
	\label{fig:pretrain_cmp_good}
\end{figure}

We first report the numerical results in Table \ref{table:pretrain_cmp_good} and Figure \ref{fig:pretrain_cmp_good} for the scenario when a ``good'' pretrained model (referred to as Pretrained Model A) is given. Here, by ``good" we mean that Model A generates samples whose distribution is very close to the true data distribution $p_0$ as measured by the KL divergence, which is obtained by implementing Algorithm \ref{algo:offline episodic} with $\beta=0$ and running the algorithm with 50,000 iterations.
Given this pretrained model, DDPO fine-tunes it for pure reward maximization (without any KL regularization). Table \ref{table:pretrain_cmp_good} shows that DDPO generates samples achieving the maximal empirical mean reward, but with a large KL divergence from the true data distribution. On the other hand, both DPOK and our q-learning method trade off two criteria: reward and KL-divergence. For a fair comparison, we compare the KL-divergence values of the two algorithms under the same level of expected reward of the generated samples. This is achieved by choosing different values of the weight $\beta$ for the reward in DPOK (while setting $\gamma = 1$ in its objective) and our model, because DPOK penalizes deviation from the pretrained model, while we penalize deviation from the true score model. Table \ref{table:pretrain_cmp_good} displays the empirical mean KL divergence to the data distribution $p_0$ (and the corresponding 95\% confidence interval) of 300 samples obtained by DPOK and our algorithm, when the terminal expected rewards are set to be 0.7, 0.8 and 0.9, respectively. We can observe that the performances of our q-learning algorithm and DPOK are similar (if the latter is slightly better) in the current scenario of a good pretrained model being available. For visualization, Figure \ref{fig:pretrain_cmp_good} plots  300 samples generated by different algorithms with a fixed expected reward level 0.8. It is clear that samples generated from DDPO diverge from the data distribution significantly, while samples obtained from DPOK and our q-learning algorithm have similar distributions to the data distribution.
% KL-divergences.

\begin{table}[h]
	\centering
	\begin{tabular}{c | c c }
		\hline Algorithm & Reward ($\mathbb{E}[h(\mathbf{y}_T)]$) $\uparrow$& KL$(p(\mathbf{y}_T) || p_0)$ $\downarrow$\\\hline
		Pretrained model B &  0.55 $\pm$ 0.005 & 0.68 $\pm$ 0.026  \\\hline\hline
		DPOK &  0.70 $\pm$ 0.004 &  0.90 $\pm$ 0.025 \\				
		q-Learning  & 0.70 $\pm$ 0.004& \textbf{0.30 $\pm$ 0.023} \\ \hline
		DPOK  & 0.80 $\pm$ 0.004 & 1.03 $\pm$ 0.027\\				
		q-Learning    & 0.80 $\pm$ 0.004& \textbf{0.42 $\pm$ 0.023}  \\\hline
		DPOK &  0.90 $\pm$ 0.003 &  1.21 $\pm$ 0.028 \\				
		q-Learning  & 0.90 $\pm$ 0.003& \textbf{0.64 $\pm$ 0.026} \\ \hline
		
	\end{tabular}
	% \captionsetup{justification=raggedright, singlelinecheck=false}
	\caption{Empirical mean rewards and KL-divergences (with 95\% confidence) based on (100 batches of) 300 samples generated by diffusion models fine-tuned/trained with different algorithms, when only a bad pretrained model is available. }
	\label{table:pretrain_cmp_bad}
\end{table}

\begin{figure}[h]
	\centering
	%\captionsetup{justification=raggedright, singlelinecheck=false}
	\includegraphics[scale=0.6]{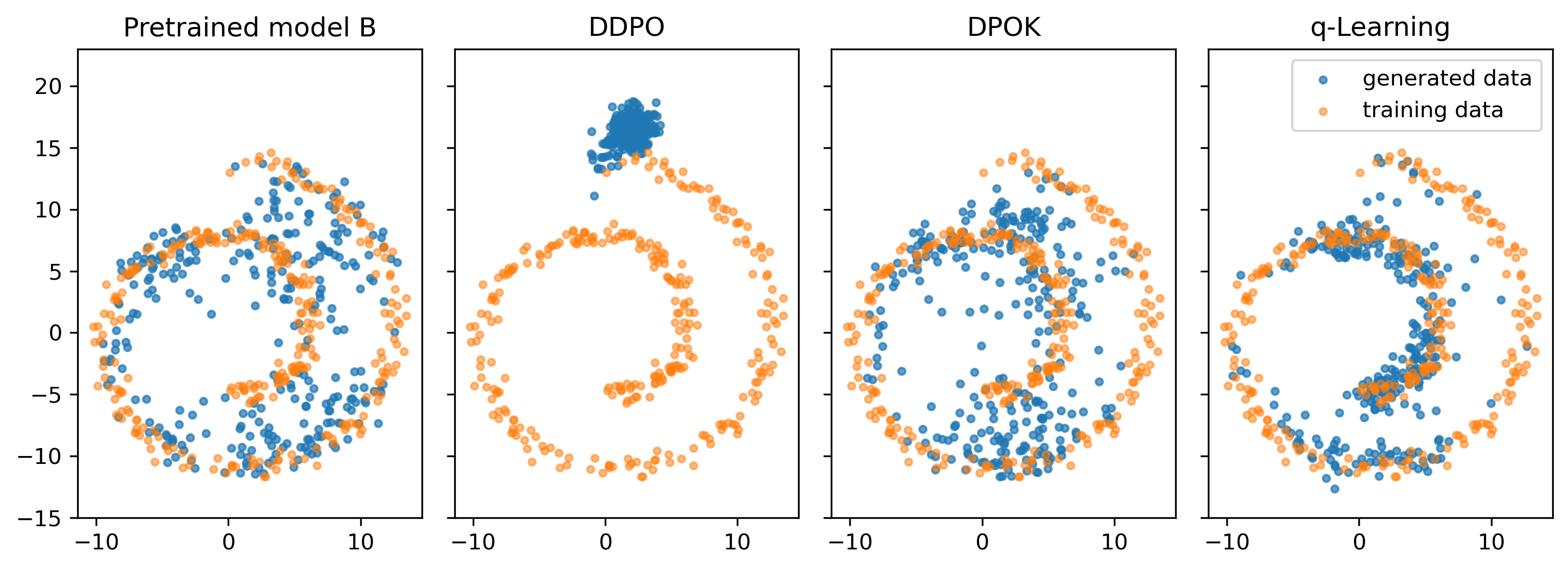}
	\caption{300 samples generated by diffusion models fine-tuned/trained by different algorithms.  DPOK and $q-$Learning have a fixed expected reward of 0.80. The samples obtained from (bad) Pretrained Model B are also plotted for reference.}
	\label{fig:pretrain_cmp_bad}
\end{figure}

While DPOK and q-Learning have similar performances in the above experiment, the two are based on very different optimization objectives. DPOK is a fine-tuning RL method, which requires the fine-tuned model to be close to the pretrained one. As a result, DPOK may not  perform well, if the pretrained model is not sufficiently good. By contrast, our method does not rely on any pretrained model and is purely data driven, resulting in  robust results.
% the fine-tuned model might encounter a performance ceiling if the pretrained one doesn't work well.
To illustrate, we repeat the above experiment but with a ``bad" pretrained model. This is referred to as Pretrained Model B, which is obtained by implementing Algorithm \ref{algo:offline episodic} with $\beta=0$ and running the algorithm with only 30,000 (instead of 50,000) iterations. In particular, the generated distribution from Model B has a much higher KL divergence (0.68 vs. 0.17 for Model A) with respect to the true data distribution; see also the first panel of Figure \ref{fig:pretrain_cmp_bad} for a visual.  Table \ref{table:pretrain_cmp_bad} reports the corresponding performances of DPOK and the q-learning algorithm. For a fixed target reward level, DPOK now performs much worse than our algorithm: the former generates samples with a distribution much further away from the true data distribution than the latter does. Interestingly, our algorithm can even improve the quality of a bad pretrained model in the sense of reducing the KL-divergence from the data distribution $p_0$, while DPOK generates samples with an even bigger divergence when applied for reward maximization. For visualization, Figure \ref{fig:pretrain_cmp_bad} shows 300 samples generated by different algorithms with a fixed expected reward level 0.8. The outperformance of our algorithm is evident in the case when only a low-quality pretrained model is available.

To summarize the findings, DDPO of \cite{black2023training} suffers from the issue of reward over-optimization, where the generated distributions diverge too far from the original data distribution.
DPOK of \cite{fan2024reinforcement} has a similar performance as our q-learning algorithm, {\it provided} that the pretrained model is of good quality. However, if the pretrained model is not good enough, our q-learning algorithm outperforms DPOK significantly.

\section{Image Generations}\label{sec:image}
In this section, we test Algorithm \ref{algo:offline episodic}  on a reward-directed image generation task. The training samples are 50,000 RGB images of size 32$\times$32 from the CIFAR-10 dataset \citep{krizhevsky2009learning},  some of which are displayed in Figure \ref{fig: cifar10_exp}(a). We consider a reward function from \cite{black2023training} based on imcompressibility, where the reward is the file size of generated images after JPEG compression. Note that the resolution of generated samples is fixed at 32$\times$32 pixels, so the file size is determined solely by the compressibility of the images.
In the following, we first describe the experimental settings (which are different from the settings for the two toy examples discussed earlier) and then report the numerical results.

\subsection{Experimental Settings}
\begin{itemize}

\item \textbf{SDE/Environment simulator.}

In this experiment, we consider the following environment simulator inspired by the celebrated DDPM model \citep{Ho2020} for image generations: for some small $\Delta t$ and $t < T=1$,
\begin{align}\label{eq: image_simulator}
    \mathbf{y}_{t + \Delta t}^{\boldsymbol{\pi}} = \frac{1}{\sqrt{1 - \beta(1 - t)\cdot \Delta t}}\left(\mathbf{y}_{t}^{\boldsymbol{\pi}} + \beta(1 - t)\cdot\Delta t\cdot a_{t}^{\boldsymbol{\pi}}\right) + \sqrt{\beta(1 - t)\cdot\Delta t}\cdot \xi,
\end{align}
where $\xi \sim \mathcal{N}(\mathbf{0}, I_d)$, $\beta(t)$ is a linear function of $t$, and $\mathbf{y}_{0}^{\boldsymbol{\pi}}\sim \mathcal{N}(\mathbf{0}, I_d)$. The simulator \eqref{eq: image_simulator} corresponds to an appropriate discretization of the
controlled dynamic \eqref{eq:samplestate} where $f(t)= \frac{1}{2} g(t)^2$ and $f$ is linear in $t$, as well as to the VP-SDE model introduced in \cite{SongICLR2021}. To match the settings in \cite{Ho2020}, we use the linear schedule $\beta(t)$ such that $\beta(\frac{1}{1000}) = 10^{-1}\cdot \frac{1}{1000\Delta t}$ and $\beta(1) = 20 \cdot \frac{1}{1000\Delta t}$. In particular,
if we use $\Delta t = 0.001$ and simulate \eqref{eq: image_simulator} in $K= \frac{T}{\Delta t} =1000$ steps, the simulator (with the action replaced by the trained score function obtained from denoising score matching) corresponds to the sampler used in the DDPM model in \citep{Ho2020}. In the implementation of our RL algorithm, we do not use $K=1000$ because it is time-consuming to simulate 1000 steps in each episode. Instead,
we consider $\Delta t = 0.05$ so that we run \eqref{eq: image_simulator} with $K=20$ steps in each episode. In this case, the linear schedule we use becomes
$\beta(t) = \frac{389}{999}t + \frac{8}{4995}$.
We find that $K=20$ sampling steps are enough to generate images of an acceptable quality. This is demonstrated in Figure \ref{fig: cifar10_exp}(b), where we display image samples generated by running
\eqref{eq: image_simulator} with $K=20$ steps, and replacing the action $a_{t}^{\boldsymbol{\pi}}$ by the score function approximator trained by DDPM.

\item \textbf{Actor and Critic Networks}

For the actor network $\mu^{\psi}(\cdot,\cdot)$, we adopt U-Net \citep{ronneberger2015u}, which is a popular neural network for image generation tasks; see e.g. \cite{Ho2020, SongICLR2021}. We initialize the actor network with the pre-trained DDPM model in the implementation of our q-Learning algorithm, which speeds up the training process.

The critic network $J^{\Theta}(\cdot,\cdot)$ consists of a U-Net that shares the same architecture with $\mu^\psi$ which we denote by $U^{\Theta_1}$, a convolution layer $\text{Conv}^{\Theta_2}$ that has one output channel,
% \footnote{We use the 'Conv2d' function in Pytorch, see more details on \url{https://docs.pytorch.org/docs/stable/generated/torch.nn.Conv2d.html}},
and a multilayer perception denoted by $\text{MLP}^{\Theta_3}$ for which we use a dropout function to deal with the output of each layer to prevent overfitting. Technically, we have
\begin{align*}
    J^\Theta(\mathbf{y}, t) := \text{MLP}^{\Theta_3}(\text{Conv}^{\Theta_2}(U^{\Theta_1}(\mathbf{y}, t))),
\end{align*}
where $\Theta = (\Theta_1, \Theta_2, \Theta_3)$. Further details about these  component networks are given in Table \ref{table:NN_setting_cifar10}.

\begin{figure}[h]
% \advance\leftskip-1cm
\begin{minipage}{\linewidth}
\captionsetup{justification=centering}
\centering
    \subcaptionbox{Original CIFAR-10 Samples}{\includegraphics[scale=0.65]{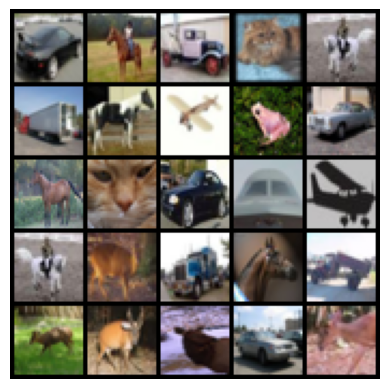}}
    \hspace{0.05\linewidth}
    \subcaptionbox{Samples from a Pre-trained Model}{\includegraphics[scale=0.65]{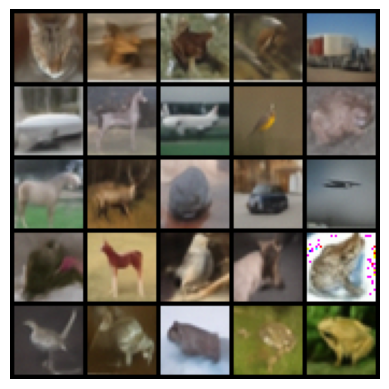}}
\end{minipage}
\caption{Sample images from the CIFAR-10 dataset (left) and those generated by the pre-trained DDPM model (right) with $20$ sampling steps.}
\label{fig: cifar10_exp}
\end{figure}

\begin{table}[h]
    \centering
    \begin{tabular}{c | c c}
        \hline Layer ID & Input Dimension & Output Dimension \\
        \hline
        U-Net $U^{\Theta_1}$& $3 \times 32 \times 32 + 1$ & $3 \times 32 \times 32$\\
        $\text{Conv}^{\Theta_2}$ & $3 \times 32 \times 32$ & $1 \times 25 \times 25$\\
        $\text{MLP}^{\Theta_3}$: Input & $25 \times 25$ &  256\\
        $\text{MLP}^{\Theta_3}$ - Hidden 1 & 256 &  64\\
        $\text{MLP}^{\Theta_3}$ - Hidden 2 & 64 &  16\\	
         $\text{MLP}^{\Theta_3}$ - Output & 16 & 1\\ \hline
    \end{tabular}
    \captionsetup{justification=centering}
    \caption{Components of the critic network for the experiment on CIFAR-10}
    \label{table:NN_setting_cifar10}
\end{table}

\item \textbf{Hyperparameters.}
% Apart from those hyperparameters specified for the simulator \eqref{eq: image_simulator} and the actor and value network, we also perform some hyperparameter search for our q-Learning algorithm.

We set the temperature parameter $\theta=10^{-3}$, the learning rates $\alpha_\psi=1\times 10^{-5}$ and $\alpha_\Theta=2\times 10^{-5}$. These key hyperparamters are tuned via grid search over the space $\{j \times 10^{-k}:\  j, k =1,2 ,3, 4,5\}$ to optimize the training process.
We use a batch size of $B=32$ in each episode. We also test $B=64$ and $B=128$, and the results are similar.
Finally, we use a sample size of $m=1$ to calculate the score estimator \eqref{eq:approxi-score}. This choice will be discussed in details later in Section~\ref{sec:m-effect}. Note that when $m=1$, the lower bound parameter $\epsilon$ in \eqref{eq:approxi-score} no longer matters, which will be dropped in implementing the q-Learning algorithm.

\end{itemize}

\subsection{Numerical results}\label{sec:results-image}

We now report the numerical results and compare the performance of our algorithm with that of DPOK in \eqref{eq:objective_DPOK}. We do not consider DDPO here because it is prone to reward over-optimization as we have already discussed in the 2D example in Section \ref{sec:swissroll}. For DPOK, we also consider $K=20$ sampling steps in each episode and set $\alpha_\psi=1\times 10^{-5}$ and $B=32$ as with our q-Learning algorithm.
Similar as in Section~\ref{sec:finetue}, we fix the same
level of the expected reward (the incompressibility score) of the generated samples by choosing different value of the weight $\beta$ for the terminal reward in the algorithms. We then compare the FID (Fr\'{e}chet Inception Distance; \citealp{heusel2017gans}) scores for the generated images under the two algorithms. We use FID instead of KL-divegence, because the former  is a more widely-used performance metric that measures the similarity of generated images.
The results of the q-Learning algorithm and DPOK are presented in Table \ref{table: cifar10_size_cmp}, where each reward is the average incompressibility score of 20,000 images generated by the corresponding algorithm, and the FID score is calculated based on 20,000 generated images and 20,000 images from the CIFAR-10 dataset.

\begin{figure}[h]
% \advance\leftskip-1cm
\begin{minipage}{\linewidth}
\captionsetup{justification=centering}
\centering
    \subcaptionbox{q-Learning: $h=1.30$, FID=50.58}{\includegraphics[scale=0.65]{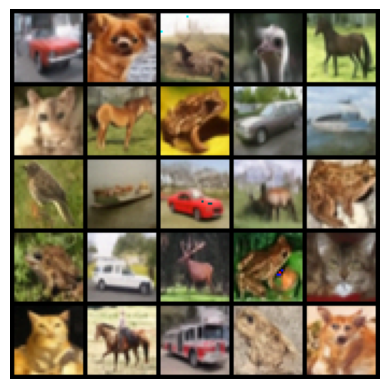}}
    \hspace{0.05\linewidth}
    \subcaptionbox{DPOK: $h=1.30$, FID=107.99}{\includegraphics[scale=0.65]{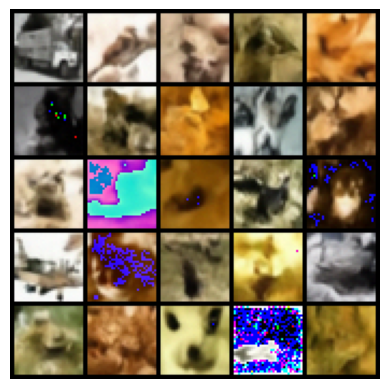}}
\end{minipage}
\caption{Sample images generated by diffusion models trained by q-Learning algorithm (left) and DPOK (right). The average reward is fixed at 1.30 KB.}
\label{fig: cifar10_size_exp}
\end{figure}

Table \ref{table: cifar10_size_cmp} shows that, with a fixed (terminal) reward level, our q-Learning algorithm generates images with much lower FID scores (lower is better) than those from DPOK. For visualization, we also
display some randomly selected samples of the generated images from our algorithm and DPOK  in Figure \ref{fig: cifar10_size_exp}, where the expected reward is fixed at 1.30 KB. We can see that the q-Learning images have more details in their background compared with those generated from the pre-trained model in Figure \ref{fig: cifar10_exp}. On the other hand, DPOK produces images of notably lower quality (higher FID), with a sizable portion barely recognizable or even unidentifiable. Indeed, approximately $15\%$ of the 20,000 images generated by DPOK exhibit a substantial amount of noise, as evident in Figure \ref{fig: cifar10_size_exp}(b).
This is primarily because DOPK penalizes deviations from the pre-trained model, yet the samples generated from the pre-trained model is not of high quality (Figure \ref{fig: cifar10_exp}(b)). By contrast, our RL formulation penalizes deviations from the true data distribution.
When the expected reward level is fixed at 1.20 KB, we observe from Table \ref{table: cifar10_size_cmp} that q-Learning even improves the pre-trained model by reducing the FID score while achieving actually a higher terminal reward (file size).

\begin{table}[h]
\captionsetup{justification=centering}
\advance\leftskip-1cm
    \centering
    \begin{tabular}{c | c | c c}
        \hline Algorithm & Reward (KB) & \makecell{FID} $\downarrow$ %& \makecell{Number of \\ Episodes}  $\downarrow$
        & \makecell{Training \\ Time (hrs.)} $\downarrow$  \\\hline
        Pre-trained Model & 1.10 & 29.90  & - \\
        \hline
        q-Learning & \multirow{2}{*}{1.20}& \textbf{26.64} %& 11,000
        & $\sim$ 8 \\
        DPOK & & 77.94 %& \textbf{7,000}
        & $\sim$ \textbf{4} \\\hline	

        q-Learning & \multirow{2}{*}{1.30}& \textbf{50.58} % & 11,000
        & $\sim$ 8\\
        DPOK & & 107.99 %&  \textbf{10,000}
        & $\sim$ \textbf{6}\\ \hline

        q-Learning & \multirow{2}{*}{1.40}& \textbf{89.79} %& 12,000
        & $\sim$ 9 \\
        DPOK & & 138.89 %&  \textbf{11,000}
        & $\sim$ \textbf{7}\\
        \hline
    \end{tabular}
    \caption{Comparison between q-Learning algorithm and DPOK.}
    \label{table: cifar10_size_cmp}
\end{table}

\begin{figure}[h]
% \advance\leftskip-1cm
\begin{minipage}{\linewidth}
\captionsetup{justification=centering}
\centering
    \subcaptionbox{Reward Curve}{\includegraphics[scale=0.345]{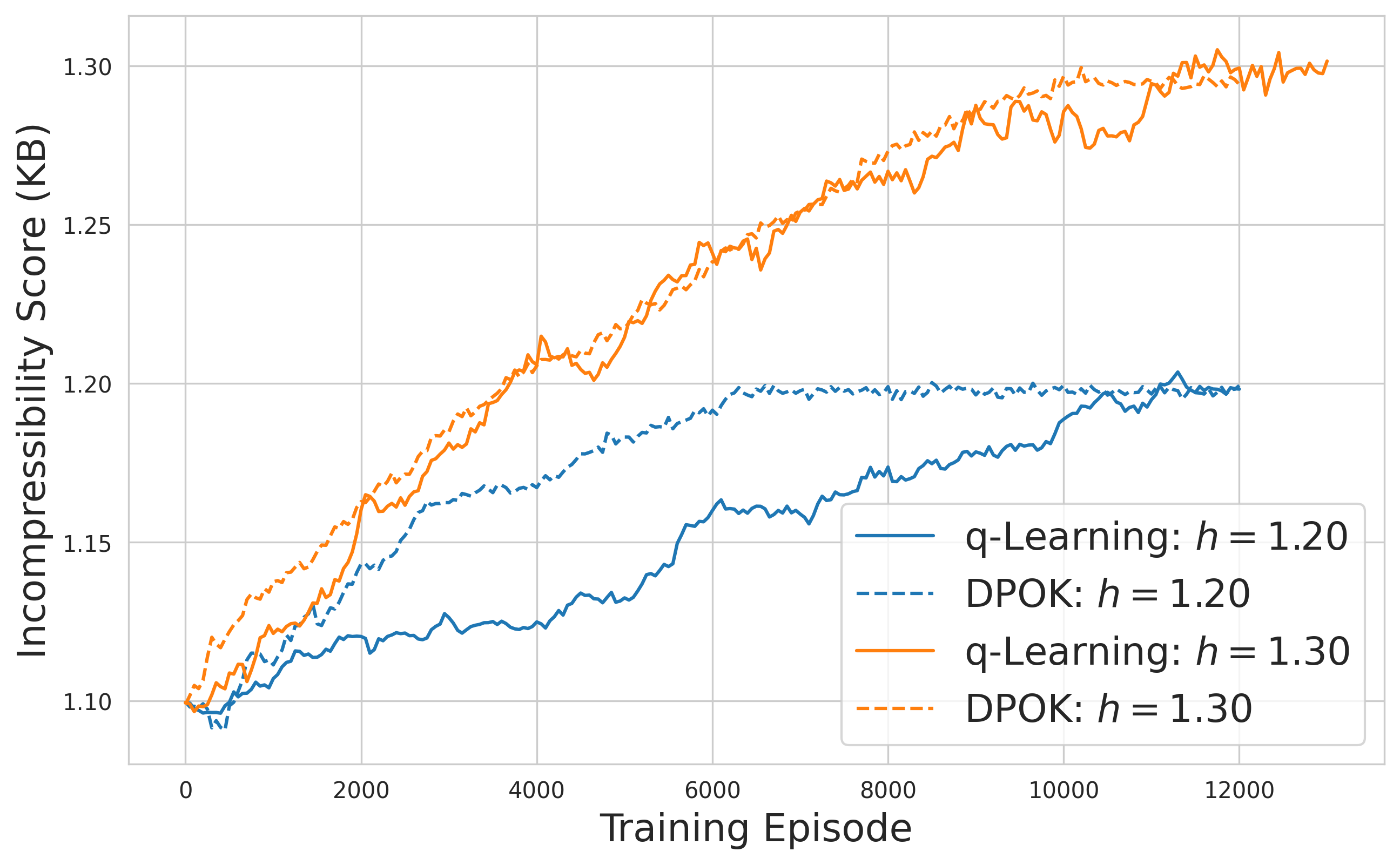}}
    \subcaptionbox{FID (to CIFAR-10) Curve}{\includegraphics[scale=0.345]{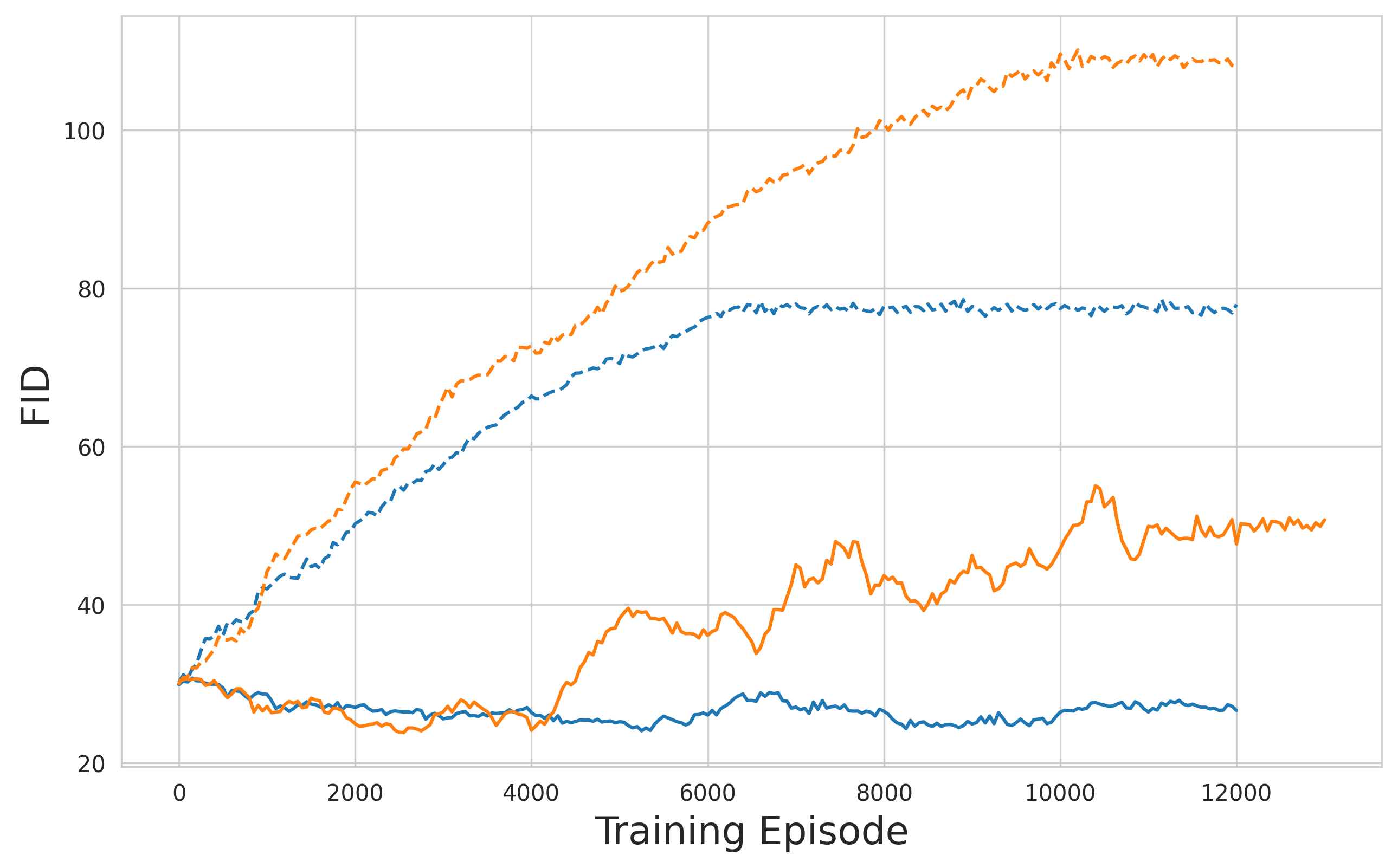}}
\end{minipage}
\caption{Training processes of q-Learning  (solid lines) and DPOK (dashed lines). The blue and orange curves correspond to final average terminal reward at 1.20 KB and 1.30 KB, respectively. }
\label{fig: cifar10_size_cmp}
\end{figure}

We next briefly discuss the training processes, including the training times for q-learning and DPOK\footnote{Our experiments are conducted on a
64-bit Linux operating system with 2 Intel Xeon Gold 5320 CPUs @ 2.20GHz and 4 A30 GPUs, each with 24GB of memory. Each algorithm in the
experiments is trained on a single GPU, and the corresponding training time is recorded.}. The last column of Table \ref{table: cifar10_size_cmp} indicates that q-Learning takes longer to converge than DPOK. Here, the training time of each algorithm is  recorded when the performance metrics including the reward and FID curve converge or stabilize. The longer training time of the q-Learning algorithm is primarily due to the additional value/critic network involved. In particular, while DPOK only optimizes the policy/actor network, q-Learning has to update parameters of both the actor and critic networks at each episode, inevitably  increasing its computational time. In our experiment, it takes around 2.55 and 2.02 seconds to run one episode for q-Learning and DPOK, respectively.
Figure \ref{fig: cifar10_size_cmp}
 further displays the training processes (learning curves) of the two methods in terms of reward and FID, where each reward and FID value is estimated based on 20,000 images generated by models saved during the training process. Figure \ref{fig: cifar10_size_cmp} (a) shows a slower convergence of q-Learning to the desired incompressibility score of 1.20 KB (while the convergence speeds are similar for the score of 1.30 KB): DPOK only needs less than 7,000 episodes while q-Learning takes around 11,500 episodes to reach the target of 1.20 KB. Again, the reason is because
 optimizing the actor and critic networks simultaneously compels more episodes for  q-Learning to converge.  Despite the longer training time, however, Figure \ref{fig: cifar10_size_cmp} (b) indicates that q-learning outperforms DPOK decisively and significantly in terms of keeping the generated images closer to the original CIFAR-10 dataset.

\subsection{Choice of sample size $m$ in ratio estimator (\ref{eq:approxi-score})} \label{sec:m-effect}
In this subsection, we discuss the impact  of the sample size $m$ in the ratio estimator \eqref{eq:approxi-score} on the performance of the q-Learning algorithm in image generations.

For the results presented in Table~\ref{table: cifar10_size_cmp} we use a sample size $m=1$ in \eqref{eq:approxi-score} for obtaining the running reward signal during training. In general, images are considered samples from a high-dimensional distribution; as such
using a larger sample size $m$ will significantly increase training time which may greatly outweigh the benefit of an increased quality of the trained model.
To see this, we conduct an experiment where three diffusion models with $m=1, 10$ and $100$ are trained via q-Learning to reach a terminal incompressibility score of 1.30 KB by tuning the parameter $\beta$. The results, including performance metrics (FID) and training times, are summarized in Table \ref{table: cifar10_sample_size_cmp}. To generate this table, we calculate the FID score using 20,000 original and generated CIFAR-10 images, and the average time cost per episode is calculated by dividing the total computational time of the 10,000 episodes by 10,000. We also track the number of episodes and total training time required for each model to converge.

 Table \ref{table: cifar10_sample_size_cmp} shows that diffusion models trained using q-Learning under different sample sizes $m$ yield comparable FID scores for their generated images, indicating similar image qualities. (In fact, $m=1$ even yields the smallest FID score!) However, increasing the sample size $m$ significantly prolongs the time spent per episode, thereby substantially enlarging the total training time despite faster convergence (i.e. fewer episodes to converge). This experiment suggests that a sample size of $m=1$ is sufficient for this image generation task, allowing us to train a high-quality diffusion model for achieving a target terminal reward level in the shortest amount of time.

\begin{table}[h]
    \centering
    \begin{tabular}{c |  c c c c}
        \hline \makecell{Sample \\Size $m$} & \makecell{FID  } $\downarrow$
        & \makecell{Time per \\Episode (sec.)} $\downarrow$  & \makecell{Number of \\ Episodes}  $\downarrow$ & \makecell{Training \\Time (hrs.)} $\downarrow$  \\\hline
        1 & \textbf{50.58 }
            & \textbf{2.55 } & $\sim 11,000$ & $\sim \textbf{8}$\\
        10 &52.72
            & 3.74  & $\sim 10,000$ & $\sim 11$ \\
        100 &51.34
            & 5.61 & $\sim \textbf{9,000}$ & $\sim 15$\\\hline
%         1 & \textbf{47.17}
%            & 15.05 & $\sim \textbf{4,500}$ & $\sim 20$\\	
%        \hline
    \end{tabular}
    \caption{Impact of sample size $m$ in the ratio estimator \eqref{eq:approxi-score} on the performance of the q-Learning algorithm. The expected terminal reward level is fixed at 1.30 (KB). }
    \label{table: cifar10_sample_size_cmp}
\end{table}

This seemingly surprising result can be explained from two perspectives, one at a general/conceptual level and the other at a specific/technical one. The conceptual one is due to the RL approach employed.  Although at a given time the algorithm generates a reward signal (i.e. a ratio estimator) using only one sample at random, at the next time point it generates another signal using another sample. When the number of episodes is large and/or the time step is small, a large number of samples will actually be used to produce these signals during the {\it entire} learning process. Consequently, all the noises in the reward signals will be eventually averaged out. The RL agent can
improve her policies by interacting with the environment and observing such reward (reinforcement) signals, and each of these signals does not need to be accurate and can be noisy.

The technical reason is specific to the image generation task and the associated score estimator \eqref{eq:approxi-score}. Ignoring $\epsilon$, we note that the score estimator $\widehat{ \nabla_{\mathbf{x}}\log p_{t}(\mathbf{x}) }$ in \eqref{eq:approxi-score} is based on the weighted combination of $m$ samples:
\begin{align}\label{eq:ratio-m}
 \sum_{i=1}^m  \frac{     p_{t|0}(\mathbf{x} | \mathbf{x}^{i}_{0})  }{  \sum_{i=1}^m   p_{t|0}(\mathbf{x} | \mathbf{x}^i_{0}) }  \cdot  {\mathbf{x}^{i}_{0}  }.
\end{align}
By \eqref{eq: conditional_density}, the conditional density $p_{t|0}(\mathbf{x} | \mathbf{x}^i_{0}) \propto  \exp\left(-  \Vert\mathbf{x}-e^{-\int_{0}^{t}f(s)ds}\mathbf{x}^{i}_{0}\Vert^{2}  \right)$. However, in a high-dimensional setting such as CIFAR-10 images with dimension $3072$ $(3 \times 32 \times 32)$, the data points are typically sparse. Hence for a given $\mathbf{x}$, $p_{t|0}(\mathbf{x} | \mathbf{x}^i_{0})$ are typically very small  and thus the weights $\frac{     p_{t|0}(\mathbf{x} | \mathbf{x}^{i}_{0})  }{  \sum_{i=1}^m   p_{t|0}(\mathbf{x} | \mathbf{x}^i_{0}) }$  close to zero, except for the data point, say $\mathbf{x}^j_{0}$, that is {\it closest} to $ \mathbf{x} \cdot e^{\int_{0}^{t} f(s)ds}$.
To illustrate this, consider a concrete example where we have only $m=2$ samples $\mathbf{x}^{1}_{0}$ and $ \mathbf{x}^{2}_{0}$, i.e., 2 images sampled from the CIFAR-10 dataset, along with a specific initial state $\mathbf{x}$ sampled from the standard multivariate normal distribution.
The specific values of $\mathbf{x},  \mathbf{x}^{1}_{0}$ and $\mathbf{x}^{2}_{0}$ are not given explicitly here because they are all vectors in $\mathbb{R}^{3072}$.
We find that $p_{1|0}(\mathbf{x} | \mathbf{x}^{1}_{0}) = \exp(-4200.7139)$, and $p_{1|0}(\mathbf{x} | \mathbf{x}^{2}_{0}) = \exp(-4229.2939)$. This yields a ratio  $p_{1|0}(\mathbf{x} | \mathbf{x}^{1}_{0}) / p_{1|0}(\mathbf{x} | \mathbf{x}^{2}_{0}) =\exp(28.5799)\approx10^{12}$, indicating that the weighted combination in \eqref{eq:ratio-m} is essentially determined by $\mathbf{x}^{1}_{0}$ only. 
This observation holds for $t< T=1$ and $m>2$ as well.
This provides an intuition for why in our training process, a very small number of samples or indeed even one sample may be sufficient for computing the score estimator in \eqref{eq:approxi-score} in high-dimensional spaces, when the number of episodes is large and/or the time step is small.

 The preceding argument also suggests that one might enhance the accuracy of the score estimator  by using nearest-neighbor search to compute \eqref{eq:ratio-m}, i.e., finding the sample image $\mathbf{x}^i_{0}$ closest to $\mathbf{x} \cdot e^{\int_{0}^{t} f(s)ds}$ at $(\mathbf{x}, t)$. However, as expected this method is computationally costly. To demonstrate this, we conduct an experiment where we train our model using the q-Learning algorithm with a fixed expected terminal reward level of 1.30 KB, employing nearest-neighbor search to compute the score estimator. Although the resulting model achieves a slightly lower FID score of 47.17, the time per episode increases significantly to 15.05 seconds, and the entire training process takes approximately 20 hours to complete (in contrast to the results reported in Table \ref{table: cifar10_sample_size_cmp}). In comparison, our simplified approach of randomly selecting a data point (with $m=1$) reduces computational overhead substantially while keeping a good image quality because it still makes use of the valuable insights from our analysis.

%%%%%%%%%%%%%%%%%%%%%%%%

\section{Extensions}\label{sec:ext}

In this section, we discuss two extensions of our SDE-based formulation.

\subsection{ODE-based formulation}\label{sec:ode}
In addition to the SDE-based implementation of diffusion models, another mainstream approach for sample generation is the probability flow ODE implementation \citep{SongICLR2021}.
In this subsection, we show that our SDE-based RL framework for reward maximization can be easily extended to the ODE-based formulation.

Recall that the forward process $(\mathbf{x}_{t})_{t \in [0, T]}$ satisfies the following SDE:
\begin{equation}\label{OU:SDE2}
d\mathbf{x}_{t}=-f(t)\mathbf{x}_{t}dt+g(t)d\mathbf{B}_{t},  \qquad {\mathbf{x}}_{0}\sim p_{0},
\end{equation}
and $p_{t}(\cdot)$ denotes the probability density function of $\mathbf{x}_t$ in \eqref{OU:SDE2}. \cite{SongICLR2021} show that there exists an ODE:
\begin{align}
\frac{d\bar{\mathbf{x}}_{t}}{dt}=f(T-t)\bar{\mathbf{x}}_{t}+ \frac{1}{2}(g(T-t))^{2}\nabla_{\mathbf{x}}\log p_{T-t}(\bar{\mathbf{x}}_{t}),
\qquad\bar{\mathbf{x}}_{0}\sim p_{T},\label{eq:ODEReverse}
\end{align}
whose solution  at time $t\in[0,T]$, $\bar{\mathbf{x}}_{t}$, is distributed according to $p_{T-t}$, i.e., the ODE \eqref{eq:ODEReverse} induces the same \textit{marginal} probability density function as the SDE in \eqref{eq:yt}.
In particular, $\bar{\mathbf{x}}_{T} \sim p_0.$
The ODE \eqref{eq:ODEReverse} is called the \textit{probability flow ODE}.

Motivated by the SDE-based problem formulation \eqref{eq:CT-RL}, we now consider an ODE-based formulation (with a slight abuse of notation):
\begin{align}\label{eq:CT-RL-ODE}
\max_{{\mathbf{a}} = (a_t: 0 \le t \le T)}  \left\{ \beta \cdot \mathbb{E}[h( \mathbf{y}^{\mathbf{a}}_{T})] -  \mathbb{E} \left[ \int_0^T (g(T-t))^{2} \cdot {| \nabla\log p_{T-t}({\mathbf{y}}^{\mathbf{a}}_{t}) - a_t|^2} dt \right]  \right\}
\end{align}
where
\begin{equation}\label{eq:u-sys-ode}
d\mathbf{y}^{\mathbf{a}}_{t}=\left[f(T-t)\mathbf{y}^{\mathbf{a}}_{t}+ \frac{1}{2}(g(T-t))^{2} a_t \right]dt, \quad \mathbf{y}_{0}\sim\nu.
\end{equation}
Because the dynamics $(\mathbf{y}^{\mathbf{a}}_{t})$ is described by a controlled ODE, the Hamiltonian for this problem becomes
\begin{align}\label{eq:Hal-ODE}
H(t, y, a, p) =   -(g(T-t))^{2}  | \nabla\log p_{T-t}(y) - a|^2  +  [f(T-t) y +  \frac{1}{2} (g(T-t))^{2} a ] \circ p. %+ \frac{1}{2}(g(T-t))^{2} \circ  q,
\end{align}

\begin{remark}\label{remark:ODE_KL}
Because \eqref{eq:u-sys-ode} is an ODE, instead of  an SDE, the second term in the objective \eqref{eq:CT-RL-ODE} can no longer be directly interpreted as the KL divegence between two path measures as in \eqref{eq:KL}. However, we still keep this term in the objective as in the SDE-based formulation to encourage the action process not to deviate too much away from the true score function.
\end{remark}

Similar to Section~\ref{sec:explora}, we can write down the exploratory RL formulation of the above problem. With slight abuse of notations, the exploratory dynamic is given by the following ODE with random/normal initialization:
        \begin{align}\label{eq: exp_state_ODE}
        d \tilde {\mathbf{y}}^{\bpi}_s= \left[f(T- s) \tilde{\mathbf{y}}^{\bpi}_{s} + \frac{1}{2} g^2(T-s) {\int_{\mathbb{R}^d} a {\bpi}(a|s, \tilde {\mathbf{y}}^{\bpi}_s ) da} \right]  ds , \quad \tilde{\mathbf{y}}_{0}^{\bpi} \sim\nu.
    \end{align}
The associated entropy-regularized control objective is given by:
\begin{align} \label{eq: value_function_ode}
        &J(t, y, \bpi) \nonumber\\
        = &\mathbb{E}_{t,y}\bigg[  \int_t^T \int_{\mathbb{R}^d} \big( {r(s, \tilde{\mathbf{y}}_s^{\bpi}, a)}\ - { \theta \log \bpi(a|s, \tilde{\mathbf{y}}_{s}^{\bpi})\big) \bpi(a|s, \tilde{\mathbf{y}}_{s}^{\bpi}) da} ds + {\beta h(\tilde{\mathbf{y}}_T^{\bpi} )}\bigg],
    \end{align}
 where $r$ is given in \eqref{eq:rt}. The goal of RL is to solve the following optimization problem:
\begin{align}\label{eq:Jstar-ODE}
	\max_{\bpi \in \bPi} \int J(0, y, \bpi) d \nu(y),
\end{align}
where $\boldsymbol{\Pi}$ stands for the set of admissible stochastic policies. As in \eqref{eq: exp_state_SDE}, for actual execution of a stochastic policy $\bpi$ in the RL algorithm, we need the discretely sampled state process, which satisfies the following ODE: for all $i = 0, . . . , K - 1$ and all $s \in [t_{i}, t_{i+1})$,
\begin{align} \label{eq:samplestate-ode}
d \mathbf{y}^{\bpi, \mathbb{S}}_s
= [f(T-s) \mathbf{y}^{\bpi, \mathbb{S}}_s + \frac{1}{2}(g(T-t))^{2} {\bf a}^{\bpi}_{t_i} ] ds,
\end{align}
where $\mathbb{S}= (t_i)_{i=0, \ldots, K}$ is a grid of $[0, T]$ and ${\bf a}^{\bpi}_{t_i} \sim \bpi(\cdot|t_i, \mathbf{y}^{\bpi, \mathbb{S}}_{t_i})$.

It is easy to see that the theory and algorithm for the SDE-based formulation can be developed for the ODE-based formulation analogously. For instance, because the Hamiltonian in \eqref{eq:Hal-ODE} is still a quadratic function of the action $a$, we deduce that the optimal stochastic policy for \eqref{eq:Jstar-ODE} is still a Gaussian distribution, which has the same form as \eqref{eq:pi-star}. We can still apply Algorithm~\ref{algo:offline episodic} to solve the RL problem \eqref{eq:Jstar-ODE}, except one importance difference. In Algorithm~\ref{algo:offline episodic}, the environment simulator (i.e. the sampler) is based on the discretization (e.g. Euler--Maruyama) of the SDE in \eqref{eq:samplestate}. For the ODE-based formulation, we instead use a numerical solver of the ODE \eqref{eq:samplestate-ode} as the sampler. Many ODE solvers have been developed and employed in practice. For instance, one can use Euler \citep{SongICLR2021}, DDIM \citep{song2020denoising}, Heun's 2nd order method \citep{Karras2022}, DPM solver \citep{lu2022dpm}, exponential integrator \citep{zhang2023fast}, among others.

To test our method experimentally, we consider the Swiss roll dataset and implement Algorithm~\ref{algo:offline episodic} for two ODE-based samplers/simulators: ODE-Euler of \cite{SongICLR2021}  and DDIM of \cite{song2020denoising}. % which is based Euler discretization of a time-reparametrized ODE
Following \cite{song2020denoising}, we set $f\equiv 0$ and $g\equiv \sqrt{2}$ in our experiment. Then from \eqref{eq:samplestate-ode} we can obtain that for a uniform grid with size $\Delta t$, the update rule for the ODE-Euler simulator is given by (again with a slight abuse of notation and suppressing grid notation for clarity)
\begin{equation*}
		\mathbf{y}^{\bpi}_{t + \Delta t} = \mathbf{y}^{\bpi}_t + \mathbf{a}^{\bpi}_t \cdot \Delta t.
\end{equation*}
Moreover, the DDIM simulator, which is simply Euler's method applied to a reparameterization of the ODE \eqref{eq:samplestate-ode} with $f\equiv 0$ and $g\equiv \sqrt{2}$, is given as follows (see \citealp{song2020denoising})
\begin{equation*}
	\mathbf{y}^{\mathbf{\pi}}_{t + \Delta t} = \mathbf{y}^{\mathbf{\pi}}_t +\left( \sqrt{\frac{1 - \alpha(t)}{\alpha(t)}}- \sqrt{\frac{1 - \alpha(t + \Delta t)}{\alpha(t + \Delta t)}}\right)\cdot \sqrt{2(T - t)}\mathbf{a}^{\bpi}_t,
\end{equation*}
where $\alpha(t) = \frac{1}{2(T - t) +1}$ and $\alpha(t) = 1$ for all $t\ge T$. The DDIM simulator behaves similarly as ODE-Euler when $\Delta t \rightarrow 0,$ but differs from the latter when $\Delta t$ is not small (i.e. with fewer sampling steps).
For comparison purposes, when implementing Algorithm~\ref{algo:offline episodic}, we also include the SDE simulator, which takes the following form when $f\equiv0$ and $g \equiv \sqrt{2}$ (similar to \eqref{eq:simulator}):
\begin{equation*}
	\mathbf{y}^{\mathbf{\pi}}_{t + \Delta t} = \mathbf{y}^{\mathbf{\pi}}_t + 2\mathbf{a}^{\bpi}_t \cdot \Delta t + \sqrt{2\Delta t}\cdot \xi,\quad  \xi \sim\mathcal{N}(0, I_d).
\end{equation*}
All the three simulators share the same prior distribution $\nu := \mathcal{N}(0, 2T\cdot I_d),$ which is obtained from \eqref{eq:hatp}.

\begin{figure}[h]
\centering
% \captionsetup{justification=centering}
\includegraphics[scale=0.35]{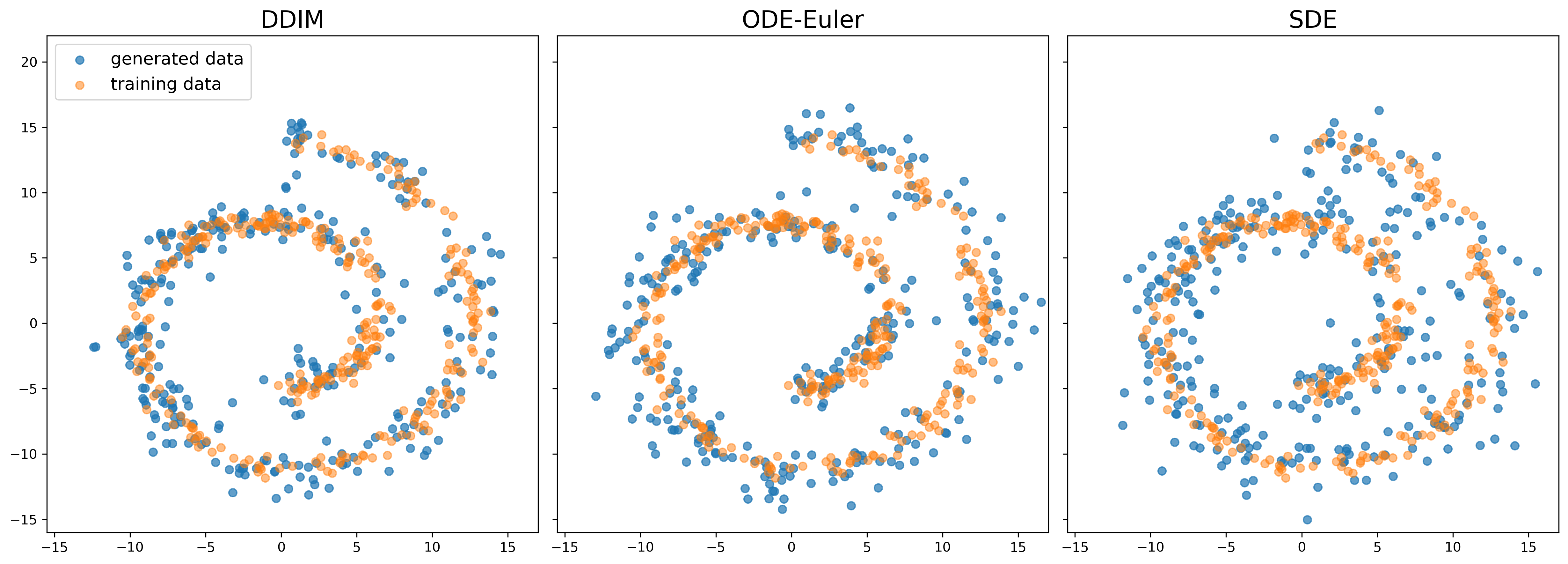}
\caption{300 samples generated by our algorithm for three simulators when $\beta = 0$ and $\Delta t = 0.5$.}
\label{fig:ddim_cmp_beta_0}
\end{figure}

In our  experiment,  we fix $T=5$ and vary $\Delta t$ (or equivalently the number of sampling steps $K := T/\Delta t$). The implementation of
Algorithm \ref{algo:offline episodic} with three different simulators/samplers is still based on the settings described in Tables \ref{table:NN_setting} and \ref{table:param_toy}. 
We first consider $\beta=0$, in which the algorithm aims to generate samples that are close to the original data without concerning the reward.
Table \ref{table:ode_cmp_beta_0} shows how the choice of a simulator  affects the output quality of the diffusion model. When there are 50 sampling steps, the RL algorithms trained under either the ODE or SDE simulator all generate reasonably good samples, although the two ODE-based ones have slightly better performances.  When the number of sampling steps is reduced, the quality of samples generated from all the simulators decrease, but the SDE-based one decreases more. 
Figure~\ref{fig:ddim_cmp_beta_0} displays visually the generated samples by different simulators.

\begin{table}[h]
    \centering
    \begin{tabular}{c | c | c c}
        \hline Simulator & Sampling Steps ($\Delta t$)& Reward ($\mathbb{E}[h(\mathbf{y}_T)]$) $\uparrow$ & KL$(p(\mathbf{y}_T) || p_0) \downarrow$ \\\hline
	DDIM & \multirow{3}{*}{50 (0.1)} & 0.44 $\pm$ 0.005 &  0.14 $\pm$ 0.018 \\	
	ODE-Euler & & 0.44 $\pm$ 0.006 &  \textbf{0.11 $\pm$ 0.016}  \\	
	SDE &   & 0.45 $\pm$ 0.005 & 0.18 $\pm$ 0.020 \\ \hline\hline
	DDIM & \multirow{3}{*}{10 (0.5)} & 0.45 $\pm$ 0.005 & \textbf{0.15 $\pm$ 0.019} \\		
	ODE-Euler & & 0.44 $\pm$ 0.006 & 0.23 $\pm$ 0.022 \\		
	SDE & & 0.46 $\pm$ 0.006 & 0.35 $\pm$ 0.024 \\\hline\hline
	DDIM & \multirow{3}{*}{5 (1.0)} & 0.46 $\pm$ 0.005 & 0.72 $\pm$ 0.022  \\	
	ODE-Euler & & 0.44 $\pm$ 0.006 & \textbf{0.56 $\pm$ 0.024}  \\	
	SDE & & 0.46 $\pm$ 0.005 & 0.83 $\pm$ 0.026  \\\hline
    \end{tabular}
% \captionsetup{justification=raggedright, singlelinecheck=false}
\caption{Empirical mean rewards and KL-divergence (with 95\% confidence) based on (100 batches of) 300 samples generated by our algorithm with ODE-based and SDE-based samplers with different sampling steps and $\beta = 0$.}
\label{table:ode_cmp_beta_0}
\end{table}

\begin{figure}[h]
	\centering
	% \captionsetup{justification=raggedright, singlelinecheck=false}
	\includegraphics[scale=0.35]{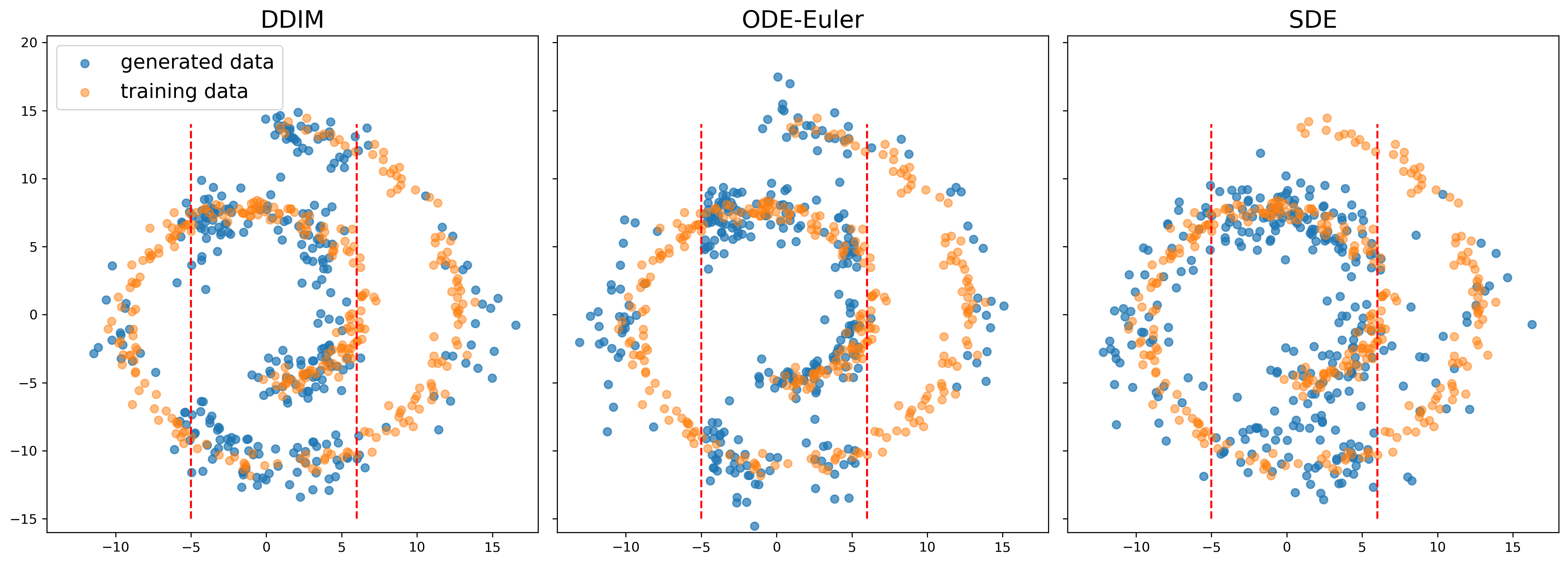}
	\caption{300 samples generated by our algorithm for three simulators with a (fixed) expected of 0.80 and $\Delta t = 0.5$.}
	\label{fig:ddim_cmp_beta_25}
\end{figure}

We next consider the reward-directed generative task with the same reward function $h(\mathbf{y}) = 1_{y_1\in[-5, 6]}$ as before, and study the performance of Algorithm~\ref{algo:offline episodic} with the three different simulators. Similar to Section~\ref{sec:finetue}, we train the diffusion model with a properly chosen reward weight $\beta$ so that the samples generated from the different simulators will earn the same mean reward of 0.80. The numerical results are presented in Table \ref{table:ode_cmp_beta_25}. We can see that when there are 50 sampling steps, the SDE-based sampler performs the best with a significantly small KL divergence between the generated distribution and the data distribution. However, as we increase $\Delta t$, the performance of the SDE-based sampler deteriorates and becomes inferior to the  ODE-based samplers. Figure~\ref{fig:ddim_cmp_beta_25} displays the generated samples out of the three simulators with a fixed expected reward of 0.8 and $\Delta t =0.5$.

To conclude this subsection, the ODE simulators prove to be effective when employed in conjunction with our RL formulation and q-Learning algorithm for reward-directed diffusion models. With a smaller number of sampling steps, the ODE simulators significantly reduce the sample generation time while maintaining an acceptable level of sample quality and reward. In our Algorithm \ref{algo:offline episodic}, running the SDE-based simulator consumes a significant proportion  of the total training time. Therefore, the ODE formulation is a competitive contender for our reward-directed RL diffusion model.

\begin{table}[h]
	\centering
	\begin{tabular}{c | c | c c}
		\hline Simulator & Sampling Steps ($\Delta t$) & Reward ($\mathbb{E}[h(\mathbf{y}_T)]$) $\uparrow$& KL$(p(\mathbf{y}_T) || p_0) \downarrow$  \\\hline
		DDIM & \multirow{3}{*}{50 (0.1)}& 0.80 $\pm$ 0.004 & 0.50 $\pm$ 0.027 \\	
		ODE-Euler && 0.80 $\pm$ 0.005 & 0.62 $\pm$ 0.022 \\	
		SDE &   & 0.80 $\pm$ 0.005 & \textbf{0.24 $\pm$ 0.024} \\ \hline\hline
		DDIM & \multirow{3}{*}{10 (0.5)}&  0.80 $\pm$ 0.004 & \textbf{0.61 $\pm$ 0.025} \\	
		ODE-Euler & &0.80 $\pm$ 0.004 & 0.72 $\pm$ 0.024  \\		
		SDE & & 0.80 $\pm$ 0.004 &0.77 $\pm$ 0.024\\\hline\hline
		DDIM & \multirow{3}{*}{5 (1.0)}&  0.80 $\pm$ 0.004 & 1.22 $\pm$ 0.028\\	
		ODE-Euler & &0.80 $\pm$ 0.005 & \textbf{1.07 $\pm$ 0.025} \\		
		SDE & & 0.80 $\pm$ 0.004 & 1.29 $\pm$ 0.027 \\\hline
	\end{tabular}
	% \captionsetup{justification=raggedright, singlelinecheck=false}
	\caption{Empirical mean rewards and KL-divergence (with 95\% confidence) based on (100 batches of) 300 samples generated by our algorithm with ODE-based and SDE-based samplers with different sampling steps when the (fixed) expected reward is 0.8.}
	\label{table:ode_cmp_beta_25}
\end{table}

%%%%%%%%%%%%%%%%%%%%%%%%%%%%%%%%%%%%

\subsection{Conditional diffusion models}
%cf. https://arxiv.org/pdf/2404.07771
Diffusion models are often used for conditional data generations, e.g., in text-to-image models. In this section, we briefly discuss the extension of our framework to conditional diffusion models. We take the SDE-based formulation as an illustration.

We first introduce some notations. Let $(\mathbf{x}_{0}, \mathbf{C})$ be a random vector in $\mathbb{R}^d \times \mathbb{R}^{d_c}$, where $\mathbf{x}_{0}$ represents the data (e.g. images) and $\mathbf{C}$ represents the condition/context (e.g. a text prompt or a class label). Denote by $P_{\mathbf{C}}$ the (marginal) distribution of $\mathbf{C}$, which is assumed to be either known or accessible through i.i.d. samples.
Denote $ p_{0}(\cdot|c) := \mathbb{P}(\mathbf{x}_{0} \in \cdot |\mathbf{C} = c )$ the conditional distribution of $\mathbf{x}_{0}$. Given $\mathbf{C} = c$,
the standard conditional diffusion model aims to generate samples from the (unknown) conditional data distribution $p_{0}(\cdot|c)$.

For a conditional diffusion model,
the forward process $\mathbf{x}_{t}$ (with a slight abuse of notations) at time $t$ is given by (see e.g. Section 2.2 of \citealp{chen2024overview}):
\begin{equation}\label{OU:SDE-condition}
d\mathbf{x}_{t}=-f(t)\mathbf{x}_{t}dt+g(t)d\mathbf{B}_{t},  \quad \mathbf{x}_{0} \sim p_{0}(\cdot|c),
\end{equation}
where the noise is added to the data $\mathbf{x}_{0}$, {\it but not to the context $c$}.
Denote by $p_{t}(\cdot|c)$ the conditional density function of $\mathbf{x}_t$ at time $t$ given $\mathbf{C} = c$.
Then the reverse-time SDE $(\mathbf{z}_{t})$ satisfies
\begin{equation}\label{eq:zt-c}
d\mathbf{z}_{t}=\left[f(T-t)\mathbf{z}_{t}+(g(T-t))^{2}\nabla\log p_{T-t}(\mathbf{z}_{t} |c)\right]dt
+g(T-t)d W_{t}, \quad \mathbf{z}_{0}\sim\nu,
\end{equation}
where the initialization distribution $\nu$ is still chosen to follow the normal distribution in \eqref{eq:hatp}, which is independent of $\mathbf{C} = c$. In \eqref{eq:zt-c}, the quantity $\nabla\log p_{T-t}(\cdot |c)$ is referred to as the conditional score function. In text-to-image models, classifier guidance \citep{dhariwal2021diffusion} or classifier-free guidance methods \citep{ho2022classifier} are often used to estimate such conditional score functions.

To adapt standard conditional diffusion models to reward maximization, we consider the following problem:
\begin{align}\label{eq:CT-RL-y}
 \max_{{\mathbf{a}} = (a_t: 0 \le t \le T)} \left\{ \beta \cdot \mathbb{E}[h( \mathbf{y}^{\mathbf{a}}_{T}, \mathbf{C})] -  \mathbb{E} \left[ \int_0^T (g(T-t))^{2} \cdot {| \nabla\log p_{T-t}({\mathbf{y}}^{\mathbf{a}}_{t} | \mathbf{C} ) - a_t|^2} dt \right]  \right\}
\end{align}
where
\begin{equation}\label{eq:u-sys-con}
d\mathbf{y}^{\mathbf{a}}_{t}=\left[f(T-t)\mathbf{y}^{\mathbf{a}}_{t}+(g(T-t))^{2} a_t \right]dt
+g(T-t)d W_{t}, \quad \mathbf{y}_{0}\sim\nu,
\end{equation}
and the expectation is taken with respect to the randomness in $(\mathbf{y}^{\mathbf{a}}_{t})$ and $\mathbf{C}$. In contrast to the unconditional diffusion model, here the reward function $h$ now depends also on the condition $\mathbf{C}$. Moreover, the state at time $t$ can be viewed as $(t, \mathbf{y}_{t}, \mathbf{C})$, and hence an optimal action process will also depend on the condition $\mathbf{C}.$
Define the running reward for this problem as follows:
\begin{align}\label{eq:rt-y}
 r(t, \mathbf{y}, c, a) := -(g(T-t))^{2} \cdot {| \nabla\log p_{T-t}({\mathbf{y}}|c) - a|^2}.
\end{align}

The exploratory formulation can be defined similarly as in Section~\ref{sec:explora}.
We let $\bpi: (t, y, c)  \rightarrow  \bpi (\cdot| t, y, c) \in \mathcal{P} (\mathbb{R}^d)$ be a given stochastic feedback policy, which now depends also on the condition $c$.
The exploratory value function is defined by
\begin{align*}
        &J(t, y, c, \bpi) \nonumber\\
        = &\mathbb{E}_{t,y, c}\bigg[  \int_t^T \int_{\mathbb{R}^d} \big( {r(s, \tilde{\mathbf{y}}_s^{\bpi}, c, a)}\ - { \theta \log \bpi(a|s, \tilde{\mathbf{y}}_{s}^{\bpi}, c)\big) \bpi(a|s, \tilde{\mathbf{y}}_{s}^{\bpi}, c) da} ds + {\beta h(\tilde{\mathbf{y}}_T^{\bpi}, c )}\bigg],
    \end{align*}
% \begin{align*} % \label{eq:J1-y}
% & J(t, y, c, \bpi)\\  %& = \mathbb{E}^{\bar{\mathbb{P}}}_{t,u, y}\left[  \int_t^T \left( r(s, \mathbf{y}_s^{\bpi}, y, a_s^{\bpi}) - \theta \log \bpi(a_s^{\bpi}|s, \mathbf{y}_{s}^{\bpi} ,y )\right) ds + \beta h(\mathbf{y}_T^{\bpi} , y )\right] \\
% & =  \mathbb{E}_{t, y, c}\left[  \int_t^T \left(  - g^2(T-t) \cdot {| \nabla\log p_{T-s}(\mathbf{y}_s^{\bpi}|c) -a_s^{\bpi}|^2}  - \theta \log \bpi(a_s^{\bpi}|s, \mathbf{y}_{s}^{\bpi}, c)\right) ds + \beta h(\mathbf{y}_T^{\bpi}, c)\right], \nonumber
% \end{align*}
where $\{\tilde{\mathbf{y}}_{s}^{\bpi}: 0 \le s \le T\}$ is the exploratory state process satisfying the SDE \eqref{eq: exp_state_SDE}, and
$\mathbb{E}_{t, y, c}$ denotes the expectation conditioned on $(t, \mathbf{y}_t^{\bpi}, \mathbf{C}) = (t, y , c )$.
We aim to solve the following optimization problem: %\textcolor{red}{does the averaging over initial distribution change results in q-learning?}
\begin{align*}% \label{eq:Jstar-y}
	\max_{\bpi } \iint J(0, y, c, \bpi) d \nu(y) dP_{\mathbf{C}}(c).
\end{align*}

This problem can be solved similarly as in the unconditional setting.
The running reward signal can be obtained in a similar manner as in the unconditional diffusion model discussed earlier. Specifically, given condition $\mathbf{C} = c,$ we have
\begin{align*}% \label{eq:scoreMC-cond}
\nabla_{\mathbf{x}}\log p_{t}(\mathbf{x}|c) & = \frac{ \nabla_{\mathbf{x}} p_{t}(\mathbf{x}|c)}{p_{t}(\mathbf{x}|c) } = \frac{\mathbb{E}_{\mathbf{x}_{0} \sim p_0(\cdot|c) } [  \nabla_{\mathbf{x}}  p_{t|0}(\mathbf{x} | \mathbf{x}_{0})]}{ \mathbb{E}_{\mathbf{x}_{0} \sim p_0(\cdot|c) } [ p_{t|0}(\mathbf{x} | \mathbf{x}_{0})] }  \nonumber \\
& = \frac{\mathbb{E}_{\mathbf{x}_{0} \sim p_0(\cdot|c) } \left[  p_{t|0}(\mathbf{x} | \mathbf{x}_{0})  \cdot  \frac{-( \mathbf{x}-e^{-\int_{0}^{t}f(s)ds}\mathbf{x}_{0}  ) }{ \int_{0}^{t}e^{-2\int_{s}^{t}f(v)dv}(g(s))^{2}ds} \right] }{ \mathbb{E}_{\mathbf{x}_{0} \sim p_0(\cdot|c) } [ p_{t|0}(\mathbf{x} | \mathbf{x}_{0})] } \nonumber \\
& = \frac{1} { \int_{0}^{t}e^{-2\int_{s}^{t}f(v)dv}(g(s))^{2}ds} \cdot \left( -  \mathbf{x} +  \frac{\mathbb{E}_{\mathbf{x}_{0} \sim p_0(\cdot|c) } \left[  p_{t|0}(\mathbf{x} | \mathbf{x}_{0})  \cdot  {\mathbf{x}_{0}  } \right] }{ \mathbb{E}_{\mathbf{x}_{0} \sim p_0(\cdot|c) } [ p_{t|0}(\mathbf{x} | \mathbf{x}_{0})] } \cdot e^{-\int_{0}^{t}f(s)ds}  \right).
\end{align*}
Therefore, given $m$ i.i.d. samples $( \mathbf{x}^{i}_{0} )$ from the conditional data distribution $p_0(\cdot |c),$
a simple ratio estimator for the true score $\nabla_{\mathbf{x}}\log p_{t}(\mathbf{x}|c) $ at given $t$ and $\mathbf{x}$ is given by
\begin{align*}%\label{eq:approxi-score}
\widehat{ \nabla_{\mathbf{x}}\log p_{t}(\mathbf{x} | c) }   = \frac{1} { \int_{0}^{t}e^{-2\int_{s}^{t}f(v)dv}(g(s))^{2}ds} \cdot \left( -  \mathbf{x} +  \frac{ \sum_{i=1}^m  \left[  p_{t|0}(\mathbf{x} | \mathbf{x}^{i}_{0})  \cdot  {\mathbf{x}^{i}_{0}  } \right] }{ \max\{ \sum_{i=1}^m  [ p_{t|0}(\mathbf{x} | \mathbf{x}^i_{0})], m \epsilon \} } \cdot e^{-\int_{0}^{t}f(s)ds}  \right),
\end{align*}
where $\epsilon>0$ is some prespecified small value. It  follow that we can obtain a noisy sample of the instantaneous reward \eqref{eq:rt-y} at time $t$ given by
\begin{align}\label{eq:sample-r-cont}
\hat r_t = -g^2(T-t) \cdot {|  \widehat{ \nabla\log p_{T- t}(\mathbf{y}_t |c) }  - a_t|^2}.
\end{align}

From the problem~\eqref{eq:CT-RL-y}, the Hamiltonian becomes
\begin{align*}% \label{eq:Hal-y}
&H(t, y, c, a, p,q) \nonumber \\
&=   -(g(T-t))^{2}  | \nabla\log p_{T-t}(y|c) - a|^2  +  [f(T-t) y +(g(T-t))^{2} a ] \circ p + \frac{1}{2}(g(T-t))^{2} \circ  q.
\end{align*}
This is still a quadratic function of $a$, and hence the optimal stochastic policy is still Gaussian as in Proposition~\ref{prop:optimal-policy}.
%Given the condition $y$, the RL problem is similar as for the unconditional model, and
Hence, given the condition $c$, we can consider Gaussian policies in the RL algorithm design:
\begin{align}\label{eq:pi-psi-y}
\bpi^{\psi}(\cdot|t, y, c )   \sim  \mathcal{N}\left(\mu^{\psi}(t, y, c), \frac{\theta}{2 g^2(T-t)}  \cdot I_{d} \right) \quad \text{for all $(t, y, c)$.}%\left(\cdot \mid  \right)
\end{align}
We also use $q^{\psi}$ as the function approximator for the optimal $q$-function which is given below:
 \begin{align}\label{eq:q-psi-y}
q^{\psi}(t, y, c, a) = - g^2(T-t) \cdot |a - \mu^{\psi}(t,y, c)|^2 - \frac{\theta d}{2} \log \left(\frac{\pi\theta}{g^2(T-t)} \right).
%T [\Sigma^{\psi}(t,u)]^{-1} [a - \mu^{\psi}(t,u)] + \frac{\theta}{2} \log \det   [\Sigma^{\psi}(t,u)]^{-1} - \frac{m \theta}{2} \log 2 \pi.
\end{align}

Algorithm~\ref{algo-cond:offline episodic} summarizes the resulting algorithm for reward maximization in the conditional diffusion model.

%%%%%%%%%%%%%%%%%%%%%%%%%%%%%%%%%%%%%%%%%%%%%
\begin{algorithm}[hbtp]
\caption{q-Learning Algorithm (SDE-based conditional generation)}
\textbf{Inputs}: condition/context distribution $P_{\mathbf{C}}$, horizon $T$, time step $\Delta t$, number of episodes $N$, number of mesh grids $K = T/\Delta t$, initial learning rates $\alpha_{\Theta},\alpha_{\psi}$ and a learning rate schedule function $l(\cdot)$ (a function of the number of episodes), functional forms  of parameterized  value function $J^{\Theta}(\cdot,\cdot, \cdot)$ and  $\mu^{\psi}(\cdot,\cdot, \cdot)$, temperature parameter $\theta$, functions $f, g$ in \eqref{OU:SDE}, and $\epsilon$ in \eqref{eq:approxi-score}.

%, functional forms of test functions $\bm{\xi}(t,u_{\cdot \wedge t},a_{\cdot \wedge t})$ and $\bm{\zeta}(t,u_{\cdot \wedge t},a_{\cdot \wedge t})$

\textbf{Required program}: environment simulator $(y', \hat r) = \textit{Environment}_{\Delta t}(t,y,a)$ that takes current time--state pair $(t,y)$ and action $a$ as inputs and generates state $y'$ (by a numerical solver of SDE \eqref{eq:u-sys-con}) at time $t+\Delta t$ and sample {instantaneous reward $\hat r$ (see \eqref{eq:sample-r-cont}) at time $t$ as outputs. Policy $\bpi^{\psi}(\cdot|t, y, c)$ in \eqref{eq:pi-psi-y}, and q-function $q^{\psi}(t, y, c, a)$ in \eqref{eq:q-psi-y}}.

\textbf{Learning procedure}:
\begin{algorithmic}
\STATE Initialize $\Theta,\psi$.
\FOR{episode $j=1$ \TO $N$} \STATE{Initialize $k = 0$. {Sample $c \sim P_{\mathbf{C}}$.} Sample initial state $y_0 \sim\nu$ and store $y_{t_k} \leftarrow  y_0$.

\WHILE{$k < K$} \STATE{
{Generate action $a_{t_k}\sim \bm{\pi}^{\psi}(\cdot|t_k,y_{t_k}, c)$.}

Apply $a_{t_k}$ to environment simulator $(y, \hat r) = Environment_{\Delta t}(t_k, y_{t_k}, a_{t_k})$, and observe new state $y$ and reward $\hat r$ as outputs. Store $y_{t_{k+1}} \leftarrow y$ and $r_{t_k} \leftarrow \hat r$.

Update $k \leftarrow k + 1$.
}
\ENDWHILE

For every $i = 0,1,\cdots,K-1$, compute and store test functions
{
\begin{align*}
\xi_{t_i}= \frac{\partial J^{\Theta}}{\partial \Theta}(t_{i},y_{t_{i}}, c ),  \quad \zeta_{t_i} = \frac{\partial q^{\psi}}{\partial \psi}(t_{i}, y_{t_{i}}, c, a_{t_i}).
\end{align*}	
Compute  % (note when $i=K-1$, we use $J^{\Theta}(t_{K}, u_{t_{K}}) =\beta h(u_{t_{K}}) = \beta h(u_T)$ instead of the value computed from NN)
\[ \Delta \Theta = \sum_{i=0}^{K-1} \xi_{t_i} \big[ J^{\Theta}(t_{i+1}, y_{t_{i+1}}, c ) - J^{\Theta}(t_{i},y_{t_{i}}, c) + r_{t_i}\Delta t -q^{\psi}(t_{i}, y_{t_{i}}, c, a_{t_i})\Delta t  \big], \]
\[
\Delta \psi =   \sum_{i=0}^{K-1}\zeta_{t_i}\big[ J^{\Theta}(t_{i+1},y_{t_{i+1}}, c) - J^{\Theta}(t_{i}, y_{t_{i}}, c) + r_{t_i}\Delta t - q^{\psi}(t_{i},y_{t_{i}}, c, a_{t_i})\Delta t \big] .
\]
}
Update $\Theta$ and $\psi$ by
\begin{align*}
 &\Theta \leftarrow \Theta + l(j)\alpha_{\Theta} \Delta \Theta , \\
& \psi \leftarrow \psi + l(j)\alpha_{\psi} \Delta \psi .
\end{align*}

}
\ENDFOR
\end{algorithmic}
\label{algo-cond:offline episodic}
\end{algorithm}

%%%%%%%%%%%%%%%%%%%%%%%%%%%%%%%%%%%%%%%%%%%

\section{Conclusions}\label{sec:conclusion}
In this paper, we provide a continuous-time RL framework for adapting score-based diffusion models to generate samples that maximize some reward function, without requiring the availability of a pretrained score function or attempting to learn the score function. The key idea is that of {\it model-free} and {\it data-driven}: optimization is based on a stream of (possibly noisy) score signals, instead of on a trained score model obtained from score matching. Our framework is general and applicable to both SDE and probability flow ODE based implementations of diffusion models. It also can be adapted readily to cover both pure score matching and fine-tuning pretrained models as special cases. Numerically, the resulting RL algorithms are shown to perform well on both low-dimensional synthetic data sets and high-dimensional CIFAR-10 image datasets. Our approach is conceptually different from the hitherto mainstream method of pretraining score functions followed by fine-tuning, offering an overarching way of accomplishing  generate AI tasks via diffusion models.

There are a couple of notable limitations of the paper that beg for further investigations.
As discussed in Section~\ref{sec:image}, while q-Learning does not need to learn the score function which is the most expensive part in diffusion models, its actor--critic structure requires substantial training time to achieve convergence at least for image generations. A promising future research direction lies in accelerating the training process, particularly when dealing with complex, high-dimensional datasets and diverse reward functions. On the other hand, the convergence analysis of the q-Learning algorithm in Section~\ref{sec:convergence} is very preliminary, with the result relying on certain assumptions, including the existence of Lyapunov functions, that are hard to verify. A specific interesting question is to identify  structural properties of neural network approximations for actor/critic  and of data distributions that would enable the validation of these assumptions.
In general, convergence and regret analysis in continuous-time RL for diffusion processes is an uncharted territory where most exciting research awaits.

%%%%%%%%%%%%%%%%%%%%%%%%%%%%%%%%%%%%%%%%%%%%%%%%%%%%%%%

\newpage
 \bibliographystyle{chicago}
\bibliography{generative}

\begin{thebibliography}{}

\bibitem[\protect\citeauthoryear{Anderson}{Anderson}{1982}]{Anderson1982}
Anderson, B. D.~O. (1982).
\newblock Reverse-time diffusion equation models.
\newblock {\em Stochastic Processes and their Applications\/}~{\em 12\/}(3),
  313--326.

\bibitem[\protect\citeauthoryear{Bengio, Lahlou, Deleu, Hu, Tiwari, and
  Bengio}{Bengio et~al.}{2023}]{bengio2023gflownet}
Bengio, Y., S.~Lahlou, T.~Deleu, E.~J. Hu, M.~Tiwari, and E.~Bengio (2023).
\newblock Gflownet foundations.
\newblock {\em Journal of Machine Learning Research\/}~{\em 24\/}(210), 1--55.

\bibitem[\protect\citeauthoryear{Benveniste, M{\'e}tivier, and
  Priouret}{Benveniste et~al.}{2012}]{benveniste2012adaptive}
Benveniste, A., M.~M{\'e}tivier, and P.~Priouret (2012).
\newblock {\em Adaptive algorithms and stochastic approximations}, Volume~22.
\newblock Springer Science \& Business Media.

\bibitem[\protect\citeauthoryear{Bhandari, Russo, and Singal}{Bhandari
  et~al.}{2021}]{bhandari2021finite}
Bhandari, J., D.~Russo, and R.~Singal (2021).
\newblock A finite time analysis of temporal difference learning with linear
  function approximation.
\newblock {\em Operations Research\/}~{\em 69\/}(3), 950--973.

\bibitem[\protect\citeauthoryear{Black, Janner, Du, Kostrikov, and
  Levine}{Black et~al.}{2024}]{black2023training}
Black, K., M.~Janner, Y.~Du, I.~Kostrikov, and S.~Levine (2024).
\newblock Training diffusion models with reinforcement learning.
\newblock In {\em The Twelfth International Conference on Learning
  Representations}.

\bibitem[\protect\citeauthoryear{Cattiaux, Conforti, Gentil, and
  L{\'e}onard}{Cattiaux et~al.}{2023}]{cattiaux2023time}
Cattiaux, P., G.~Conforti, I.~Gentil, and C.~L{\'e}onard (2023).
\newblock Time reversal of diffusion processes under a finite entropy
  condition.
\newblock {\em Annales de l'Institut Henri Poincar{\'e} (B) Probabilit{\'e}s et
  Statistiques\/}~{\em 59\/}(4), 1844--1881.

\bibitem[\protect\citeauthoryear{Chen, Mei, Fan, and Wang}{Chen
  et~al.}{2024}]{chen2024overview}
Chen, M., S.~Mei, J.~Fan, and M.~Wang (2024).
\newblock An overview of diffusion models: Applications, guided generation,
  statistical rates and optimization.
\newblock {\em arXiv preprint arXiv:2404.07771\/}.

\bibitem[\protect\citeauthoryear{Clark, Vicol, Swersky, and Fleet}{Clark
  et~al.}{2024}]{clark2023directly}
Clark, K., P.~Vicol, K.~Swersky, and D.~J. Fleet (2024).
\newblock Directly fine-tuning diffusion models on differentiable rewards.
\newblock In {\em The Twelfth International Conference on Learning
  Representations}.

\bibitem[\protect\citeauthoryear{Dhariwal and Nichol}{Dhariwal and
  Nichol}{2021}]{dhariwal2021diffusion}
Dhariwal, P. and A.~Nichol (2021).
\newblock Diffusion models beat {GAN}s on image synthesis.
\newblock {\em Advances in Neural Information Processing Systems\/}~{\em 34},
  8780--8794.

\bibitem[\protect\citeauthoryear{Fan, Watkins, Du, Liu, Ryu, Boutilier, Abbeel,
  Ghavamzadeh, Lee, and Lee}{Fan et~al.}{2023}]{fan2024reinforcement}
Fan, Y., O.~Watkins, Y.~Du, H.~Liu, M.~Ryu, C.~Boutilier, P.~Abbeel,
  M.~Ghavamzadeh, K.~Lee, and K.~Lee (2023).
\newblock Reinforcement learning for fine-tuning text-to-image diffusion
  models.
\newblock {\em Advances in Neural Information Processing Systems\/}~{\em 36}.

\bibitem[\protect\citeauthoryear{Haarnoja, Zhou, Abbeel, and Levine}{Haarnoja
  et~al.}{2018}]{haarnoja2018soft}
Haarnoja, T., A.~Zhou, P.~Abbeel, and S.~Levine (2018).
\newblock Soft actor-critic: Off-policy maximum entropy deep reinforcement
  learning with a stochastic actor.
\newblock In {\em International Conference on Machine Learning}, pp.\
  1861--1870. PMLR.

\bibitem[\protect\citeauthoryear{Haussmann and Pardoux}{Haussmann and
  Pardoux}{1986}]{haussmann1986time}
Haussmann, U.~G. and E.~Pardoux (1986).
\newblock Time reversal of diffusions.
\newblock {\em The Annals of Probability\/}, 1188--1205.

\bibitem[\protect\citeauthoryear{Heusel, Ramsauer, Unterthiner, Nessler, and
  Hochreiter}{Heusel et~al.}{2017}]{heusel2017gans}
Heusel, M., H.~Ramsauer, T.~Unterthiner, B.~Nessler, and S.~Hochreiter (2017).
\newblock {GAN}s trained by a two time-scale update rule converge to a local
  {N}ash equilibrium.
\newblock In {\em Advances in Neural Information Processing Systems},
  Volume~30.

\bibitem[\protect\citeauthoryear{Ho, Jain, and Abbeel}{Ho
  et~al.}{2020}]{Ho2020}
Ho, J., A.~Jain, and P.~Abbeel (2020).
\newblock Denoising diffusion probabilistic models.
\newblock In {\em Advances in Neural Information Processing Systems},
  Volume~33.

\bibitem[\protect\citeauthoryear{Ho and Salimans}{Ho and
  Salimans}{2022}]{ho2022classifier}
Ho, J. and T.~Salimans (2022).
\newblock Classifier-free diffusion guidance.
\newblock {\em arXiv preprint arXiv:2207.12598\/}.

\bibitem[\protect\citeauthoryear{Hoogeboom, Satorras, Vignac, and
  Welling}{Hoogeboom et~al.}{2022}]{hoogeboom2022equivariant}
Hoogeboom, E., V.~G. Satorras, C.~Vignac, and M.~Welling (2022).
\newblock Equivariant diffusion for molecule generation in 3d.
\newblock In {\em International conference on machine learning}, pp.\
  8867--8887. PMLR.

\bibitem[\protect\citeauthoryear{Hyv{\"a}rinen and Dayan}{Hyv{\"a}rinen and
  Dayan}{2005}]{hyvarinen2005estimation}
Hyv{\"a}rinen, A. and P.~Dayan (2005).
\newblock Estimation of non-normalized statistical models by score matching.
\newblock {\em Journal of Machine Learning Research\/}~{\em 6\/}(4), 695--708.

\bibitem[\protect\citeauthoryear{Jia, Ouyang, and Zhang}{Jia
  et~al.}{2025}]{jia2025accuracy}
Jia, Y., D.~Ouyang, and Y.~Zhang (2025).
\newblock Accuracy of discretely sampled stochastic policies in continuous-time
  reinforcement learning.
\newblock {\em arXiv preprint arXiv:2503.09981\/}.

\bibitem[\protect\citeauthoryear{Jia and Zhou}{Jia and Zhou}{2022a}]{JZ21}
Jia, Y. and X.~Y. Zhou (2022a).
\newblock Policy evaluation and temporal-difference learning in continuous time
  and space: A martingale approach.
\newblock {\em Journal of Machine Learning Research\/}~{\em 23}, (154):1--55.

\bibitem[\protect\citeauthoryear{Jia and Zhou}{Jia and Zhou}{2022b}]{JZ22}
Jia, Y. and X.~Y. Zhou (2022b).
\newblock Policy gradient and actor-critic learning in continuous time and
  space: Theory and algorithms.
\newblock {\em Journal of Machine Learning Research\/}~{\em 23}, (275):1--50.

\bibitem[\protect\citeauthoryear{Jia and Zhou}{Jia and Zhou}{2023}]{jia2023q}
Jia, Y. and X.~Y. Zhou (2023).
\newblock q-learning in continuous time.
\newblock {\em Journal of Machine Learning Research\/}~{\em 24\/}(161), 1--61.

\bibitem[\protect\citeauthoryear{Jia and Zhou}{Jia and
  Zhou}{2025}]{jia2025erratum}
Jia, Y. and X.~Y. Zhou (2025).
\newblock Erratum to “q-learning in continuous time”.
\newblock {\em Available at
  \url{https://www.columbia.edu/~xz2574/download/err.pdf}\/}.

\bibitem[\protect\citeauthoryear{Karras, Aittala, Aila, and Laine}{Karras
  et~al.}{2022}]{Karras2022}
Karras, T., M.~Aittala, T.~Aila, and S.~Laine (2022).
\newblock Elucidating the design space of diffusion-based generative models.
\newblock In {\em Advances in Neural Information Processing Systems},
  Volume~35.

\bibitem[\protect\citeauthoryear{Kingma and Ba}{Kingma and
  Ba}{2015}]{kingma2014adam}
Kingma, D.~P. and J.~Ba (2015).
\newblock Adam: A method for stochastic optimization.
\newblock In {\em 3rd International Conference on Learning Representations}.

\bibitem[\protect\citeauthoryear{Krizhevsky}{Krizhevsky}{2009}]{krizhevsky2009learning}
Krizhevsky, A. (2009).
\newblock Learning multiple layers of features from tiny images.

\bibitem[\protect\citeauthoryear{Kushner and Yin}{Kushner and
  Yin}{2003}]{kushner2003stochastic}
Kushner, H. and G.~Yin (2003).
\newblock Stochastic approximation and recursive algorithms.
\newblock In {\em Stochastic Modelling and Applied Probability}, Volume~35.
  Springer-Verlag NY.

\bibitem[\protect\citeauthoryear{Lahlou, Deleu, Lemos, Zhang, Volokhova,
  Hern{\'a}ndez-Garc{\i}a, Ezzine, Bengio, and Malkin}{Lahlou
  et~al.}{2023}]{lahlou2023theory}
Lahlou, S., T.~Deleu, P.~Lemos, D.~Zhang, A.~Volokhova,
  A.~Hern{\'a}ndez-Garc{\i}a, L.~N. Ezzine, Y.~Bengio, and N.~Malkin (2023).
\newblock A theory of continuous generative flow networks.
\newblock In {\em International Conference on Machine Learning}, pp.\
  18269--18300. PMLR.

\bibitem[\protect\citeauthoryear{Lai, Takida, Murata, Uesaka, Mitsufuji, and
  Ermon}{Lai et~al.}{2023}]{lai2023fp}
Lai, C.-H., Y.~Takida, N.~Murata, T.~Uesaka, Y.~Mitsufuji, and S.~Ermon (2023).
\newblock {FP-D}iffusion: Improving score-based diffusion models by enforcing
  the underlying score fokker-planck equation.
\newblock In {\em International Conference on Machine Learning}, pp.\
  18365--18398. PMLR.

\bibitem[\protect\citeauthoryear{Lee, Lu, and Tan}{Lee
  et~al.}{2022}]{leeconvergence}
Lee, H., J.~Lu, and Y.~Tan (2022).
\newblock Convergence for score-based generative modeling with polynomial
  complexity.
\newblock In {\em Advances in Neural Information Processing Systems},
  Volume~35.

\bibitem[\protect\citeauthoryear{Lee, Liu, Ryu, Watkins, Du, Boutilier, Abbeel,
  Ghavamzadeh, and Gu}{Lee et~al.}{2023}]{lee2023aligning}
Lee, K., H.~Liu, M.~Ryu, O.~Watkins, Y.~Du, C.~Boutilier, P.~Abbeel,
  M.~Ghavamzadeh, and S.~S. Gu (2023).
\newblock Aligning text-to-image models using human feedback.
\newblock {\em arXiv preprint arXiv:2302.12192\/}.

\bibitem[\protect\citeauthoryear{Lu, Zhou, Bao, Chen, Li, and Zhu}{Lu
  et~al.}{2022}]{lu2022dpm}
Lu, C., Y.~Zhou, F.~Bao, J.~Chen, C.~Li, and J.~Zhu (2022).
\newblock {DPM}-solver: A fast {ODE} solver for diffusion probabilistic model
  sampling in around 10 steps.
\newblock In {\em Advances in Neural Information Processing Systems},
  Volume~35, pp.\  5775--5787.

\bibitem[\protect\citeauthoryear{Ramesh, Dhariwal, Nichol, Chu, and
  Chen}{Ramesh et~al.}{2022}]{ramesh2022hierarchical}
Ramesh, A., P.~Dhariwal, A.~Nichol, C.~Chu, and M.~Chen (2022).
\newblock Hierarchical text-conditional image generation with {CLIP} latents.
\newblock {\em arXiv preprint arXiv:2204.06125\/}.

\bibitem[\protect\citeauthoryear{Rombach, Blattmann, Lorenz, Esser, and
  Ommer}{Rombach et~al.}{2022}]{rombach2022high}
Rombach, R., A.~Blattmann, D.~Lorenz, P.~Esser, and B.~Ommer (2022).
\newblock High-resolution image synthesis with latent diffusion models.
\newblock In {\em Proceedings of the IEEE/CVF Conference on Computer Vision and
  Pattern Recognition}, pp.\  10684--10695.

\bibitem[\protect\citeauthoryear{Ronneberger, Fischer, and Brox}{Ronneberger
  et~al.}{2015}]{ronneberger2015u}
Ronneberger, O., P.~Fischer, and T.~Brox (2015).
\newblock U-net: Convolutional networks for biomedical image segmentation.
\newblock In {\em Medical image computing and computer-assisted
  intervention--MICCAI 2015: 18th international conference, Munich, Germany,
  October 5-9, 2015, proceedings, part III 18}, pp.\  234--241. Springer.

\bibitem[\protect\citeauthoryear{Schulman, Moritz, Levine, Jordan, and
  Abbeel}{Schulman et~al.}{2015}]{schulman2015high}
Schulman, J., P.~Moritz, S.~Levine, M.~Jordan, and P.~Abbeel (2015).
\newblock High-dimensional continuous control using generalized advantage
  estimation.
\newblock {\em arXiv preprint arXiv:1506.02438\/}.

\bibitem[\protect\citeauthoryear{Sohl-Dickstein, Weiss, Maheswaranathan, and
  Ganguli}{Sohl-Dickstein et~al.}{2015a}]{Sohl2015}
Sohl-Dickstein, J., E.~Weiss, N.~Maheswaranathan, and S.~Ganguli (2015a).
\newblock Deep unsupervised learning using nonequilibrium thermodynamics.
\newblock In {\em International Conference on Machine Learning}, Volume~37,
  pp.\  2256--2265. PMLR.

\bibitem[\protect\citeauthoryear{Sohl-Dickstein, Weiss, Maheswaranathan, and
  Ganguli}{Sohl-Dickstein et~al.}{2015b}]{sohl2015deep}
Sohl-Dickstein, J., E.~Weiss, N.~Maheswaranathan, and S.~Ganguli (2015b).
\newblock Deep unsupervised learning using nonequilibrium thermodynamics.
\newblock In {\em International conference on machine learning}, pp.\
  2256--2265. PMLR.

\bibitem[\protect\citeauthoryear{Song, Meng, and Ermon}{Song
  et~al.}{2020}]{song2020denoising}
Song, J., C.~Meng, and S.~Ermon (2020).
\newblock Denoising diffusion implicit models.
\newblock In {\em International Conference on Learning Representations}.

\bibitem[\protect\citeauthoryear{Song, Durkan, Murray, and Ermon}{Song
  et~al.}{2021}]{song2021maximum}
Song, Y., C.~Durkan, I.~Murray, and S.~Ermon (2021).
\newblock Maximum likelihood training of score-based diffusion models.
\newblock {\em Advances in Neural Information Processing Systems\/}~{\em 34},
  1415--1428.

\bibitem[\protect\citeauthoryear{Song and Ermon}{Song and
  Ermon}{2019}]{SongErmon2019}
Song, Y. and S.~Ermon (2019).
\newblock Generative modeling by estimating gradients of the data distribution.
\newblock In {\em Advances in Neural Information Processing Systems},
  Volume~32.

\bibitem[\protect\citeauthoryear{Song, Garg, Shi, and Ermon}{Song
  et~al.}{2020}]{song2020sliced}
Song, Y., S.~Garg, J.~Shi, and S.~Ermon (2020).
\newblock Sliced score matching: A scalable approach to density and score
  estimation.
\newblock In {\em Uncertainty in Artificial Intelligence}, pp.\  574--584.
  PMLR.

\bibitem[\protect\citeauthoryear{Song, Sohl-Dickstein, Kingma, Kumar, Ermon,
  and Poole}{Song et~al.}{2021}]{SongICLR2021}
Song, Y., J.~Sohl-Dickstein, D.~P. Kingma, A.~Kumar, S.~Ermon, and B.~Poole
  (2021).
\newblock Score-based generative modeling through stochastic differential
  equations.
\newblock In {\em International Conference on Learning Representations}.

\bibitem[\protect\citeauthoryear{Tang}{Tang}{2024}]{tang2024fine}
Tang, W. (2024).
\newblock Fine-tuning of diffusion models via stochastic control: entropy
  regularization and beyond.
\newblock {\em arXiv preprint arXiv:2403.06279\/}.

\bibitem[\protect\citeauthoryear{Tang, Zhang, and Zhou}{Tang
  et~al.}{2022}]{tang2022exploratory}
Tang, W., Y.~P. Zhang, and X.~Y. Zhou (2022).
\newblock Exploratory hjb equations and their convergence.
\newblock {\em SIAM Journal on Control and Optimization\/}~{\em 60\/}(6),
  3191--3216.

\bibitem[\protect\citeauthoryear{Uehara, Zhao, Black, Hajiramezanali, Scalia,
  Diamant, Tseng, Biancalani, and Levine}{Uehara et~al.}{2024}]{uehara2024fine}
Uehara, M., Y.~Zhao, K.~Black, E.~Hajiramezanali, G.~Scalia, N.~L. Diamant,
  A.~M. Tseng, T.~Biancalani, and S.~Levine (2024).
\newblock Fine-tuning of continuous-time diffusion models as
  entropy-regularized control.
\newblock {\em arXiv preprint arXiv:2402.15194\/}.

\bibitem[\protect\citeauthoryear{Vincent}{Vincent}{2011}]{vincent2011connection}
Vincent, P. (2011).
\newblock A connection between score matching and denoising autoencoders.
\newblock {\em Neural Computation\/}~{\em 23\/}(7), 1661--1674.

\bibitem[\protect\citeauthoryear{Wang, Zariphopoulou, and Zhou}{Wang
  et~al.}{2020}]{Wang20}
Wang, H., T.~Zariphopoulou, and X.~Y. Zhou (2020).
\newblock Reinforcement learning in continuous time and space: A stochastic
  control approach.
\newblock {\em Journal of Machine Learning Research\/}~{\em 21}, (198):1--34.

\bibitem[\protect\citeauthoryear{Wang, Kulkarni, and Verd{\'u}}{Wang
  et~al.}{2009}]{wang2009divergence}
Wang, Q., S.~R. Kulkarni, and S.~Verd{\'u} (2009).
\newblock Divergence estimation for multidimensional densities via $ k
  $-nearest-neighbor distances.
\newblock {\em IEEE Transactions on Information Theory\/}~{\em 55\/}(5),
  2392--2405.

\bibitem[\protect\citeauthoryear{Wu, Gong, Liu, Ye, and Liu}{Wu
  et~al.}{2022}]{wu2022diffusion}
Wu, L., C.~Gong, X.~Liu, M.~Ye, and Q.~Liu (2022).
\newblock Diffusion-based molecule generation with informative prior bridges.
\newblock {\em Advances in Neural Information Processing Systems\/}~{\em 35},
  36533--36545.

\bibitem[\protect\citeauthoryear{Yang, Zhang, Song, Hong, Xu, Zhao, Shao,
  Zhang, Cui, and Yang}{Yang et~al.}{2023}]{yang2022diffusion}
Yang, L., Z.~Zhang, Y.~Song, S.~Hong, R.~Xu, Y.~Zhao, Y.~Shao, W.~Zhang,
  B.~Cui, and M.-H. Yang (2023).
\newblock Diffusion models: A comprehensive survey of methods and applications.
\newblock {\em ACM Copmuting Surveys\/}~{\em 56\/}(4), 1--39.

\bibitem[\protect\citeauthoryear{Yong and Zhou}{Yong and
  Zhou}{2012}]{yong2012stochastic}
Yong, J. and X.~Y. Zhou (2012).
\newblock {\em Stochastic controls: Hamiltonian systems and HJB equations},
  Volume~43.
\newblock Springer Science \& Business Media.

\bibitem[\protect\citeauthoryear{Zhang, Chen, Malkin, and Bengio}{Zhang
  et~al.}{2022}]{zhang2022unifying}
Zhang, D., R.~T. Chen, N.~Malkin, and Y.~Bengio (2022).
\newblock Unifying generative models with gflownets and beyond.
\newblock {\em arXiv preprint arXiv:2209.02606\/}.

\bibitem[\protect\citeauthoryear{Zhang, Zhang, Gu, Zhang, Susskind, Jaitly, and
  Zhai}{Zhang et~al.}{2024}]{zhang2024improving}
Zhang, D., Y.~Zhang, J.~Gu, R.~Zhang, J.~Susskind, N.~Jaitly, and S.~Zhai
  (2024).
\newblock Improving {GFlowNets} for text-to-image diffusion alignment.
\newblock {\em arXiv preprint arXiv:2406.00633\/}.

\bibitem[\protect\citeauthoryear{Zhang and Chen}{Zhang and
  Chen}{2023}]{zhang2023fast}
Zhang, Q. and Y.~Chen (2023).
\newblock Fast sampling of diffusion models with exponential integrator.
\newblock In {\em International Conference on Learning Representations}.

\bibitem[\protect\citeauthoryear{Zhao, Haoxian, Zhang, Yao, and Tang}{Zhao
  et~al.}{2024}]{tang2024score}
Zhao, H., C.~Haoxian, J.~Zhang, D.~Yao, and W.~Tang (2024).
\newblock Scores as actions: a framework of fine-tuning diffusion models by
  continuous-time reinforcement learning.
\newblock {\em arXiv preprint arXiv:2409.08400\/}.

\end{thebibliography}

\end{document}